\documentclass{article} 
\usepackage[preprint]{colm2026_conference}

\usepackage{microtype}
\usepackage{hyperref}
\usepackage{url}
\usepackage{array}
\usepackage{booktabs}
\usepackage{multirow}
\usepackage{titletoc}
\usepackage{graphicx}
\usepackage{tikz}
\usetikzlibrary{positioning,arrows.meta,calc}
\usepackage{listings}
\lstdefinestyle{promptstyle}{%
  basicstyle=\ttfamily\footnotesize,
  breaklines=true,
  breakatwhitespace=true,
  columns=fullflexible,
  frame=leftline,
  rulecolor=\color{black},
  framerule=0.4pt,
  xleftmargin=0.6em,
  framexleftmargin=0.35em,
}
\usepackage{numprint}
\usepackage{xspace}
\usepackage{placeins}
\usepackage{longtable}
\usepackage{caption}
\usepackage[breakable, skins]{tcolorbox}
\usepackage{fvextra}
\usepackage{floatrow}

\usepackage{lineno}
\usepackage{framed}
\usepackage{enumitem}

\usepackage{amsmath,amsfonts,amssymb,bm}

\def\eqref#1{equation~\ref{#1}}

\def\1{\bm{1}}

\DeclareMathAlphabet{\mathsfit}{\encodingdefault}{\sfdefault}{m}{sl}
\SetMathAlphabet{\mathsfit}{bold}{\encodingdefault}{\sfdefault}{bx}{n}

\newcommand{\name}{\textsc{CoShop}\xspace}
\newcommand{\namestructured}{\textsc{CoShop-structured}\xspace}

\newcommand{\frameworkname}{\textsc{CoPref}\xspace}
\newcommand{\taskhm}{H\&M\xspace}
\newcommand{\hmtask}{\taskhm}
\newcommand{\movielens}{MovieLens\xspace}
\newcommand{\goodreads}{Goodreads\xspace}

\newcommand{\claudehaiku}{\texttt{claude-haiku-4.5}\xspace}
\newcommand{\claudesonnet}{\texttt{claude-sonnet-4.6}\xspace}
\newcommand{\gptmini}{\texttt{gpt-4.1-mini}\xspace}
\newcommand{\gptfive}{\texttt{gpt-5.2}\xspace}

\newcommand{\oss}{\texttt{gpt-oss-120b}\xspace}

\newcommand{\qwenvlfourbinstruct}{\texttt{Qwen/Qwen3-VL-4B-Instruct}\xspace}

\newcommand{\qwenthreeembeddingeightb}{\texttt{Qwen-3-Embedding-8B}\xspace}

\newcommand{\update}[2]{\text{\texttt{update}}(#1 \mid #2)}

\newenvironment{walkuserturn}{%
  \par\noindent
  \setlength{\FrameSep}{3pt}%
  \setlength{\FrameRule}{0.4pt}%
  \begin{framed}%
  \noindent{\small\bfseries User turn}%
}{%
  \end{framed}%
  \noindent\ignorespacesafterend
}

\newenvironment{walkagentturn}{%
  \par\noindent
  \setlength{\FrameSep}{3pt}%
  \setlength{\FrameRule}{0.4pt}%
  \begin{framed}%
  \noindent{\small\bfseries Agent turn}%
}{%
  \end{framed}%
  \noindent\ignorespacesafterend
}
\newenvironment{walkuserstate}{%
  \noindent\textbf{\footnotesize Current state} $S_t \leftarrow \update{S_{t-1}}{m_{t-1}^{\text{agent}}}$\par\vspace{2pt}%
  \footnotesize
}{%
  \vspace{2pt}%
}

\newenvironment{walkactions}{%
  \noindent\textit{\footnotesize Actions:} \footnotesize\ignorespaces
}{%
  \par\vspace{2pt}%
}

\newenvironment{walkmessage}{%
  \par\noindent\textit{\footnotesize Message:} \footnotesize\ignorespaces
}{%
}

\newenvironment{walkactivations}{%
\footnotesize
  \begin{tabular}{@{}p{0.26\linewidth}p{0.32\linewidth}p{0.32\linewidth}@{}}%
  \toprule
  Feature & Revealed value & Activation source \\
  \midrule
}{%
  \bottomrule
  \end{tabular}%
  \vspace{2pt}%
}

\newcommand{\walkactivationrow}[3]{#1 & #2 & #3 \\}

\newcommand{\agentsearch}[2]{\textit{Searched for ``#1''; reviewed #2 item\ifnum#2=1\relax\else s\fi.}}

\DefineVerbatimEnvironment{walkjson}{Verbatim}{%
  breaklines=true,
  breakanywhere=true,
  breaksymbolleft={},
  breaksymbolright={},
  breaksymbolindentleftnchars=0,
  breaksymbolindentrightnchars=0,
  fontsize=\footnotesize,
}

\newcolumntype{C}{>{\footnotesize\raggedright\arraybackslash}p}

\definecolor{darkblue}{rgb}{0, 0, 0.5}
\hypersetup{colorlinks=true, citecolor=darkblue, linkcolor=darkblue, urlcolor=darkblue}

\title{Beyond expert users: agents should help users construct preferences, not just elicit them}

\author{Irena Saracay, Ludwig Schmidt, and Carlos Guestrin \\
Stanford University\\
\texttt{irena@cs.stanford.edu}
}

\begin{document}

\ifcolmsubmission
\linenumbers
\fi

\maketitle

\begin{abstract}
Agents typically assume an expert user ---
one with well-formed preferences about what they want ---
and default to clarifying questions whenever the task is underspecified.
We argue this assumption is unrealistic.
Users often lack the domain knowledge to have completely specified preferences;
if asked about their preference on some feature,
the user may be unable to answer without the agent helping the user to \textit{learn} some domain knowledge needed to form a preference for that feature, e.g., via examples or explanations.
To formalize these principles, we draw on the Search-Experience-Credence framework from Information Economics to introduce \frameworkname,
a model of how users construct preferences based on agent dialog actions.
We then study these ideas concretely in agentic recommender systems,
proposing \name, an interactive benchmark.
In \name, an agent converses with and makes recommendations for a \frameworkname user.
The agent's performance depends on whether it can help the user gain the knowledge needed to specify the task well.
Evaluating five frontier models, 
we find that no agent exceeds 56\% accuracy on \name despite five turns of interaction.
Failures stem not from agents' ability to find items, but from how little the interaction expands what users know about what they want.
\end{abstract}

\section{Introduction}\label{sec:intro}
\begin{figure}[bt]
    \centering
    \includegraphics[width=0.9\textwidth]{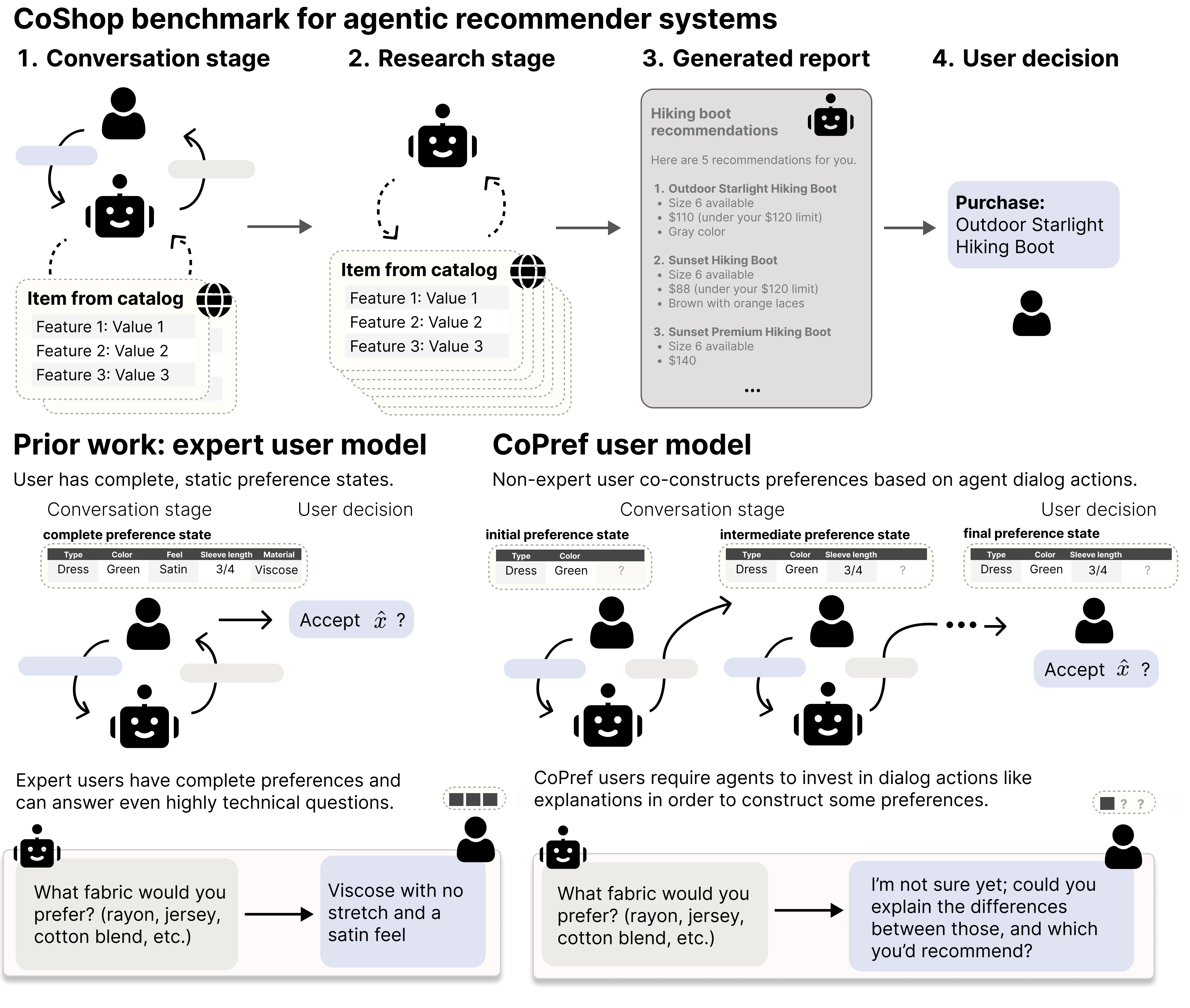}
    \caption{\textit{Top:} Interactive agents, like agentic recommender systems, chat with users to understand their task preferences. \textit{Bottom:} Rather than assuming these users are \textit{experts} with static, well-formed task preferences that are simply retrieved, we propose \frameworkname, a model of users which \textit{co-construct} preferences based on agents' dialog actions. We then introduce \name, a benchmark evaluating agents with \frameworkname users in the recommendation setting.\vspace{-1em}}
    \label{fig:hero}
\end{figure}

Language models are increasingly deployed as agents that interact with users to solve some task.
For example, agentic recommender systems, 
like ChatGPT Shopping Research \citep{chatgptshopping}, Perplexity Shopping \citep{perplexityshopping}, and Amazon's Rufus \citep{rufus}, 
elicit a user's shopping intent via chat, search for relevant items using tools, and recommend items in reports for users to select from (Figure \ref{fig:hero} top).
A quiet assumption underlying most such systems --- and the interactive benchmarks that evaluate them --- 
is that agents are interacting with \textit{expert users}: 
users who can specify their preferences fully when pressed 
\citep{qian2025userbench,pan2025benchmarks,vijayvargiya2025interactive}.
In this work, we challenge this assumption.

Imagine asking an agentic recommender system for hiking boot recommendations.
You make a good faith attempt to describe your requirements (size, price, preferred color),
and the agent returns a written report with a set of recommendations.
Whether you end up with the right boot depends not on the agent's search skill, but on whether it helped you develop your preferences for features that matter --- even ones you didn't know to ask about. 
As a non-expert hiker, you may not realize that boots can run narrow or wide, or that lug depth affects what terrain you can safely hike on --- but once you do, these features change both whether you can give the agent useful information to search with, and whether you can evaluate its recommendations. 
A truly helpful agent surfaces important, originally unknown features unprompted, before they become a bad purchase.

This example illustrates that real users are often not experts. Users may lack the domain knowledge to have well-formed preferences in the first place --- they not only underspecify their initial requests, but are frequently unaware of what features they could form preferences over in the first place \citep{kim1999interactive,haubl2000consumer,song2021purchase}. 
A user who has never heard of lug depth cannot answer a question about it; agents must instead \textit{teach} this domain knowledge --- through intentional examples and proactive explanations, not just clarifying questions --- before users can form a preference over it. 
The same gap also limits a user's ability to act on recommendations: even if a report contains the ideal pair of boots, a user who hasn't yet formed a preference about width can't reliably tell it apart from similar-looking options.

To formalize these principles, we propose \frameworkname (\textbf{Co}-constructed \textbf{Pref}erences), a model of how non-expert users \textit{construct} --- rather than simply reveal --- preferences based on agent dialog actions \citep{fischhoff1991value,bettman1998constructive}. 
In \frameworkname, users maintain a preference state --- a subset of their task preferences they are aware of, which begins incomplete and evolves each turn depending on agent actions (Figure \ref{fig:hero} bottom). 
Critically, not all features are equally easy to construct preferences over, a distinction we formalize using the \textbf{Search-Experience-Credence (SEC)} framework from Information Economics \citep{nelson1970information, darby1973free}: some features can be elicited by a simple question, some can only be recognized once the user has seen or compared real options, 
and some require the agent to provide a technical explanation that increases the user's domain knowledge. 
This means an agent cannot rely on clarifying questions alone --- it must learn which features typically demand which dialog action across users.
A user's final preference state, accumulated over the course of the interaction, also determines how well they can evaluate solutions.

We study these ideas concretely in the setting of agentic recommender systems, 
where we introduce \name, a benchmark evaluating agents with \frameworkname users 
across fashion, movie, and book recommendation domains.
\name measures performance by the user-agent \textit{team's} item accuracy: 
in each instance, an agent converses with a \frameworkname user to learn about their preferences, searches for relevant items, and writes a report of $k$ recommendations (Figure \ref{fig:hero} top). 
The user then selects one of the $k$ items based on their preference state at that point in the conversation.
Team accuracy requires two things to go right: the target item must be among the $k$  recommendations, 
and the user's preference state must be rich enough for them to identify it.
An agent that searches well but teaches the user nothing leaves the task unspecified and the user unable to select the correct item.

In evaluations of RAG agents built from five frontier language models, 
we find that all models significantly drop in performance when faced with \frameworkname users ---
e.g., while \claudesonnet scores 94.3\% accuracy given fully-specified preferences upfront, 
it scores only 27.6\% team accuracy on \name after 5 turns of conversation.
No model exceeds 56\% on any domain in \name, 
and all exhibit gaps between standalone agent performance and team accuracy.
An error analysis reveals two failure modes. 
The first is \textit{poor report quality}: agents omit or misrepresent item features, leaving users unable to evaluate recommendations even when the right item is present. The second is \textit{underdeveloped user preferences}: agents fail to take the dialog actions needed to expand what users know, leaving them unable to recognize the correct item.

In summary, our contributions are:
\begin{itemize}
    \item \frameworkname, a model of preference construction in which different features --- depending on how they're learned --- demand different agent dialog actions.
    \item \name, an interactive benchmark of agentic recommender systems, which is open-sourced at \href{https://irenasaracay.github.io/coshop/}{this URL}.
\end{itemize}
\section{\frameworkname : Modeling users with developing preference states}\label{sec:user}
We study agents that converse with a user over a bounded number of turns before proposing a solution. Formally, an agent interacts with a user for $T \le B_\text{turns}$ turns,\footnote{We count a turn as one user message followed by an agent message.} then returns a solution $x$ from a solution space $\mathcal{X}$ (e.g., a set of hiking boots). Solutions vary along a set of features $\mathcal{F}$ that users may care about, such as price or lug depth; each solution's values on these features are given by a feature map $\phi: \mathcal{X} \to \mathbb{R}^{|\mathcal{F}|}$.

We assume the user has some latent target item $x^\ast$ in mind, and that their preferences are determined by its feature values: for each feature $f \in \mathcal{F}$, the \textbf{preferred value} is $\phi_f(x^*)$, i.e. the value the target takes on that feature. For instance, $\phi_\text{price}(x^*) = ``{\le}\$50''$ is a budget preference: the user wants items priced at or under \$50. Critically, however, a non-expert user need not be aware of --- or hold an opinion on --- every feature in $\mathcal{F}$ at once. We capture this with a \textbf{preference state} $S_t \subseteq \mathcal{F}$ the subset of features over which the user has actually formed a preference by turn $t$. This state determines what the user can ultimately evaluate: at the end of the conversation, the user judges a proposed solution $\hat x$ based on how well it matches their preferences over the features in their final state $S_{T+1}$.

\paragraph{Beyond expert users: modeling preference construction.}
Prior work models users as \textit{experts} ---
their preference state at every turn includes all features, i.e. $S_t = \mathcal F$ for all $t$.
This means the user can answer any question about any feature $f$ with $\phi_f(x^*)$ perfectly (Figure \ref{fig:hero} bottom left).
Not only is this assumption unrealistic,
it rewards undesirable agent behavior:
since the user can answer any question,
an agent's best strategy is to bombard users with questions upfront.
We observe exactly this in Section \ref{sec:analysis},
where models like \gptfive ask a median of 9 clarifying questions.
The social science literature offers a richer model:
\citet{bettman1998constructive} and \citet{fischhoff1991value} argue that preferences
are not pre-formed and waiting to be retrieved --- they are \textit{constructed}
through interaction, shaped by what the user encounters and learns from the agent.
\citet{knijnenburg2010effect,wang2016effects,kostric2021soliciting,shen2024multi}
further document that users lack confident, fully-formed preferences,
particularly in domains where they have limited expertise.
The agent's role is thus not just to query a static preference state,
but to actively participate in constructing it.

We operationalize this by modeling preference states as dependent on agent actions. $S_1$ is initialized to be sparse, matching empirical studies suggesting consumers start with few predefined preferences \citep{song2021purchase}. At turn $t$, the user updates their preference state based on the previous agent message (Figure \ref{fig:hero} bottom right).

\begin{figure}[bt]
    \centering
    \includegraphics[width=0.95\textwidth]{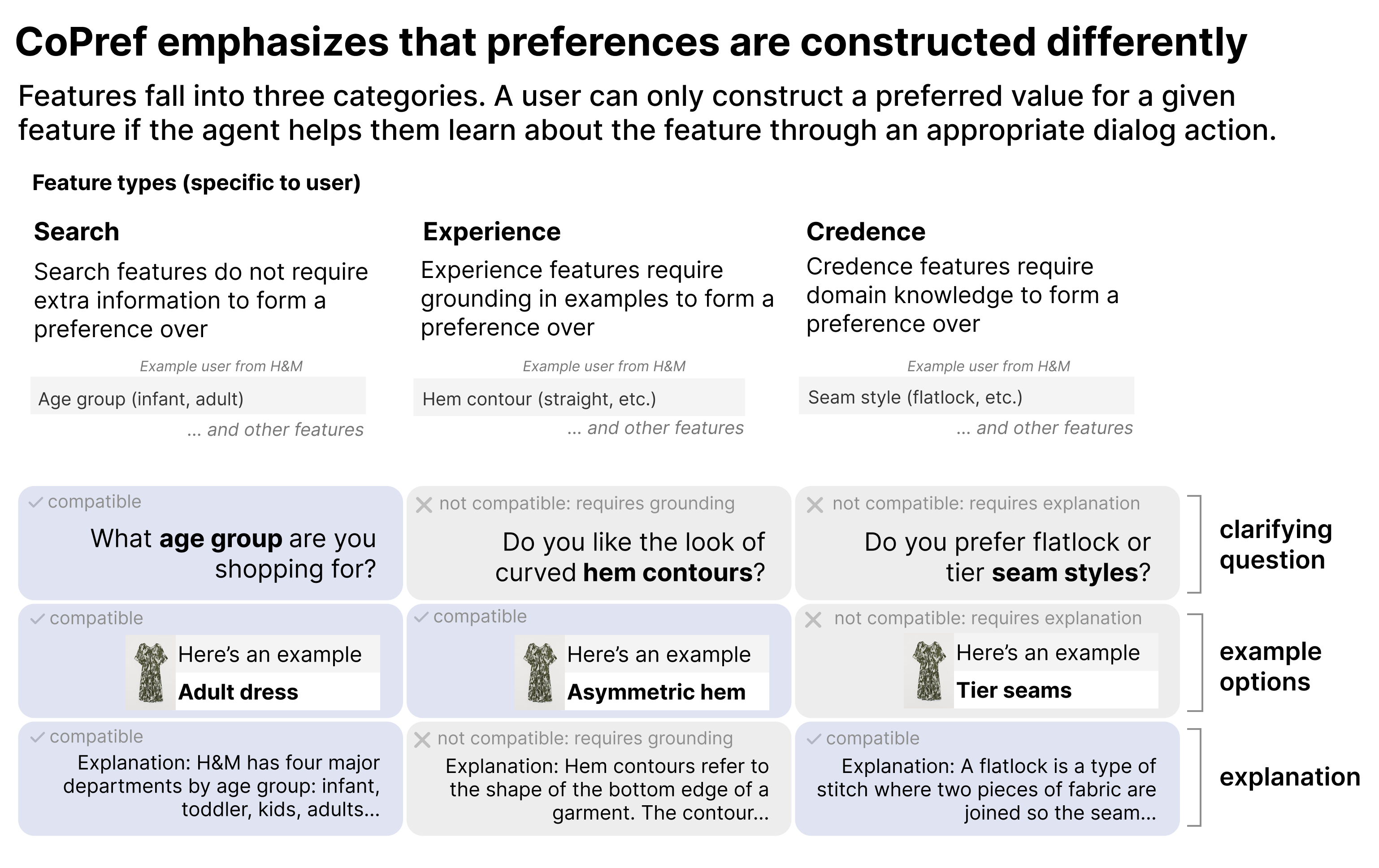}
    \caption{Agent dialog actions determine which user preferences are constructed during conversation.
We adopt the \textbf{Search-Experience-Credence} (SEC) framework from Information Economics,
which partitions item features by the information a user needs to form a preference over them. 
An agent's turn may contain three types of dialog actions (clarifying questions, example options, and technical explanations), which unlock features if they match an appropriate SEC type.
A strong interaction thus shapes what the user learns about themselves.
}
    \label{fig:problem}
\end{figure}

\paragraph{Not all preferences are equally easy to construct.}
Information Economics offers a relevant distinction: not all product attributes require the same kind of information before a consumer can form a judgment about them. The \textbf{Search-Experience-Credence (SEC)} framework \citep{nelson1970information, darby1973free, ford1990consumer} formalizes this by classifying features according to what a consumer needs in order to assess them, and empirical work testing the framework \citep{girard2010validating} confirms that consumer behavior really does differ systematically across these types. 
We adopt this typology and partition $\mathcal{F} = \mathcal{F}_\text{s} \cup \mathcal{F}_\text{e} \cup \mathcal{F}_\text{c}$ accordingly. 
\begin{enumerate}
\item \textbf{Search features} $\mathcal{F}_\text{s}$ are easy for the user to form preferences over;
    they do not require additional information for the user to assess what they prefer (e.g., price, size).
\item \textbf{Experience features} $\mathcal{F}_\text{e}$ require grounding in exemplar solutions $x \in \mathcal X$
    for a user to decide what their preferred value is. A user knows what they like or don't like when they see it,
    but not beforehand.
\item \textbf{Credence features} $\mathcal{F}_\text{c}$ require domain knowledge to form a preference for. For example, 
    a preference for viscose vs. lyocell fabric requires the user to do additional research to understand.
\end{enumerate}

Each SEC type suggests a different way an agent can help a user construct a preference over it (Figure \ref{fig:problem}). A \textbf{clarifying question} is enough to surface a search feature, since the user already knows their preference and just needs to be asked. \textbf{Example options} can ground an experience feature --- and, trivially, a search feature too --- by letting the user react to something concrete rather than an abstract attribute. A \textbf{technical explanation} can unlock a credence feature --- and, again, a search feature --- by supplying the domain knowledge the user was missing.
Concretely, a feature is added to the user's preference state $S_t$ once the agent's turn-$t$ message references it through an action capable of unlocking its type.

In the classical SEC literature, the search, experience, and credence partitions are fixed across all users. However, we hypothesize that SEC splits should vary by the individual; two users may bring different domain knowledge to the same feature.

\begin{figure}[bt]
    \centering
    \includegraphics[width=\textwidth]{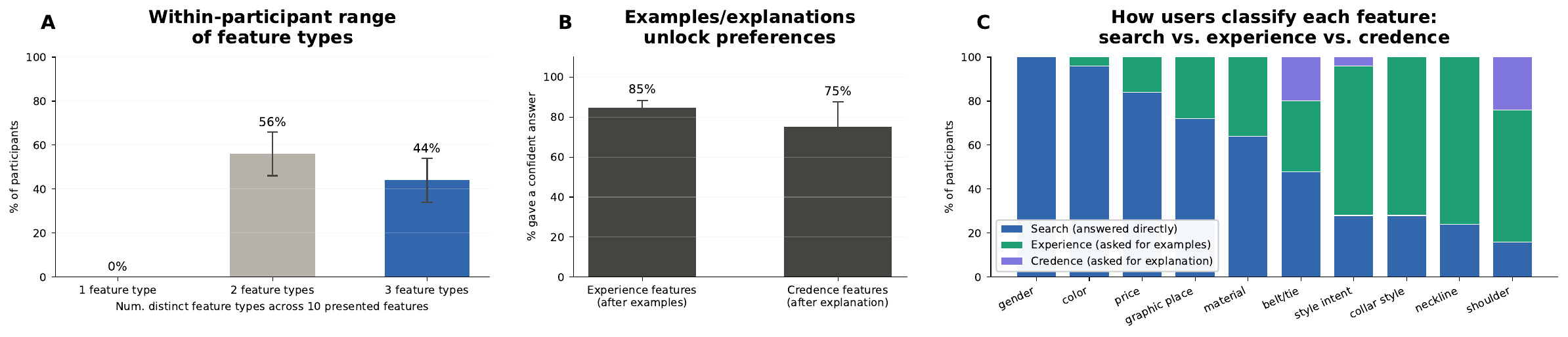}
    \caption{Our human study validates that (A) users do have a mix of search, experience, and credence features; (B) preferences can be “unlocked” for experience / credence features if shown the appropriate dialog action; and (C) there is individual variation in how search, experience, and credence features are split across users.
    }
    \label{fig:human_study}
\end{figure}

\subsection{Human study: validating \frameworkname}\label{sec:human_study}
\frameworkname rests on three claims: that non-expert users have a mix of search, experience, and credence features, rather than just search features; that the dialog actions we propose --- questions, examples, explanations --- succeed in unlocking preferences over their corresponding feature type; and that a feature's SEC type may be a property of the individual rather than fixed across users. We validated these with a human study.

We recruited $n=25$  participants for a simulated clothing shopping task. Participants imagined shopping for a sweater with an AI agent that needed to understand their preferences before recommending items. Participants self-reported a range of expertises with sweaters (28\% claimed to be very familiar with sweaters). To elicit each participant's SEC split, we administered 10 rounds, one per sweater feature drawn from the \taskhm dataset in Section \ref{sec:benchmark} (e.g., price, neckline, material). In each round, participants saw a clarifying question about the feature and could either answer directly, in which case we classified the feature as a search feature for that participant; or indicate uncertainty by requesting examples or a technical explanation, in which case we classified the feature as an experience or credence feature, respectively. Those requesting examples or explanations were then shown the corresponding content and asked again to state a preference. Additional details are in Appendix \ref{app:human_study}.

\paragraph{Non-expert users have heterogeneous preferences, and dialog actions unlock them.}
Our results validate the first two claims. 
Individuals have a mix of search, experience, and credence features: 0 participants had all search features, as every participant asked for examples and/or an explanation on at least one feature; 56\% asked only for examples or only for explanations, suggesting they had search-and-experience or search-and-credence features among the presented set, while the remaining 44\% had all three kinds (Figure \ref{fig:human_study}). 
Appropriate dialog actions also do unlock preferences for experience and credence features: of (participant, feature) pairs where the participant requested examples after indicating uncertainty, 85\% were subsequently "unlocked" (i.e., the participant went on to provide a non-empty free-text preference); of pairs where the participant requested an explanation, 75\% were unlocked.

\paragraph{SEC type varies across individuals.}
Aside from gender, no feature was uniformly classified as search, experience, or credence across participants. At the same time, credence classifications were sparse and concentrated in a small number of features (shoulder style, belt/tie presence), suggesting that while credence features are not universal, they cluster around a handful of technically demanding attributes. 
This suggests that features carry some base rate of difficulty, but their SEC classifications vary across individuals based on individual domain knowledge.

\subsection{Limitations of \frameworkname} 
\frameworkname makes two simplifying assumptions about preference construction. First, users form a preference over a feature based on a single agent action, rather than requiring multiple interactions. Second, once formed, preferences are stable and never forgotten: we assume that $S_t$ is monotonically non-decreasing. Real preference construction may be more iterative and less stable than this model assumes — a user might need to see examples before realizing they need a technical explanation, or might revise a preference they had previously settled on. 
These simplifications make \frameworkname users easier to collaborate with --- however, we find in Section \ref{sec:results} that frontier language models still fail to help users develop preferences, even under these favorable simplifications.

\section{\name: An interactive benchmark for agentic recommender systems}\label{sec:benchmark}
As a case study of how well agents work with \frameworkname users, we instantiate an interactive benchmark for agentic recommender systems. 
Shopping is a natural fit for our problem setting:
product attributes like price and color give us a well-defined feature set $\mathcal{F}$, and a first-time buyer rarely knows all features that will matter to them until they review real options while shopping. 

\begin{figure}[tb]
    \centering
    
    \begin{minipage}[c]{0.27\linewidth}
        \centering
        \includegraphics[width=\linewidth]{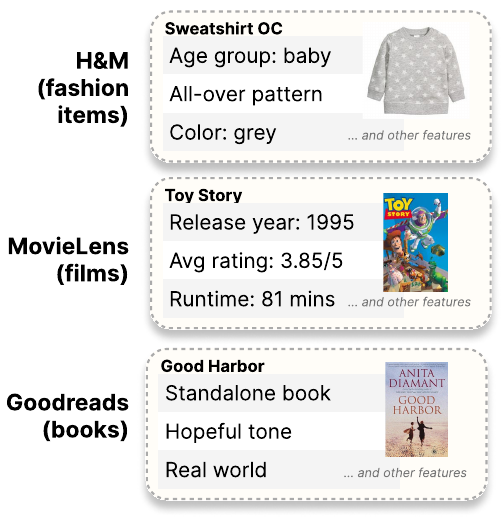}
        \label{fig:dataset_examples}
    \end{minipage}
    \hfill
    \begin{minipage}[c]{0.7\linewidth}
        \centering
        \resizebox{\linewidth}{!}{
            \begin{tabular}{lccc}
    \toprule
    & \textbf{\taskhm} & \textbf{\movielens} & \textbf{\goodreads} \\
    \midrule
    Number of users & 100 & 100 & 100 \\
    Catalog size $|\mathcal X|$ & \numprint{37570} & \numprint{26637} & \numprint{39050} \\
    Number of domain features $|\mathcal F|$ & 77 & 92 & 102 \\
    Initial state size $|S_{1}|$ & 3.92 (6.05) & 1.67 (1.20) & 3.76 (11.08) \\
    Max state size $|\mathcal F(x^*)|$ & 36.69 (9.07) & 87.59 (2.14) & 88.24 (8.20) \\
    Number of search features $|\mathcal{F}_{\text{s}}|$ & 6.11 (6.84) & 8.60 (16.74) & 7.76 (11.92)\\
    Number of experience features $|\mathcal{F}_{\text{e}}|$ & 21.47 (7.33) & 70.20 (12.72) & 66.98 (65.22) \\
    Number of credence features $|\mathcal{F}_{\text{c}}|$ & 9.11 (7.42) & 8.79 (2.35) & 13.50 (70.79) \\
    \bottomrule
    \end{tabular}

        }
    \end{minipage}
    \captionof{table}{\name covers three product domains: 
    fashion recommendations from the \taskhm dataset \citep{hm}, 
    movie recommendations from \movielens \citep{movielens},
    and books from \goodreads \citep{goodreads1, goodreads2}. Numbers reported as mean (standard deviation).}
    \label{tab:datasets}
\end{figure}

\paragraph{Problem definition.}
In recommendation, the solution space $\mathcal{X}$ is a catalog of items, the target $x^\ast$ is the item the user has in mind, and the features $\mathcal{F}$ are catalog attributes such as price or material. 
Not all features apply to all items; we use $\mathcal F(x)$ to denote the features relevant to $x$ (e.g., sleeve length isn't a preference one can hold about shoes).
As in Section~\ref{sec:user}, the preference state $S_t \subseteq \mathcal{F}(x^*)$ tracks which features the user has formed an opinion on by turn $t$. 
Each rollout proceeds in four stages (Figure \ref{fig:hero}, top):
\begin{enumerate}
    \item \textbf{Conversation.} The agent converses with the user for up to $B_\text{turns} = 5$ turns or until the agent emits a special ``end conversation'' token. The agent can call a catalog search tool, which takes a query string $q$ and returns the $m$ most relevant items with features $\phi(x_1), \dots, \phi(x_m)$. This stage is limited to $B_\text{preview} = 50$ retrievals, 20 clarifying questions, and 10 unique items shown to the user, based on prior estimates of how many questions users tolerate before dropping out of elicitation \citep{zou2020empirical} (we provide a sensitivity analysis in Appendix \ref{app:budget_sensitivity}).
    \item \textbf{Research.} The agent searches more thoroughly, with an expanded budget of $B_\text{search} = 250$ retrievals.
    \item \textbf{Report.} The agent writes up its final $k = 5$ recommendations.
    \item \textbf{Selection.} The user picks one item $\hat x$ from the report, evaluating it against their final preference state $S_{T+1}$.
\end{enumerate}

Performance is measured by the user-agent \textit{team's} accuracy: whether $\hat x$ matches $x^\ast$. Evaluating by team accuracy stresses the importance of preference construction: without a preference state $S_{T+1}$ rich enough to distinguish the target from its $k-1$ distractors, the user may be unable to break ties well.

\paragraph{Product domains.}
We source \name data from three existing datasets of real human-product interactions  (Table \ref{tab:datasets}): 
fashion purchases from the \taskhm dataset \citep{hm}, 
movie recommendations from \movielens \citep{movielens},
and books from \goodreads \citep{goodreads1, goodreads2}.
Each dataset provides a product catalog $\mathcal{X}$ with domain-specific columns  $\mathcal{F}$; to increase the size of $\mathcal F$, we enriched the original datasets' columns using a mix of regex and LLM-based methods (Appendix~\ref{app:data}).

\begin{table}[bht]
    \centering
    \fbox{%
        \begin{minipage}{\dimexpr0.9\textwidth-2\fboxsep-2\fboxrule\relax}

{\setlength{\parindent}{0pt}\footnotesize
\textbf{\taskhm}\par\smallskip
\hangindent=2em\hangafter=1 \textbf{Search:} Product category \textcolor{gray}{(100\%)}, Gender \textcolor{gray}{(100\%)}, Age group \textcolor{gray}{(100\%)}, Neckline or collar \textcolor{gray}{(27\%)}, Color \textcolor{gray}{(26\%)}\par
\hangindent=2em\hangafter=1 \textbf{Experience:} Windproof or wind-resistant \textcolor{gray}{(100\%)}, Waterproof or water-resistant \textcolor{gray}{(100\%)}, Has sequins or glitter embellishment \textcolor{gray}{(100\%)}, Has detachable hood \textcolor{gray}{(100\%)}, Has text or slogan graphic \textcolor{gray}{(100\%)}\par
\hangindent=2em\hangafter=1 \textbf{Credence:} Vegan \textcolor{gray}{(100\%)}, Toxin-free certification \textcolor{gray}{(100\%)}, Woven fabric \textcolor{gray}{(100\%)}, Sustainable \textcolor{gray}{(100\%)}, Has yoke panel \textcolor{gray}{(100\%)}\par
}\par\medskip

{\setlength{\parindent}{0pt}\footnotesize
\textbf{\movielens}\par\smallskip
\hangindent=2em\hangafter=1 \textbf{Search:} Genres of the movie \textcolor{gray}{(100\%)}, Popularity score of the movie \textcolor{gray}{(59\%)}, Whether the film has been released on a physical home media format \textcolor{gray}{(59\%)}, Average rating of the movie out of 5 \textcolor{gray}{(55\%)}, Number of ratings for the movie \textcolor{gray}{(53\%)}\par
\hangindent=2em\hangafter=1 \textbf{Experience:} Whether the film features unrequited love \textcolor{gray}{(100\%)}, Whether the film centers on an unlikely or unexpected friendship \textcolor{gray}{(100\%)}, Whether the film is a slow burn — building tension or emotion gradually over time \textcolor{gray}{(100\%)}, Whether the film features a mentor-protege relationship \textcolor{gray}{(100\%)}, Whether the film features martial arts \textcolor{gray}{(100\%)}\par
\hangindent=2em\hangafter=1 \textbf{Credence:} Production companies of the movie \textcolor{gray}{(100\%)}, Whether the film has been selected for the US National Film Registry \textcolor{gray}{(100\%)}, Whether the film appears on an IMDb ranked list such as the IMDb Top 250 \textcolor{gray}{(100\%)}, Collections the movie was featured in \textcolor{gray}{(100\%)}, Whether the film passes or fails the Bechdel Test \textcolor{gray}{(100\%)}\par
}\par\medskip

{\setlength{\parindent}{0pt}\footnotesize
\textbf{\goodreads}\par\smallskip
\hangindent=2em\hangafter=1 \textbf{Search:} Genres of the book \textcolor{gray}{(100\%)}, Primary emotional register \textcolor{gray}{(39\%)}, audiobook \textcolor{gray}{(36\%)}, Primary intended readership age group \textcolor{gray}{(33\%)}, Number of books in the series \textcolor{gray}{(32\%)}\par
\hangindent=2em\hangafter=1 \textbf{Experience:} Trope: secret baby \textcolor{gray}{(99\%)}, Trope: rags to riches \textcolor{gray}{(99\%)}, Trope: mistaken or swapped identity \textcolor{gray}{(99\%)}, Trope: found family \textcolor{gray}{(99\%)}, Trope: forced proximity \textcolor{gray}{(99\%)}\par
\hangindent=2em\hangafter=1 \textbf{Credence:} Whether the book features an explicit, rule-based magic system \textcolor{gray}{(100\%)}, Whether the book is tagged as a book club selection \textcolor{gray}{(100\%)}, Author name \textcolor{gray}{(100\%)}, Number of ratings for the book \textcolor{gray}{(79\%)}, Technological era of the story's setting \textcolor{gray}{(77\%)}\par
}\par\medskip

        \end{minipage}}
    \caption{The five most common search, experience, and credence features in each dataset, ranked by the percentage of the 100 test users for whom the feature receives that classification (shown in parentheses).}
    \label{tab:sample_sec}
\end{table}

\paragraph{Constructing test users.} 
For each domain, agents are evaluated in interactions with 100 test users.
Each user is anchored to a real human from the original dataset.
The user's target item $x^\ast$ is the last item that the user purchased or rated at least 4 out of 5 stars.
Motivated by our human study results in Section \ref{sec:human_study}, each user's SEC split is personalized based on their purchase history (Appendix~\ref{app:data_sec}).
Concretely, for each feature $f$ and target value $v = \phi(x^*)_f$, we compute the pointwise mutual information between the catalog marginal $p_\text{catalog}(f{=}v)$ and the user's empirical purchase frequency $p_\text{user}(f{=}v)$.
Features the user consistently over-selects relative to the catalog are classified as \emph{search}; features selected at roughly the catalog rate as \emph{experience}; and features that are either rare in the catalog or the user under-selects, as \emph{credence}.
The most common search, experience, and credence features for each dataset are given in Table \ref{tab:sample_sec}.
Each user's initial state $S_{1}$ is a subset of their search features $\mathcal F_S$. We construct $S_1$ so that it is one feature short of identifying $x^\ast$: we build $S_{1}$ for each user by randomly adding features from the user's search split, stopping just before \oss can recall $x^\ast$ directly from the research stage at a budget of 250. This is to ensure challenging, but not overly adversarial initializations.

Unlike typical LLM-based user simulators, which prompt a model once with a persona and let it improvise an entire conversation, 
we explicitly anchor our simulator LLM \oss to the current preference state $S_t$. 
Concretely, we generate natural language messages conditioned only on the current preference state $S_t$, and we have \oss select a final item conditioned only on the final preference state $S_{T+1}$ and the agent's report (with item order randomized to prevent position bias).

\paragraph{Parsing dialog actions from agent messages.}
In order to study agents in as natural a setting as possible, agents emit natural language messages. 
To compute state updates, we used \oss to parse the agent's message into dialog actions and, critically, identify the features being referenced by each parsed action. 
This parsing step introduces some error, mainly around identifying the referenced features:
in Appendix~\ref{app:simulator_evaluation}, 
we found that \oss has question parsing accuracy of 74.0\% and item parsing accuracy of 94.0\%.
In Appendix \ref{app:structured}, we also experimented with removing this parsing step: we required agents to emit structured dialog actions during the conversation stage (\texttt{ask\_question}, \texttt{show\_item\_for\_feedback}, or \texttt{explain}), each of which must explicitly reference the target feature name.
This removes the confound of parsing error at the cost of realism, and it may change the distribution of what agents say.
We found that removing the parsing step dramatically degraded performance for tested models, which tended to hallucinate column names (e.g., \texttt{color} instead of \texttt{perceived\_colour\_master\_name}), even when the model's natural language output was coherent.

\section{Benchmark results on \name}\label{sec:results}
We evaluated five frontier language models on \name: 
one smaller and one larger model from the GPT and Claude families 
(\gptmini, \gptfive, \claudehaiku, \claudesonnet),
as well as one open-source model \oss. 
We use a simple RAG agent harness which gives the model access to a catalog search tool, a private scratchpad, a reminder to be cognizant of the user's limited attention,
and a summary of the SEC theory along with the various budgets $B_\text{turn}, B_\text{preview}, B_\text{search}$ 
(prompts in Appendix \ref{app:baselines}).
We also compared agents to non-LLM baselines which retrieve 500 items using a query based on $S_1$ and then rerank either randomly or by catalog popularity.

\begin{figure}[tb]
    \centering
    \includegraphics[width=\textwidth]{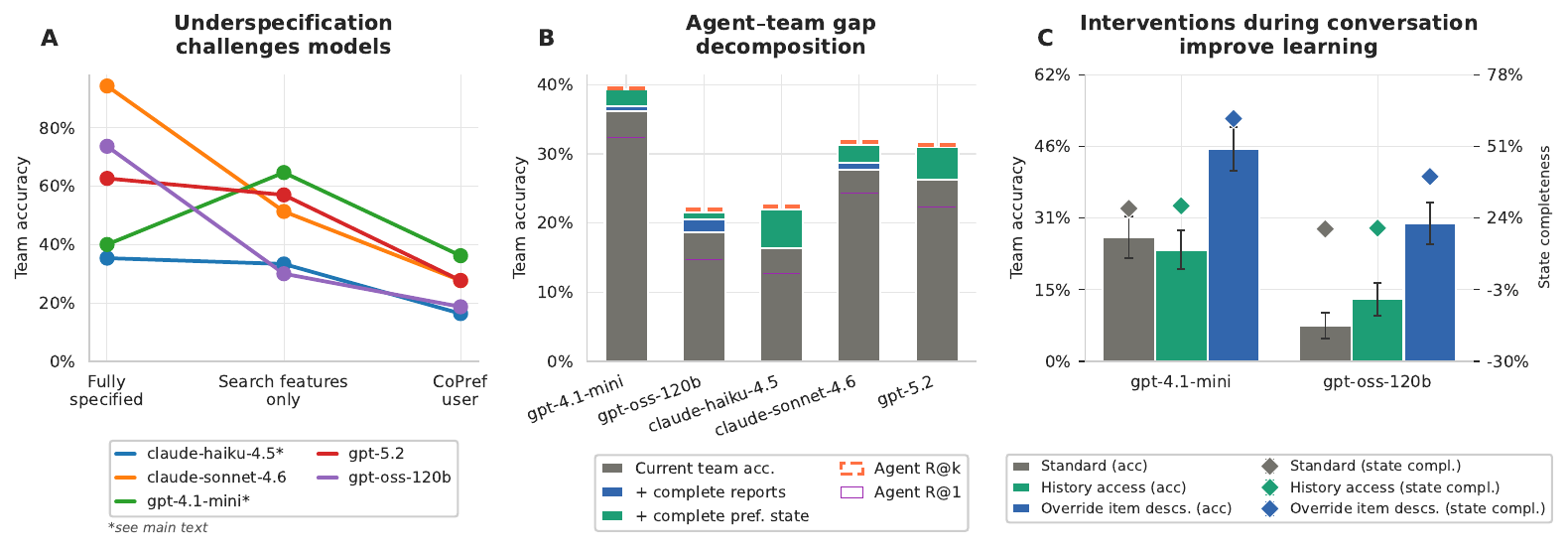}
    \caption{%
    \textbf{(A) Underspecification causes large, unrecovered performance drops.}
    All models score highly with fully-specified preferences,
    but no model recovers this through conversation under either interactive setting.
    \textbf{(B) The agent--team gap stems from report quality and underdeveloped preference states.}
    Complete reports (blue) and full preference states (green) recover most of the gap,
    with a fully-informed user exceeding agent recall@$1$ for all models.
    \textbf{(C) Effect of interventions.}
    Providing agents with user rating history yields minimal gains,
    while injecting full catalog descriptions for items mentioned during conversation
    substantially improves state completeness and team accuracy.
}
    \label{fig:underspecification_curve}
\end{figure}

\subsection{Agents excel at search but fail to resolve underspecification}
We begin by isolating agents' task execution ability ---
how well can agents find and recommend items given the user's full preference set $\phi(x^*)$ upfront?
In Figure \ref{fig:underspecification_curve}A,
the leftmost \textit{fully-specified} setting shows that models score highly in this fully-specified setting, with
\claudesonnet scoring an average of 94.3\% across datasets.
There are two exceptions to this rule.
\claudehaiku scored 98\% on \taskhm, but on both \movielens and \goodreads hit context limit errors on 96 of 100 runs.
\gptmini scored 86\% on \taskhm but only 17\% on \movielens and \goodreads;
on these datasets, in nearly half of runs it failed to terminate after exceeding the research budget, instead issuing tool calls indefinitely (Appendix Table \ref{tab:full_knowledge_by_dataset}).

Next, we ask how performance changes as users become more underspecified.
In the \textit{search features only} setting, 
users develop preferences over the course of the conversation as in \frameworkname,
but all features are search features ($\mathcal F_\text{s} = \mathcal F$).
This means that any dialog action can add any feature to the preference state;
preferences are simply \textit{retrieved}, and clarifying questions can recover all preferences.
This mimics the setups of \citet{murakhovs2023salespeople,chen2024chatshop,kim2026discoverllm}.
Finally, we use \frameworkname users from Section \ref{sec:user}, 
where features vary in construction difficulty depending on the user's SEC split.

Figure \ref{fig:underspecification_curve}A traces final accuracy across these three user types.
All models drop in performance when non-search features are introduced.
For example, while \gptfive scores 57.0\% accuracy when all features are search features,
it scores only 27.7\% accuracy when some features require actions besides clarifying questions.

\subsection{Team accuracy reveals human-facing failures}

We now report detailed results on \name using three families of metrics,
each isolating a different component of the pipeline.
\textit{State completeness} measures the fraction of the user's full preference set recovered by end of conversation:
$|S_{T+1}| / |\mathcal{F}(x^*)|$.
\textit{Agent recall} measures whether $x^\ast$ appears in the agent's top-$k$ recommendations after research.
\textit{Team accuracy} --- our primary metric --- measures whether the user ultimately selects $x^\ast$
from the agent's report, and is upper bounded by agent recall at $k$.
Table \ref{tab:main_results} (RAG Agents section) reveals two things:
first, no model recovers more than 54\% of the user's preferences over the course of the conversation. 
This means agents are leaving a substantial portion of users' preferences undeveloped.
Second, team accuracy is consistently and substantially lower than agent recall:
even when agents surface $x^\ast$ in their top $k$, users often cannot identify it as better than the other $k-1$ presented options.
For example, \gptfive scores 65\% recall at $k$ on \taskhm but only 56\% team accuracy, a 9-point gap.

\begin{table}[tb]
    \centering
    \resizebox{\linewidth}{!}{
        \begin{tabular}{cccccccccc}
\toprule
\multicolumn{1}{c}{} & \multicolumn{3}{c}{\taskhm} & \multicolumn{3}{c}{\movielens} & \multicolumn{3}{c}{\goodreads} \\
\cmidrule(lr){2-4} \cmidrule(lr){5-7} \cmidrule(lr){8-10}
\multicolumn{1}{c}{} & \multicolumn{1}{c}{User} & \multicolumn{1}{c}{Agent} & \multicolumn{1}{c}{Team} & \multicolumn{1}{c}{User} & \multicolumn{1}{c}{Agent} & \multicolumn{1}{c}{Team} & \multicolumn{1}{c}{User} & \multicolumn{1}{c}{Agent} & \multicolumn{1}{c}{Team} \\
\cmidrule(lr){2-2} \cmidrule(lr){3-3} \cmidrule(lr){4-4} \cmidrule(lr){5-5} \cmidrule(lr){6-6} \cmidrule(lr){7-7} \cmidrule(lr){8-8} \cmidrule(lr){9-9} \cmidrule(lr){10-10}
 & State compl. & R@k & Acc. & State compl. & R@k & Acc. & State compl. & R@k & Acc. \\
\midrule
\midrule
\multicolumn{10}{c}{\textit{Baselines}} \\
\midrule
Random item & --- & 0.01 (0.00) & 0.01 (0.00) & --- & 0.01 (0.00) & 0.01 (0.00) & --- & 0.00 (0.00) & 0.00 (0.00) \\
Popularity & --- & 0.00 (0.00) & 0.00 (0.00) & --- & 0.00 (0.00) & 0.00 (0.00) & --- & 0.00 (0.00) & 0.00 (0.00) \\
\midrule
\multicolumn{10}{c}{\textit{RAG Agents}} \\
\midrule
\oss & 0.48 (0.01) & 0.32 (0.05) & 0.26 (0.04) & 0.21 (0.01) & 0.24 (0.04) & 0.23 (0.04) & 0.19 (0.01) & 0.10 (0.03) & 0.08 (0.03) \\
\claudehaiku & 0.31 (0.01) & 0.44 (0.05) & 0.33 (0.05) & 0.19 (0.01) & 0.20 (0.04) & 0.15 (0.04) & 0.18 (0.01) & 0.03 (0.02) & 0.01 (0.01) \\
\claudesonnet & 0.36 (0.01) & 0.45 (0.05) & 0.37 (0.05) & 0.24 (0.01) & 0.36 (0.05) & 0.35 (0.05) & 0.25 (0.01) & 0.14 (0.04) & 0.11 (0.03) \\
\gptmini & \textbf{0.54 (0.01)} & 0.49 (0.05) & 0.42 (0.05) & \textbf{0.40 (0.01)} & \textbf{0.48 (0.05)} & \textbf{0.46 (0.05)} & \textbf{0.38 (0.02)} & \textbf{0.21 (0.04)} & \textbf{0.20 (0.04)} \\
\gptfive & 0.46 (0.01) & \textbf{0.65 (0.05)} & \textbf{0.56 (0.05)} & 0.28 (0.01) & 0.23 (0.04) & 0.21 (0.04) & 0.24 (0.01) & 0.06 (0.02) & 0.06 (0.02) \\
\midrule
\multicolumn{10}{c}{\textit{Intervention: user history access}} \\
\midrule
\oss & 0.31 (0.01) & 0.31 (0.05) & 0.18 (0.04) & 0.16 (0.01) & 0.18 (0.04) & 0.16 (0.04) & 0.13 (0.01) & 0.07 (0.03) & 0.06 (0.02) \\
\gptmini & 0.37 (0.01) & 0.40 (0.05) & 0.34 (0.05) & 0.25 (0.01) & 0.24 (0.04) & 0.24 (0.04) & 0.23 (0.01) & 0.17 (0.04) & 0.14 (0.03) \\
\midrule
\multicolumn{10}{c}{\textit{Intervention: override item descriptions}} \\
\midrule
\oss & 0.68 (0.01) & 0.44 (0.05) & 0.40 (0.05) & 0.25 (0.01) & 0.38 (0.05) & 0.37 (0.05) & 0.25 (0.01) & 0.13 (0.03) & 0.12 (0.03) \\
\gptmini & 0.73 (0.01) & 0.56 (0.05) & 0.48 (0.05) & 0.57 (0.02) & 0.64 (0.05) & 0.63 (0.05) & 0.53 (0.02) & 0.26 (0.04) & 0.26 (0.04) \\
\bottomrule
\end{tabular}
    }
    \captionof{table}{Main results on \name across three metric families:
    state completeness (fraction of user preferences developed during conversation),
    agent recall (whether $x^\ast$ appears in the agent's recommendations),
    and team accuracy (whether the user ultimately selects $x^\ast$).
    Additional results in Appendix~\ref{app:additional_results}.\vspace{-1em}}
    \label{tab:main_results}
\end{table}

We isolate the sources of this gap experimentally in Figure \ref{fig:underspecification_curve}B.
The first source is \textit{poor report quality}.
Since users select $\hat x$ based solely on the agent's generated report,
if an agent misrepresents an item (e.g., omits or hallucinates features),
or if their report is strongly biased towards a particular item,
this directly impairs team accuracy.
To measure the extent to which this happens, we replayed the user selection step with complete reports ---
programmatically generated descriptions of $\phi(x)$, which includes all features in $\mathcal F$ ---
in place of the agent's report.
Team accuracy increases significantly for \oss, and slightly for \gptmini and \claudesonnet (blue bars).

The second source for the team-agent gap is \textit{underdeveloped preference states}.
Even with complete reports, users sometimes cannot distinguish between items
because the final state $S_{T+1}$ lacks the discriminative features needed to break apparent ties ---
a direct consequence of failing to develop the user's preference state during conversation.
Replaying selection with both complete reports \textit{and} the full preference set $\phi(x^*)$
improves accuracy further (green bars).
For all models outside of \oss, underdeveloped preference states explain most of the team-agent gap.
Note that underdeveloped preference states are also responsible for low agent recall@$k$ in the first place. Agents that fail to grow the user's preference state leave the search task greatly underspecified.

Further, with complete reports and full preference states,
team accuracy substantially exceeds agent recall@$1$ for all models (Figure \ref{fig:underspecification_curve}B).
This shows that a well-informed user is a better selector than the original agent ---
teams with well-informed users can outperform the agent alone.

\subsection{Interventions on agent knowledge and communication}\label{sec:intervention_results}
The agents above interact with users zero-shot, with no knowledge of their prior rating history. In principle, history access could help in two ways: since our SEC splits are derived from purchase history, models might better anticipate whether a given feature is likely to be search, experience, or credence for a particular user, and might also infer preferences directly from past behavior without needing to ask. Table \ref{tab:main_results} (history access) and Figure \ref{fig:underspecification_curve}C show that for the models we tested, neither benefit materializes significantly.

A more direct lever is how agents communicate during conversation. Most models show example items frequently, but describe them incompletely --- rather than proactively surfacing features the user may not have considered, agents tend to rehash already-known ones, missing the key opportunity to grow $S_t$ that examples afford. In Table \ref{tab:main_results} (override item descriptions) and Figure~\ref{fig:underspecification_curve}C, we overrode agent messages during conversation crudely, replacing their item descriptions with programmatic ones that mention all catalog features. We found substantial gains: \gptmini's state completeness jumps from 44\% to 61\% and team accuracy from 36\% to 46\%. This suggests that improving how agents use dialog actions --- particularly examples and explanations --- is a tractable path to better performance with \frameworkname users, even holding model capability fixed.
\subsection{Analysis of interaction strategies on \name}\label{sec:analysis}
To understand what drives underdeveloped preference states, 
we turn to Table~\ref{tab:budgets}, which reveals that models cluster into qualitatively different interaction strategies.

\begin{table}[tb]
    \centering
    \resizebox{\linewidth}{!}{
        \begin{tabular}{ccccccccc}
\toprule
\multicolumn{1}{c}{} & \multicolumn{5}{c}{Budget util. (median)} & \multicolumn{3}{c}{Behavior freq. (avg)} \\
\cmidrule(lr){2-6} \cmidrule(lr){7-9}
 & Turns & Qs & Examples & Explanations & Research & Confirmation Qs & Loops & Hallucinations \\
\midrule
\claudehaiku & 5.00 & 9.50 & 6.33 & 0.00 & 96.00 & 0.10 (0.01) & 0.24 (0.04) & 0.47 (0.05) \\
\claudesonnet & 3.67 & 8.17 & 5.33 & 0.00 & 106.50 & 0.10 (0.01) & 0.15 (0.04) & 0.00 (0.00) \\
\gptmini & 5.00 & 7.58 & 7.75 & 0.00 & 40.00 & 0.07 (0.01) & 0.35 (0.04) & 0.01 (0.01) \\
\gptfive & 2.67 & 9.00 & 6.00 & 0.00 & 190.33 & 0.20 (0.01) & 0.03 (0.01) & 0.00 (0.00) \\
\oss & 5.00 & 6.50 & 4.67 & 0.00 & 52.25 & 0.06 (0.01) & 0.20 (0.04) & 0.31 (0.05) \\
\bottomrule
\end{tabular}
    }
    \captionof{table}{Models adopt qualitatively different interaction strategies on \name,
    varying in question volume, turn budget utilization, and item description completeness.\vspace{-1em}}    \label{tab:budgets}
\end{table}

\paragraph{Conversation vs. research.} We observe two groups of models when looking at the balance of the conversation and research stages (Table \ref{tab:budgets}).
Smaller models (\oss, \claudehaiku, \gptmini) exhaust the conversation turn budget (median 5.00 turns) while using relatively little of the research budget.
The larger models (\claudesonnet, \gptfive) prematurely end the conversation, but use more of the 250 item research budget.
In 37\% of conversations, \gptfive ends the conversation after two turns.
This suggests an ``execution bias'' for the larger models --- these are overly cautious about using user time and try to do more work offline, which fails on \name because the task remains underspecified. 

\paragraph{Distribution of dialog actions.} All models ask questions and show example items, but no model proactively provides explanations for technical features. As a result, no model unlocks credence features during conversation.

When showing examples, some models frequently hallucinate items that do not exist in the catalog: \claudehaiku in 47\% of conversations and \oss in 31\%. In these cases the user simulator prompts the model to verify against the catalog, after which it corrects itself, but the opportunity to surface genuine features from a real item is lost. 
Even when models show real items, they often omit features in their descriptions, such that the examples fail to grow $S_t$. The \textit{loops} column in Table \ref{tab:budgets} captures this: a \textit{loop} occurs when the model shows an item satisfying the user's known preferences and the user simulator --- programmed to always request more detail in this case --- receives a response in which the model just restates the same features rather than introducing new ones. All models except \gptfive loop frequently --- in 35\% of \gptmini's conversations, a request for more detail yields nothing new.

Models also vary in the types of questions they ask. 
20\% of the questions that \gptfive issues are confirmation questions: they ask the user to confirm preferences which have already been clearly stated in a prior turn.
For example, when a user requests a popular 1970s--80s ensemble film with ``a strong adult female lead'' and an ``ensemble team vibe with only mild action,'' \gptfive spends successive turns probing how strict each preference is rather than searching for items that satisfy them:
\begin{quote}
\small
\textit{(6) Is ``mystery/thriller'' a hard requirement (primary genre), or just a strong preference? (7) Is ``survival plot'' a hard requirement (characters trapped / disaster / hunted / must endure), or were you using ``survival'' more loosely (career/social survival, etc.)? (8) For the ``ensemble team vibe,'' do you mean a group solving the central mystery together, or is it okay if the lead drives the case and there's just a strong supporting cast? (9) Is comedy actually desired (e.g., ``mystery-comedy'' / ``comedic thriller''), or are you fine with a straight thriller as long as the ensemble mystery-solving is present? (10) How strict is the ``strong adult female lead'' requirement versus simply having an ensemble with at least one prominent adult woman?}
\end{quote}
We provide the full list of model questions from this interaction in Appendix Figure~\ref{fig:questions_movielens}; Appendix Figure~\ref{fig:questions} shows an analogous \taskhm example.

Together, these patterns --- confirmation questions that retrieve rather than construct preferences, hallucinated items, and loops that fail to introduce new features --- explain why no model recovers more than 54\% of the user's preference state despite using the full turn budget. Appendix \ref{app:qualitative} provides an extended qualitative analysis of transcripts.
\section{Related Work}\label{sec:related_work}
\textit{(Appendix \ref{app:related_work} surveys additional related work.)}

A body of work evaluates how language models \textbf{resolve underspecified requests} in task-oriented settings \citep{li2024mediq,yao2024tau,luo2024duetsim,shao2024collaborative,wu2025collabllm,qian2025userbench,pan2025benchmarks,vijayvargiya2025interactive},
but these works make the expert user assumption that we argued against in Section \ref{sec:user}.

Most \textbf{shopping agent evaluations} are non-interactive:
agents are prompted with fully-specified preferences $\phi(x^*)$ upfront
and evaluated solely on recall@$k$ \citep{yao2022webshop,zhou2023webarena,lyu2025deepshop,peeters2025webmall,wang2026shopsimulator}.
\citet{murakhovs2023salespeople} and \citet{chen2024chatshop} evaluate conversational recommender systems in underspecified settings, 
but both still simulate users with the expert user model, where all features are search features.
\citet{murakhovs2023salespeople} additionally measure how much the agent teaches the user about the domain;
we similarly emphasize the importance of user learning, but we evaluate this more directly via the user's state completeness and center it by focusing on team accuracy.

In \name, agents are evaluated based on \textit{team accuracy}: if agents fail to build up the user's domain knowledge, the user is too weak to verify the recommended slate. 
These ideas relate to the problem of \textbf{scalable oversight} \citep{christiano2018supervising,bowman2022measuring}, which studies how a non-expert human can effectively supervise a stronger model.
Our work considers a setting where the agent can augment the human's expertise so that the human might be better able to supervise.

A new line of work studies how to \textbf{simulate users} for training and evaluating task-oriented agents \citep{shi2019build,naous2025flipping,abdulhai2026consistently,suh2026quantifying,wu2026humanlm}.
These works are split by whether they take a bottom-up, data-driven approach (i.e., finetuning language models on user utterances) 
or a top-down, model-based approach (i.e., defining a model of human behavior to condition a language model on).
Because of the limited amount of multi-turn user data, the former community tends to emphasize simulating human utterance styles,
while the latter emphasizes simulating human behaviors.
Our work falls in the latter camp, where we specifically model preference construction.

Concurrent work by \citet{kim2026discoverllm} models user preferences as hierarchical. 
\frameworkname instead emphasizes \textit{heterogeneity}: 
some features (e.g., experience and credence features) require non-question dialog actions to construct. 
\citet{kim2026discoverllm} train agents against synthetic tasks with their user model and show this improves real user satisfaction, whereas we benchmark agents against \frameworkname users on recommendation tasks.
\section{Conclusion}
We introduced \frameworkname, a model of how users construct rather than retrieve preferences over the course of an interaction, and \name, a benchmark instantiating this model in agentic recommendation. 
Our results show that current agents systematically underdevelop user preference states.
We hope future work (1) explores methods to improve agent quality on \name, such as training with \frameworkname users, and (2) extends \frameworkname beyond the recommendation setting.


\paragraph{Using the \name package.}
\name is open-sourced as a package to support future agent and model evaluations. It can be accessed at \href{https://irenasaracay.github.io/coshop/}{this site}.
\section*{Acknowledgements}
The authors thank Theodora Worledge, Nicole Meister, Qinan Yu, Ed Chen, and other members of the Guestrin lab, as well as Miguel Liu-Schiaffini, Vignesh Kothapalli, and Marco Tulio Ribeiro, for discussions and feedback on this project. 

\bibliography{main}
\bibliographystyle{colm2026_conference}

\newpage
\appendix
\startcontents[appendices]
\printcontents[appendices]{}{1}{\section*{Appendix contents}}

\section{Human study}\label{app:human_study}
Below we provide additional details on the study design, participant demographics, and results for each hypothesis validated in Section~\ref{sec:human_study}.
This study was approved by our institution's IRB. 

\subsection{Study design}

We administered a Google Form structured around a simulated clothing shopping task. Participants were asked to imagine they were shopping for a sweater and that an AI shopping agent needed to understand their preferences before making recommendations. The form proceeded in two stages.

\paragraph{Stage 1: Pre-survey.} Participants first answered questions about their demographics, shopping habits, AI tool usage, and familiarity with sweaters.

\paragraph{Stage 2: Preference elicitation.} Participants completed 10 rounds of questions, one per sweater feature drawn from the \taskhm dataset (Section~\ref{sec:benchmark}). The features were drawn from the H\&M catalog; all participants saw the same 10 features: price, gender, target style intent, belt or tie presence, shoulder style, collar style, graphic placement, material, color family, and neckline.

In each round, the participant was shown a direct clarifying question about the feature (e.g., ``What neckline do you prefer for the sweater?'') and asked to respond in one of four ways:
\begin{enumerate}
    \item Answer directly in a free-text box.
    \item ``I'm not sure --- show me examples.''
    \item ``I'm not sure --- show me an explanation of what this means.''
    \item ``I don't care.''
\end{enumerate}
Responses (1) and (4) were interpreted as settled preferences; the feature was classified as a \textit{search} feature for that participant.
Responses (2) and (3) indicated that the participant needed additional grounding; the feature was classified as requiring \textit{experience} or \textit{credence} grounding, respectively.

Participants selecting option (2) were shown a set of example items illustrating variation along that feature. Participants selecting option (3) were shown a brief technical explanation of the feature generated by Claude Sonnet 4.6. In both cases, participants were then asked to state their preference in free text or select ``I don't care.'' A non-empty free-text response was interpreted as evidence that the participant successfully formed a preference after seeing the provided content.

\begin{figure}[tbh]
    \centering
    \includegraphics[width=\textwidth]{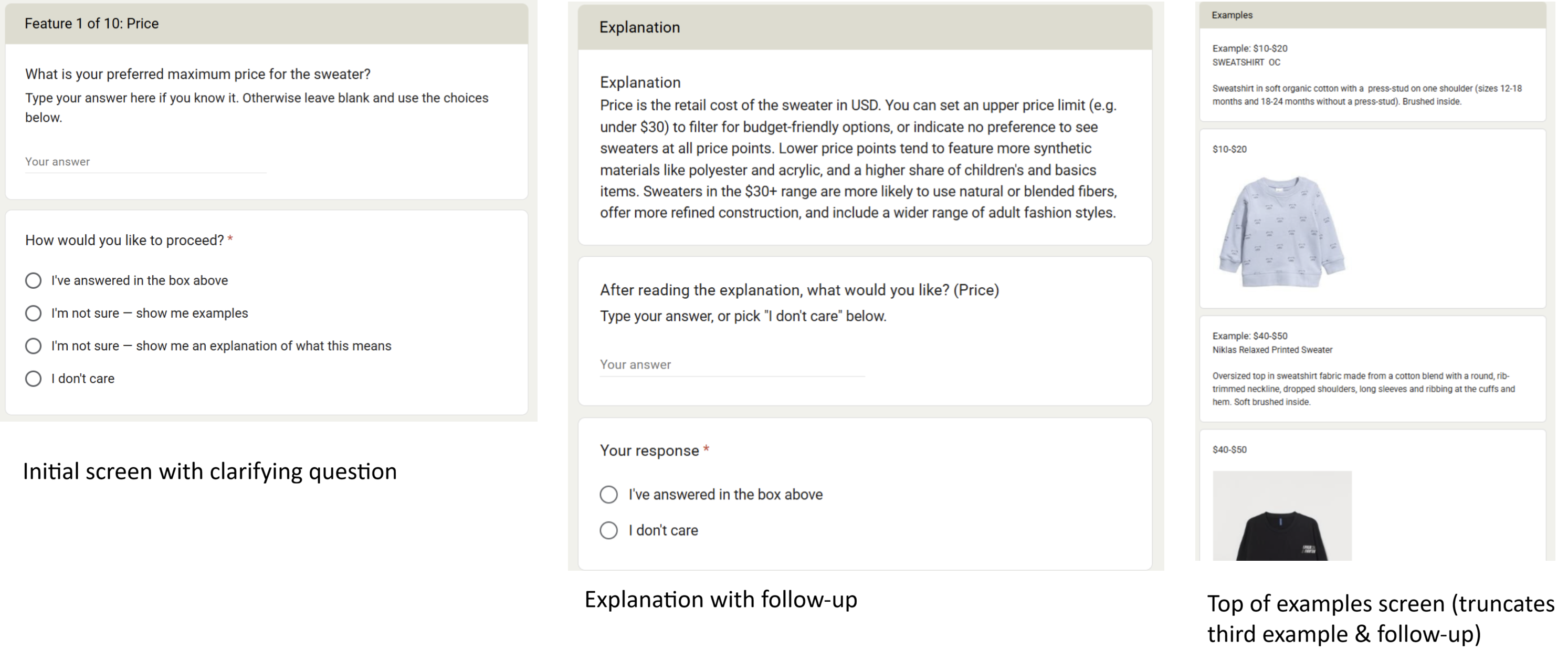}
    \caption{Screenshots from the human study survey illustrating the preference elicitation rounds. Left: the initial clarifying question with four response options. Right: an example of follow-up content (examples or a technical explanation) shown to participants who indicated uncertainty.}
    \label{fig:survey}
\end{figure}

\subsection{Participants}

\begin{figure}[tbh]
    \centering
    \includegraphics[width=\textwidth]{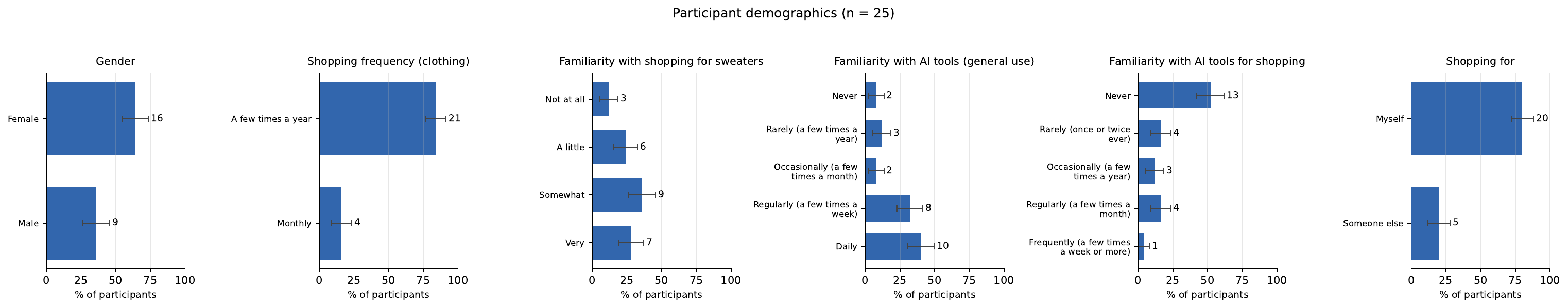}
    \caption{Participant demographics from the human study ($n=25$).}
    \label{fig:demographics}
\end{figure}

We recruited $n=25$ participants via word of mouth. The sample skewed young; we estimate the majority of participants were under 30. 16 participants identified as female and 9 as male. Most were frequent users of AI tools such as ChatGPT, Claude, and Gemini, though over 50\% had never used AI for shopping or gift recommendations. Participants self-reported a range of sweater-shopping expertise: only 7/25 (28\%) claimed to be very familiar with sweaters, consistent with our focus on non-expert users. Figure~\ref{fig:demographics} summarizes participant demographics.

\subsection{Limitations}

Our study design asks participants to decide in a single shot whether a non-search feature is experience or credence. In practice, a user might request examples and then realize they need a technical explanation after seeing those examples --- a more iterative process than our one-shot classification captures. Additionally, our sample is small ($n=25$) and skews young, which may limit generalizability. These limitations are consistent with the simplifying assumptions of \frameworkname discussed in Section~\ref{sec:user}.

\section{\name dataset details}\label{app:data}
\name tests agents on three datasets, each representing a separate product domain:
fashion recommendations (\taskhm),
movie recommendations (\movielens),
and book recommendations (\goodreads).
For each dataset, we source the catalog $\mathcal X$ and user profiles (including a target item $x^\ast$) from the original datasets.
We enrich the domain feature set $\mathcal F$ using regex-based feature extraction.
Section~\ref{app:data_sec} describes how we compute the search-experience-credence (SEC) split and the initial known feature set $S_1$ for each user.
In each per-dataset feature table below, the \textbf{S}, \textbf{E}, and \textbf{C} columns report the percentage of the 100 test users for whom that feature is classified as search, experience, or credence, respectively (these need not sum to 100\%, since a feature is excluded for users whose target value is missing or expresses no preference).

\subsection{\taskhm preprocessing}\label{app:data_hm}

\paragraph{Overview.}
The agent's goal is to help a user find a fashion item they would like from the \taskhm catalog \citep{hm}.
There are \numprint{37570} catalog items and $|\mathcal{F}| = 77$ features per item.
Across users, the average SEC split is $|\mathcal{F}_\text{s}| \approx 6.1$ search, $|\mathcal{F}_\text{e}| \approx 21.5$ experience, and $|\mathcal{F}_\text{c}| \approx 9.1$ credence features, with an average initial known feature set size of $|S_1| \approx 3.9$.

\paragraph{Users.}
We start from the original dataset's catalog and transaction log.
For each customer, we take their newest transaction as the target item $x^\ast$.
All remaining in-catalog purchases by the same customer form their purchase history.
We keep users who have at least 30 history items and retain the first 100 valid users.

\paragraph{Catalog.}
Each article's price is imputed as the maximum observed transaction price; articles without a price are dropped.
We exclude catalog items that are missing an image, belong to the undergarment section, or are not purchased by any of the 100 users in our evaluation set.
To enrich the catalog, we use \qwenvlfourbinstruct to generate a visual description from each product image and then use regular expressions to extract structured features.
The $|\mathcal{F}| = 77$ features span: \emph{basic attributes} (product category, pattern, color, price); \emph{fit and cut} (fit, sleeve length, neckline, waist rise, leg silhouette, dress length, skirt shape, silhouette shape); \emph{construction details} (lining, pockets, ribbing, drawstring, woven vs.\ knit fabric, seam type, wrap style); \emph{outerwear and performance} (jacket type, waterproof, windproof, functional finish); \emph{surface and decoration} (fabric texture, graphic type/theme/placement/scale, embroidery, sequins, trim, logo); \emph{part-specific details} (sleeve style, cuff, shoulder exposure, strap, armhole, hem, back, side features); \emph{ethical attributes} (sustainable, vegan, toxin-free); and \emph{demographics} (gender, age group, sportswear, formality).
Table~\ref{tab:app_hm_features} lists all features with their descriptions.

To avoid a situation where the user simulator asks for too specific of a price (e.g., ``Can I have a shirt priced at \$15.93?''), the simulator view of $\phi(x^*)$ coarsens the \texttt{price} feature: it is rounded up to the nearest \$10 and displayed as an upper bound (e.g.\ a price of \$15.93 becomes ``$\le$ \$20'').

{\small\setlength{\tabcolsep}{3.5pt}%
\begin{longtable}{@{}>{\raggedright\arraybackslash}p{0.60\textwidth}rrr>{\raggedright\arraybackslash}p{0.13\textwidth}@{}}
\caption{Catalog features in $\mathcal F$ for \taskhm{}.}
\label{tab:app_hm_features} \\
\toprule
\textbf{Feature description} & \textbf{S} & \textbf{E} & \textbf{C} & \textbf{Source} \\
\midrule
\endfirsthead
\caption[]{Catalog features in $\mathcal F$ for \taskhm{}. (continued)} \\
\toprule
\textbf{Feature description} & \textbf{S} & \textbf{E} & \textbf{C} & \textbf{Source} \\
\midrule
\endhead
\midrule
\multicolumn{5}{r}{{Continued on next page}} \\
\endfoot
\bottomrule
\endlastfoot
Age group (target age category: adult, child, infant) & 100 & 0 & 0 & regex \\
Armhole design (e.g. deep, standard) & 0 & 0 & 0 & regex \\
Back-specific construction details (e.g. back pleat, longer back hem, back opening, back yoke, elasticated back, back tie, back vent, back darts) & 4 & 11 & 17 & regex \\
Bodice construction (for dresses: upper-body construction style, e.g. wrap, fitted, smocked) & 0 & 0 & 5 & regex \\
Bottom garment in multipack (type of lower garment in a set, e.g. pajama shorts, pajama pants, leggings, jogger pants) & 0 & 0 & 0 & regex \\
Brand mark presence (location of brand identifier, e.g. chest logo, sleeve logo, brand tag) & 0 & 0 & 2 & regex \\
Closure type (primary fastening mechanism, e.g. zip, full front zip, button, snap/press-stud, hook-and-eye, toggle, lace) & 8 & 19 & 13 & regex \\
Color & 26 & 45 & 29 & original dataset \\
Color family & 25 & 56 & 19 & original dataset \\
Crotch design (for pant-type or full-body items: e.g. dropped crotch) & 0 & 0 & 0 & regex \\
Cuff construction (detail at the end of the sleeve, e.g. ribbed, elasticated, buttoned, thumbhole, tab-and-button, turn-up, adjustable) & 10 & 11 & 11 & regex \\
Decorative trim (type of ornamental trim applied to edges or seams, e.g. frill trim, lace trim, flounce, ruffle, scalloped, faux fur trim, broderie anglaise) & 5 & 4 & 10 & regex \\
Dress or skirt length (for full-body garments with a skirt portion: e.g. short, mini, midi, maxi, knee-length, calf-length, ankle-length) & 25 & 11 & 11 & regex \\
Fabric composition (fibre content detected in description or tags, e.g. cotton, polyester, wool, linen, silk, lyocell) & 0 & 0 & 63 & regex \\
Fabric surface finish (visual or tactile surface quality, e.g. satin sheen, brushed, creped, crinkled, washed) & 6 & 4 & 4 & regex \\
Fit (overall silhouette fit, e.g. slim, regular, relaxed, oversized, fitted, loose, muscle, bodycon) & 7 & 34 & 9 & regex \\
Fly closure type (for pant-type items: e.g. fly-front zip, button fly, fake fly) & 0 & 0 & 0 & regex \\
Formality level (e.g. casual, smart casual, formal) & 4 & 93 & 3 & regex \\
Functional performance finish (performance treatment or feature, e.g. water-repellent, fast-drying, wrinkle-resistant, thermal, breathable, UV protection, reflective) & 0 & 0 & 4 & regex \\
Garment length category (overall length, e.g. cropped, extended/longline, regular with longer back hem) & 2 & 5 & 5 & regex \\
Gender (target gender of the garment: female, male, unisex) & 100 & 0 & 0 & regex \\
Graphic location (where the graphic appears on the garment, e.g. chest, sleeve, back, front, all-over) & 5 & 19 & 4 & regex \\
Graphic size (relative scale of the graphic, e.g. small, large) & 0 & 0 & 4 & regex \\
Graphic subject matter (theme of the graphic or print, e.g. animals, floral, space/fantasy, vehicles, geometric, text/slogan, nature/nautical) & 1 & 14 & 4 & regex \\
Graphic type (nature of applied decoration, e.g. print motif, all-over print, logo, embroidery, sequin, appliqu\'e, text/slogan, glitter) & 1 & 1 & 2 & regex \\
Has belt loops (for pant-type items) & 0 & 0 & 0 & regex \\
Has cargo pockets (for pant-type items: large utility pockets on the leg) & 0 & 0 & 0 & regex \\
Has detachable hood (hood that can be removed) & 0 & 100 & 0 & regex \\
Has drawstring & 2 & 97 & 1 & regex \\
Has embroidery (embroidered decoration present) & 1 & 99 & 0 & regex \\
Has hood (whether the garment has a hood, derived from neckline) & 2 & 97 & 1 & regex \\
Has lining (whether the garment has any internal lining) & 1 & 97 & 2 & regex \\
Has pleating (for garments with a skirt portion: whether the skirt is pleated) & 16 & 28 & 8 & regex \\
Has pockets (any pockets present on the garment) & 1 & 93 & 6 & regex \\
Has ribbing (ribbing at cuffs, hems, waistbands or as a fabric panel; excludes decorative neckline trim) & 4 & 93 & 3 & regex \\
Has sequins or glitter embellishment & 0 & 100 & 0 & regex \\
Has text or slogan graphic (whether a prominent text or wording is part of the decoration) & 0 & 100 & 0 & regex \\
Has yoke panel (structured panel across the upper back or shoulders) & 0 & 0 & 100 & regex \\
Hem treatment (construction or shape of the hem, e.g. ribbed, raw-edge, scalloped, ruffled, flared, asymmetric, curved, straight-cut, elasticated, smocked) & 11 & 11 & 7 & regex \\
Interior finish (detail of the lining or inside treatment, e.g. lined, unlined, brushed, thermal lining, faux shearling) & 3 & 17 & 3 & regex \\
Is sport (whether the garment is marketed as sportswear or activewear) & 1 & 95 & 4 & regex \\
Is wrap style (garment that wraps and ties rather than closing with a fixed fastening) & 2 & 98 & 0 & regex \\
Jacket type (e.g. bomber, parka, puffer, trench coat, blazer, denim jacket, faux fur coat, shirt jacket, windbreaker) & 8 & 3 & 7 & regex \\
Leg length (for pant-type items: length of the leg, e.g. ankle, cropped, 7/8) & 0 & 0 & 0 & regex \\
Leg silhouette (for pant-type items: cut of the leg, e.g. straight, tapered, wide-leg, skinny, flare, jogger) & 0 & 0 & 0 & regex \\
Leg structure (lower-body configuration for full-body garments, e.g. playsuit/shorts, culottes, wrap skirt) & 0 & 0 & 0 & regex \\
Multipack format (for multi-piece sets: e.g. two-piece pack, three-piece pack) & 0 & 0 & 0 & regex \\
Neckline or collar (e.g. crew/round neck, V-neck, hooded, collared, off-shoulder, stand-up collar, turtleneck, funnel neck) & 27 & 43 & 12 & regex \\
Nursing or maternity design (special construction for nursing access, or garment marketed as maternity wear) & 0 & 0 & 3 & regex \\
Overall silhouette (garment outline shape, e.g. boxy, A-line, straight-cut, bodycon, peplum, flowing, fitted, draped, tailored) & 12 & 13 & 7 & regex \\
Pattern & 12 & 80 & 8 & original dataset \\
Pattern type (dominant visual pattern or colour arrangement, e.g. solid, stripe, floral, plaid/check, animal print, melange, colour block, camouflage, polka dots) & 14 & 77 & 9 & regex \\
Pocket location (where pockets are positioned, e.g. side, front, back, chest) & 1 & 10 & 6 & regex \\
Pocket style (configuration of pockets, e.g. kangaroo, side seam, welt, flap, patch, chest, zipped, cargo) & 3 & 6 & 6 & regex \\
Price & 14 & 60 & 26 & original dataset \\
Product category & 100 & 0 & 0 & original dataset \\
Seam style (visible or functional seam treatment, e.g. flatlock, taped, decorative) & 0 & 0 & 0 & regex \\
Shoulder construction (structural shoulder cut, e.g. dropped, raglan) & 5 & 17 & 3 & regex \\
Shoulder exposure (degree of shoulder baring, e.g. off-shoulder, cold-shoulder, one-shoulder, halter) & 0 & 0 & 3 & regex \\
Side construction details (features running along the side seam, e.g. side zipper, side slit, side seam pocket, wrapover side panel, ribbed panel) & 1 & 3 & 3 & regex \\
Skirt silhouette (for garments with a skirt portion: overall shape or structure, e.g. A-line, tiered, flared, pleated, straight, wrap, bodycon, flowing) & 9 & 13 & 8 & regex \\
Sleeve design (distinctive sleeve cut or shape, e.g. puff, raglan, wide, butterfly, trumpet, gathered) & 5 & 12 & 12 & regex \\
Sleeve graphic presence (decorative element located on the sleeve, e.g. sleeve logo, sleeve embroidery) & 0 & 0 & 0 & regex \\
Sleeve length (e.g. sleeveless, cap, short, 3/4, long) & 9 & 62 & 8 & regex \\
Strap design (shoulder strap style, e.g. adjustable straps, narrow/spaghetti straps, wide straps) & 1 & 0 & 1 & regex \\
Structural surface texture (fabric construction texture, e.g. quilted, mesh, ribbed, piqu\'e, creped, crinkled) & 6 & 2 & 5 & regex \\
Style intent (intended lifestyle or use context, e.g. athletic, outdoor, streetwear, loungewear/nightwear, everyday basic, occasion/formal) & 2 & 92 & 6 & regex \\
Sustainable (whether the garment uses sustainable materials or certifications, e.g. organic cotton, recycled polyester, Tencel, GOTS) & 0 & 0 & 100 & regex \\
Toxin-free certification (e.g. OEKO-TEX, bluesign) & 0 & 0 & 100 & regex \\
Top garment in multipack (type of upper garment in a set, e.g. pajama t-shirt, pajama shirt, sweatshirt top) & 0 & 0 & 0 & regex \\
Vegan (no animal-derived materials; false if wool, silk, leather, down etc. are present) & 0 & 0 & 100 & regex \\
Waist construction method (how the waist opening is structured or adjusted, e.g. elasticated, drawstring, smocked, gathered, seamed) & 0 & 0 & 19 & regex \\
Waist decorative styling (ornamental waist detail beyond the construction method, e.g. detachable belt, self-tie belt, cinched, sewn-on tie belt) & 9 & 2 & 5 & regex \\
Waist rise (for lower-body and full-body items: height of the waistband, e.g. high-waisted, mid-rise, low-rise) & 0 & 0 & 0 & regex \\
Waterproof or water-resistant & 0 & 100 & 0 & regex \\
Windproof or wind-resistant & 0 & 100 & 0 & regex \\
Woven fabric (true if the main fabric is woven, false if knit/jersey/fleece, null if indeterminate) & 0 & 0 & 100 & regex \\
\end{longtable}
}

\subsection{\movielens preprocessing}\label{app:data_movielens}

\paragraph{Overview.}
The agent's goal is to help a user find a movie they would like from the \movielens catalog \citep{movielens}.
There are \numprint{26637} catalog items and $|\mathcal{F}| = 92$ features per item.
Across users, the average SEC split is $|\mathcal{F}_\text{s}| \approx 8.6$ search, $|\mathcal{F}_\text{e}| \approx 70.2$ experience, and $|\mathcal{F}_\text{c}| \approx 8.8$ credence features, with an average initial known feature set size of $|S_1| \approx 1.7$.

\paragraph{Users.}
We start from the original dataset's rating log.
For each user, we sort ratings from newest to oldest, keeping only users who rated movies on at least five distinct calendar days.
The target item $x^\ast$ is the first movie that receives a rating of at least 4 out of 5 stars in that order.
All other rated in-catalog movies count as history; we drop users with fewer than 50 history entries and retain the first 100 valid users.

\paragraph{Catalog.}
We start from the MovieLens movie table linked to The Movie Database (TMDB) metadata: titles, release years, plot overviews, genres, vote averages and counts, popularity, runtime, budget, revenue, collection membership, production countries and companies, spoken languages, and user-contributed tags.
We exclude movies that do not appear in any of the 100 users' rating histories.
To enrich the catalog, we use regular expressions to infer additional narrative, tone, and theme attributes.
The $|\mathcal{F}| = 92$ features include: \emph{core metadata} (release year, rating, vote count, popularity, runtime, budget, revenue, genres, collection, production companies/countries, spoken languages); \emph{production attributes} (MPAA rating, animation style, award profile, source material, cinematographic format, home media availability); \emph{content flags} (nudity, remake, sequel, Bechdel test, IMDB top list, national film registry, black-and-white, era); and a large set of \emph{thematic and genre indicators} (83 binary features covering tones such as melancholic, slow burn, or dark comedy; narrative elements such as twist ending, road trip, or time travel; demographics such as female-led; and setting/subject tags such as high school, prison, or urban setting).
Table~\ref{tab:app_ml_features} lists all features with their descriptions.

To avoid a situation where the user simulator asks for too specific of a feature (e.g., a movie with exactly a \$1.3M revenue), the simulator view of $\phi(x^*)$ coarsens several fields.
\texttt{original\_language} ISO 639-1 codes are replaced with English language names.
\texttt{popularity} is replaced by a discrete label derived from the catalog distribution (e.g.\ \texttt{very popular}, \texttt{below average}).
\texttt{vote\_average} is floored to the nearest half-point and prefixed with ``$\ge$'';
\texttt{vote\_count} is floored to the nearest 100;
\texttt{budget} and \texttt{revenue} are floored to the nearest \$100{,}000;
\texttt{runtime} is rounded up to the next multiple of 30 and prefixed with ``$\le$'';
\texttt{year} is mapped to a two-decade phrase (e.g.\ ``1980s or 1990s'').

{\small\setlength{\tabcolsep}{3.5pt}%
\begin{longtable}{@{}>{\raggedright\arraybackslash}p{0.60\textwidth}rrr>{\raggedright\arraybackslash}p{0.13\textwidth}@{}}
\caption{Catalog features in $\mathcal F$ for \movielens{}.}
\label{tab:app_ml_features} \\
\toprule
\textbf{Feature description} & \textbf{S} & \textbf{E} & \textbf{C} & \textbf{Source} \\
\midrule
\endfirsthead
\caption[]{Catalog features in $\mathcal F$ for \movielens{}. (continued)} \\
\toprule
\textbf{Feature description} & \textbf{S} & \textbf{E} & \textbf{C} & \textbf{Source} \\
\midrule
\endhead
\midrule
\multicolumn{5}{r}{{Continued on next page}} \\
\endfoot
\bottomrule
\endlastfoot
Animation style of the film: 'live\_action', 'traditional\_animation', 'CGI', 'stop\_motion', or 'anime'. Null only when genre is unknown & 0 & 0 & 6 & regex \\
Average rating of the movie out of 5 & 55 & 35 & 10 & original dataset \\
Budget of the movie in USD & 0 & 0 & 70 & original dataset \\
Cinematographic formats used, e.g. ['70mm', 'anamorphic', 'imax', '3d']. Null if none detected & 0 & 0 & 10 & regex \\
Collections the movie was featured in & 0 & 0 & 100 & original dataset \\
Genres of the movie & 100 & 0 & 0 & original dataset \\
Highest level of award recognition: 'winner', 'nominated', 'festival\_selection', or null if none detected & 0 & 0 & 29 & regex \\
Languages the movie is available in & 5 & 72 & 23 & original dataset \\
Number of ratings for the movie & 53 & 15 & 32 & original dataset \\
Popularity score of the movie (higher is more popular) & 59 & 32 & 9 & original dataset \\
Production companies of the movie & 0 & 0 & 100 & original dataset \\
Production countries of the movie & 14 & 65 & 21 & original dataset \\
Release year & 13 & 58 & 29 & original dataset \\
Revenue of the movie in USD & 0 & 0 & 77 & original dataset \\
Runtime of the movie in minutes & 28 & 64 & 8 & original dataset \\
The decade in which the film is set (e.g. '1970s', '1980s'). Null if not recorded & 37 & 0 & 3 & regex \\
The type of source material the film is adapted from: 'book' (novel, memoir, or non-fiction), 'play' (stage play or TV show), or 'true\_story' (based on real events). Null if original screenplay or unknown & 22 & 5 & 0 & regex \\
Whether drug use or addiction is a significant theme & 3 & 97 & 0 & regex \\
Whether family dynamics are a central theme & 23 & 76 & 1 & regex \\
Whether friendship is a significant theme, including buddy films and unlikely alliances & 8 & 92 & 0 & regex \\
Whether isolation — physical, psychological, or social — is a significant theme or setting & 3 & 97 & 0 & regex \\
Whether mental health, psychological disorder, or emotional crisis is a significant theme & 7 & 93 & 0 & regex \\
Whether murder is a significant plot element & 12 & 88 & 0 & regex \\
Whether music is a significant thematic or narrative element (distinct from the musical genre) & 28 & 71 & 1 & regex \\
Whether revenge is a significant plot driver & 6 & 94 & 0 & regex \\
Whether survival is a central theme or challenge & 1 & 98 & 1 & regex \\
Whether terrorism is a significant theme & 2 & 98 & 0 & regex \\
Whether the film addresses racism or racial injustice & 2 & 98 & 0 & regex \\
Whether the film appears on an IMDb ranked list such as the IMDb Top 250 & 0 & 0 & 100 & regex \\
Whether the film centers on an unlikely or unexpected friendship & 0 & 100 & 0 & regex \\
Whether the film contains nudity & 21 & 79 & 0 & regex \\
Whether the film contains significant violence or gore & 18 & 82 & 0 & regex \\
Whether the film engages with political themes & 8 & 92 & 0 & regex \\
Whether the film features a found family theme & 0 & 100 & 0 & regex \\
Whether the film features a mentor-protege relationship & 0 & 100 & 0 & regex \\
Whether the film features a prison setting or incarceration as a significant element & 2 & 98 & 0 & regex \\
Whether the film features an anti-hero as its protagonist & 1 & 98 & 1 & regex \\
Whether the film features an ensemble cast rather than a single protagonist & 5 & 95 & 0 & regex \\
Whether the film features martial arts & 0 & 100 & 0 & regex \\
Whether the film features unrequited love & 0 & 100 & 0 & regex \\
Whether the film features zombies & 0 & 99 & 1 & regex \\
Whether the film follows a character's journey into maturity or self-understanding & 12 & 88 & 0 & regex \\
Whether the film has a female protagonist who drives the plot & 2 & 97 & 1 & regex \\
Whether the film has a melancholic tone & 1 & 98 & 1 & regex \\
Whether the film has a significant twist ending & 6 & 94 & 0 & regex \\
Whether the film has been released on a physical home media format (DVD, VHS, Blu-ray, Betamax, LaserDisc, etc.) & 59 & 6 & 35 & regex \\
Whether the film has been released on Blu-ray & 17 & 83 & 0 & regex \\
Whether the film has been selected for the US National Film Registry & 0 & 0 & 100 & regex \\
Whether the film has LGBTQ+ themes or characters & 3 & 97 & 0 & regex \\
Whether the film has religious or spiritual themes & 9 & 91 & 0 & regex \\
Whether the film has significant romantic content & 5 & 95 & 0 & regex \\
Whether the film has supernatural elements such as ghosts, monsters, vampires, or demons & 2 & 98 & 0 & regex \\
Whether the film involves a conspiracy or cover-up & 3 & 97 & 0 & regex \\
Whether the film involves a heist & 1 & 99 & 0 & regex \\
Whether the film involves a road trip & 0 & 99 & 1 & regex \\
Whether the film involves crime as a central element & 12 & 88 & 0 & regex \\
Whether the film involves espionage or spying & 4 & 96 & 0 & regex \\
Whether the film involves gangsters, the mob, or organized crime & 3 & 97 & 0 & regex \\
Whether the film involves time travel & 2 & 98 & 0 & regex \\
Whether the film is a comedy & 3 & 95 & 2 & regex \\
Whether the film is a dark comedy or black comedy & 11 & 89 & 0 & regex \\
Whether the film is a documentary & 0 & 99 & 1 & regex \\
Whether the film is a drama & 0 & 99 & 1 & regex \\
Whether the film is a fantasy, including magical realism and supernatural fantasy & 11 & 89 & 0 & regex \\
Whether the film is a horror film & 1 & 99 & 0 & regex \\
Whether the film is a mockumentary & 1 & 98 & 1 & regex \\
Whether the film is a musical & 6 & 94 & 0 & regex \\
Whether the film is a mystery, including detective stories and investigations & 10 & 90 & 0 & regex \\
Whether the film is a psychological thriller & 2 & 98 & 0 & regex \\
Whether the film is a remake of an earlier work & 5 & 95 & 0 & regex \\
Whether the film is a sequel or part of a franchise & 10 & 90 & 0 & regex \\
Whether the film is a slow burn — building tension or emotion gradually over time & 0 & 100 & 0 & regex \\
Whether the film is a sports film & 3 & 97 & 0 & regex \\
Whether the film is a superhero film & 2 & 98 & 0 & regex \\
Whether the film is a thriller & 13 & 87 & 0 & regex \\
Whether the film is a western & 0 & 99 & 1 & regex \\
Whether the film is about war or armed conflict & 11 & 89 & 0 & regex \\
Whether the film is an action film & 19 & 79 & 2 & regex \\
Whether the film is an adventure film & 8 & 92 & 0 & regex \\
Whether the film is available on Netflix & 12 & 88 & 0 & regex \\
Whether the film is biographical or autobiographical & 3 & 97 & 0 & regex \\
Whether the film is satirical or a parody & 11 & 89 & 0 & regex \\
Whether the film is science fiction & 8 & 91 & 1 & regex \\
Whether the film is set in a dystopian or post-apocalyptic world & 4 & 96 & 0 & regex \\
Whether the film is set in a small town & 1 & 99 & 0 & regex \\
Whether the film is set in an urban environment such as a major city & 8 & 92 & 0 & regex \\
Whether the film is set in or focused on a historical period & 8 & 92 & 0 & regex \\
Whether the film is set in or prominently features a high school & 5 & 95 & 0 & regex \\
Whether the film is set in or significantly involves outer space & 3 & 97 & 0 & regex \\
Whether the film is set in suburbia & 1 & 99 & 0 & regex \\
Whether the film is shot in black and white & 3 & 97 & 0 & regex \\
Whether the film passes or fails the Bechdel Test: 'pass', 'fail', or null if not recorded & 0 & 0 & 100 & regex \\
\end{longtable}
}

\subsection{\goodreads preprocessing}\label{app:data_goodreads}

\paragraph{Overview.}
The agent's goal is to help a user find a book they would like from the \goodreads catalog \citep{goodreads1,goodreads2}.
There are \numprint{39050} catalog items and $|\mathcal{F}| = 102$ features per item.
Across users, the average SEC split is $|\mathcal{F}_\text{s}| \approx 7.8$ search, $|\mathcal{F}_\text{e}| \approx 67.0$ experience, and $|\mathcal{F}_\text{c}| \approx 13.5$ credence features, with an average initial known feature set size of $|S_1| \approx 3.8$.

\paragraph{Users.}
We start from the original dataset's reading log, keeping interactions that have a marked read date.
For each user, we sort reads from newest to oldest.
The target item $x^\ast$ is the first book in that order with usable genre labels.
All other reads form the user's history; we drop users with fewer than 5 history entries and retain the first 100 valid users.

\paragraph{Catalog.}
We start from the public Goodreads book dump, keeping only books with non-empty titles and descriptions.
We attach series metadata, author name, and aggregated genre labels per work where available, and add a small random sample of review text per book.
We exclude works that do not appear in any of the 100 users' reading histories.
To enrich the catalog, we use regular expressions to infer additional plot, tone, and trope-level features.
The $|\mathcal{F}| = 102$ features include: \emph{bibliographic metadata} (author, genres, average rating, publication year, page count, publisher, series name/length/position, review and rating counts); \emph{format and publication} (ebook, audiobook, physical format, indie/self-published, debut); \emph{content descriptors} (world type, setting type, age group, emotional tone, violence level, NSFW, happy ending, cliffhanger, magic system, technology level, paranormal creatures); \emph{award and cultural} (award status, book club pick, classic, associated cultures, summer read, chick lit, art/music focus, based on true story, illness theme, LGBTQ themes, cookbook details, nonfiction subtype); \emph{narrative and plot devices} (narrative devices, plot devices, fantasy/mystery/sci-fi subgenre, fanfiction origin); \emph{themes} (24 binary theme flags covering death, love, family, war, identity, crime, etc.); \emph{tropes} (21 binary romance and narrative trope flags); and \emph{romance subgenres} (10 binary flags).
Table~\ref{tab:app_gr_features} lists all features with their descriptions.

The simulator view of $x^\ast$ coarsens several fields.
The \texttt{genres} list is truncated to at most the first three labels.
\texttt{series\_works\_count} is bucketed (0, 1, 2--3, 4--10, 10+); \texttt{series\_position} maps to either first-in-series or a coarse mid-series label.
\texttt{average\_rating} is floored to the nearest half-point and prefixed with ``$\ge$''; \texttt{ratings\_count} and \texttt{text\_reviews\_count} are floored to the nearest 100; \texttt{publication\_year} is floored to the nearest multiple of 100 (e.g.\ 1997 $\rightarrow$ ``$\ge$ 1900'') and prefixed with ``$\ge$''; \texttt{num\_pages} is rounded up to the next multiple of 100 and prefixed with ``$\le$''.

{\small\setlength{\tabcolsep}{3.5pt}%
\begin{longtable}{@{}>{\raggedright\arraybackslash}p{0.60\textwidth}rrr>{\raggedright\arraybackslash}p{0.13\textwidth}@{}}
\caption{Catalog features in $\mathcal F$ for \goodreads{}.}
\label{tab:app_gr_features} \\
\toprule
\textbf{Feature description} & \textbf{S} & \textbf{E} & \textbf{C} & \textbf{Source} \\
\midrule
\endfirsthead
\caption[]{Catalog features in $\mathcal F$ for \goodreads{}. (continued)} \\
\toprule
\textbf{Feature description} & \textbf{S} & \textbf{E} & \textbf{C} & \textbf{Source} \\
\midrule
\endhead
\midrule
\multicolumn{5}{r}{{Continued on next page}} \\
\endfoot
\bottomrule
\endlastfoot
Author name & 0 & 0 & 100 & original dataset \\
Average rating of the book & 11 & 71 & 18 & original dataset \\
Award recognition: winner, nominated, or absent & 0 & 0 & 12 & regex \\
Broad type of world the story takes place in: real\_world, fantasy\_world, or sci\_fi\_world & 9 & 83 & 6 & regex \\
Cultures meaningfully associated with the book, covering author origin, story setting, and subject matter — may include both specific (e.g. Korean, French) and regional (e.g. Asian, European) labels & 14 & 4 & 27 & regex \\
Fantasy type: epic\_or\_high, urban, dark\_or\_grimdark, romantic\_fantasy, fairy\_tale\_retelling, mythological, portal, cozy\_fantasy, other & 0 & 0 & 29 & regex \\
For cookbooks: cuisine type and skill level (e.g. Italian beginner, Baking all\_levels) & 0 & 0 & 2 & regex \\
For nonfiction books: memoir\_or\_autobiography, biography, travel\_writing, academic\_or\_scholarly, reference\_or\_how\_to, inspirational\_or\_devotional, self\_help, parenting\_or\_family\_guidance, cultural\_criticism, narrative\_nonfiction, investigative\_journalism, true\_crime, history, business & 0 & 0 & 32 & regex \\
Genres of the book & 100 & 0 & 0 & original dataset \\
Key plot mechanics present: heroic\_quest, whodunit, survival, political\_intrigue, forced\_partnership, heist\_or\_con, competition\_or\_tournament, undercover\_mission, artifact\_quest & 0 & 0 & 29 & regex \\
Level of LGBTQ+ content: central\_to\_plot, significant\_theme, minor\_or\_background, or absent & 9 & 0 & 5 & regex \\
Level of violence depicted: mild, moderate, or graphic & 20 & 41 & 9 & regex \\
Mystery or thriller type: cozy, hardboiled, police\_procedural, psychological\_thriller, amateur\_sleuth, supernatural\_mystery, domestic\_thriller, legal\_thriller & 0 & 0 & 37 & regex \\
Name of the series the book belongs to & 0 & 0 & 51 & original dataset \\
Number of books in the series & 32 & 41 & 27 & original dataset \\
Number of pages in the book & 31 & 26 & 21 & original dataset \\
Number of ratings for the book & 12 & 9 & 79 & original dataset \\
Number of text reviews for the book & 10 & 30 & 60 & original dataset \\
Paranormal creature types present: vampire, werewolf, angel, demon, zombie, fae, witch\_or\_mage, ghost, dragon, god\_or\_deity, mutant & 0 & 0 & 15 & regex \\
Position of the book in the series & 24 & 6 & 7 & original dataset \\
Primary emotional register: dark\_or\_grim, tense, melancholic, hopeful, humorous, whimsical, cozy\_or\_warm, bittersweet, angsty & 39 & 24 & 17 & regex \\
Primary intended readership age group: middle\_grade, young\_adult, new\_adult, adult, or academic & 33 & 51 & 16 & regex \\
Primary physical or social environment: urban, small\_town\_or\_rural, isolated\_or\_wilderness, institutional, workplace, aristocratic\_or\_court & 27 & 10 & 19 & regex \\
Publisher of the book & 0 & 0 & 72 & original dataset \\
Romance subgenre: christian or clean romance & 0 & 98 & 2 & regex \\
Romance subgenre: contemporary romance & 18 & 76 & 6 & regex \\
Romance subgenre: dark romance & 0 & 99 & 1 & regex \\
Romance subgenre: erotic romance & 15 & 68 & 17 & regex \\
Romance subgenre: historical romance & 4 & 94 & 2 & regex \\
Romance subgenre: new adult romance & 7 & 87 & 6 & regex \\
Romance subgenre: paranormal romance & 6 & 89 & 5 & regex \\
Romance subgenre: Regency romance & 1 & 98 & 1 & regex \\
Romance subgenre: romantic suspense & 2 & 94 & 4 & regex \\
Romance subgenre: sci-fi or fantasy romance & 2 & 96 & 2 & regex \\
Science fiction type: space\_opera, dystopian, cyberpunk, military, biopunk, cli\_fi, hard\_sci\_fi, other & 11 & 2 & 4 & regex \\
Structural narrative techniques used: nonlinear, dual\_timeline, flashback\_heavy, unreliable\_narrator, epistolary, frame\_narrative, slow\_revelation, social\_fable & 0 & 0 & 4 & regex \\
Technological era of the story's setting: pre\_industrial, contemporary, near\_future, or far\_future & 0 & 0 & 77 & regex \\
Theme: abuse, trauma, or domestic violence & 1 & 95 & 4 & regex \\
Theme: adventure, exploration, or quest & 11 & 78 & 11 & regex \\
Theme: crime, murder, or criminal activity & 27 & 62 & 11 & regex \\
Theme: cultural difference, clash, or cross-cultural encounter & 2 & 96 & 2 & regex \\
Theme: death, dying, or mortality & 11 & 83 & 6 & regex \\
Theme: faith, religion, or spirituality & 0 & 94 & 6 & regex \\
Theme: family relationships and dynamics & 8 & 82 & 10 & regex \\
Theme: feminism or women's rights & 7 & 87 & 6 & regex \\
Theme: freedom of speech, censorship, or expression & 0 & 99 & 1 & regex \\
Theme: friendship or found family & 16 & 78 & 6 & regex \\
Theme: horror or psychological terror & 17 & 77 & 6 & regex \\
Theme: identity, self-discovery, or coming out & 1 & 95 & 4 & regex \\
Theme: interpersonal drama or emotional conflict & 16 & 68 & 16 & regex \\
Theme: love or romantic relationships & 27 & 56 & 17 & regex \\
Theme: magic or the supernatural & 8 & 82 & 10 & regex \\
Theme: mythology, legend, or folklore & 3 & 94 & 3 & regex \\
Theme: nature, the environment, or animals & 3 & 95 & 2 & regex \\
Theme: power, politics, or political intrigue & 5 & 91 & 4 & regex \\
Theme: psychology or mental health & 11 & 83 & 6 & regex \\
Theme: redemption, sacrifice, or atonement & 0 & 97 & 3 & regex \\
Theme: survival against extreme odds & 5 & 92 & 3 & regex \\
Theme: technology or scientific progress & 7 & 87 & 6 & regex \\
Theme: time travel & 0 & 98 & 2 & regex \\
Theme: war, military conflict, or its aftermath & 6 & 90 & 4 & regex \\
Trope: alpha or brooding hero & 6 & 87 & 7 & regex \\
Trope: chosen one or prophesied hero & 0 & 99 & 1 & regex \\
Trope: coming of age & 3 & 88 & 9 & regex \\
Trope: enemies-to-lovers & 3 & 92 & 5 & regex \\
Trope: fake dating or fake relationship & 0 & 98 & 2 & regex \\
Trope: forbidden love or star-crossed romance & 0 & 99 & 1 & regex \\
Trope: forced or arranged marriage & 1 & 98 & 1 & regex \\
Trope: forced proximity & 0 & 99 & 1 & regex \\
Trope: found family & 0 & 99 & 1 & regex \\
Trope: friends-to-lovers & 2 & 94 & 4 & regex \\
Trope: love triangle & 1 & 97 & 2 & regex \\
Trope: mentor and protégé relationship & 0 & 0 & 0 & regex \\
Trope: mistaken or swapped identity & 0 & 99 & 1 & regex \\
Trope: opposites attract & 1 & 96 & 3 & regex \\
Trope: rags to riches & 0 & 99 & 1 & regex \\
Trope: redemption arc & 0 & 98 & 2 & regex \\
Trope: revenge plot & 0 & 98 & 2 & regex \\
Trope: second chance romance & 1 & 95 & 4 & regex \\
Trope: secret baby & 0 & 99 & 1 & regex \\
Trope: secret identity or double life & 0 & 98 & 2 & regex \\
Trope: slow burn romance & 1 & 97 & 2 & regex \\
Whether a serious illness or terminal diagnosis is a significant plot element & 1 & 90 & 9 & regex \\
Whether creating or inhabiting art, music, or another creative discipline is central to the plot or a main character & 3 & 86 & 11 & regex \\
Whether the book contains explicit sexual content & 6 & 72 & 22 & regex \\
Whether the book ends on an unresolved cliffhanger that requires reading the next installment & 3 & 95 & 2 & regex \\
Whether the book features an explicit, rule-based magic system & 0 & 0 & 100 & regex \\
Whether the book has been tagged as available in a physical print format (paperback or hardcover) & 14 & 79 & 7 & regex \\
Whether the book is a collection or anthology of multiple works (short stories, poems, essays, comics, etc.) rather than a single continuous work & 10 & 82 & 8 & regex \\
Whether the book is available as an audiobook & 36 & 51 & 13 & regex \\
Whether the book is available as an ebook & 0 & 84 & 4 & regex \\
Whether the book is inspired by or closely depicts real events or people & 4 & 88 & 8 & regex \\
Whether the book is part of a series & 30 & 50 & 20 & original dataset \\
Whether the book is tagged as a beach or summer read & 0 & 97 & 3 & regex \\
Whether the book is tagged as a book club selection & 0 & 0 & 100 & regex \\
Whether the book is tagged as chick-lit or women's fiction & 17 & 73 & 10 & regex \\
Whether the book is widely regarded as a literary classic & 0 & 97 & 3 & regex \\
Whether the book originated as or was adapted from fanfiction & 0 & 99 & 1 & regex \\
Whether the book resolves with a happy ending (True), an unhappy one (False), or unknown (null) & 4 & 0 & 1 & regex \\
Whether the book was self-published or released through an indie publisher & 0 & 98 & 2 & regex \\
Whether this is the author's debut published book & 0 & 98 & 2 & regex \\
Year the book was published & 0 & 74 & 2 & original dataset \\
\end{longtable}
}

\subsection{Computing user SEC splits}\label{app:data_sec}

For each user $i$ and target item $x^\ast$, we classify every feature $f$ as \emph{search}, \emph{experience}, or \emph{credence} based on how well-established the user's preference for $x^\ast$'s value was before the purchase.

Features are scored using a base-rate-adjusted surprise metric.
Let $v = \phi(x^*)_f$ be the target value and let $p_\text{user}(f{=}v)$ and $p_\text{catalog}(f{=}v)$ be the empirical frequency of the value $v$ for feature $f$ across the user's purchase history and catalog,  respectively.
The score is the pointwise mutual information between the catalog and user distributions:
\[
  s(f, x^*) \;=\; \log_2\!\frac{p_\text{catalog}(f{=}v)}{p_\text{user}(f{=}v)}
  \;=\; -\log_2 p_\text{user}(f{=}v) \;+\; \log_2 p_\text{catalog}(f{=}v).
\]
Features are then classified as:
\begin{itemize}[leftmargin=1.8em,itemsep=0pt]
  \item \textbf{Search} ($s < -1$): the user over-selects this value relative to the catalog, indicating a settled prior preference.
  \item \textbf{Experience} ($-1 \le s < 1$): the user selects this value at roughly the catalog rate; no strong directional preference signal.
  \item \textbf{Credence} ($s \ge 1$, or value never seen in history, or feature almost always missing in the catalog): the user substantially under-selects this value or has never purchased it.
\end{itemize}

Three edge cases are handled before scoring.
(1)~If $\phi(x^*)_f$ is missing or indicates no preference, the feature is excluded entirely.
(2)~If value $v$ never appears in the user's history ($p_\text{user}(f{=}v) = 0$), the score is undefined; we assign a large sentinel value and classify the feature directly as credence.
(3)~If the feature is structurally unobserved for nearly all catalog items (over 95\% missing), the catalog marginal is unreliable and the feature is classified as credence.

\subsection{Computing user initial preference states}\label{app:data_initial_state}
The initial feature set $S_1$ for each user contains the search features the agent knows at the start of the conversation.
To compute $S_1$, we simulate an incremental catalog search with a fixed retrieval budget: starting from a seed of always-known search features (e.g.\ genre), we iteratively add one search feature at a time and generate a natural-language search query from the currently known features using \oss.
The query is issued to a vector search index over catalog item descriptions with a retrieval budget of $k = 250$ items.
We stop adding features once $x^\ast$ appears in the top-$k$ results, or once all search features have been included.
The resulting feature set---the minimal set of search features sufficient for the target item to be retrievable by a search query---becomes $S_1$.
\FloatBarrier
\section{User simulator details}\label{app:simulator}
Our user simulator plays the shopper in multi-turn dialogues.
The user simulator has two main functions: to exchange messages with the agent during elicitation, 
and to select the final $x^\ast$ from the agent's report at the end of the interaction.

Following Section~\ref{sec:user}, the simulator maintains a \emph{preference state} $S_t := \phi_{S_t}(x^*)$ as a growing set of known catalog columns.
Unlike typical LLM-based user simulators, which prompt a model once with just a persona and let it improvise an entire conversation,
we explicitly anchor our simulator LLM \oss to the current preference state $S_t$.
Concretely, we generate natural language messages conditioned only on the current preference state $S_t$, and we have \oss select a final item conditioned only on the final preference state $S_{T+1}$ and the agent's report (with item order randomized to prevent position bias).

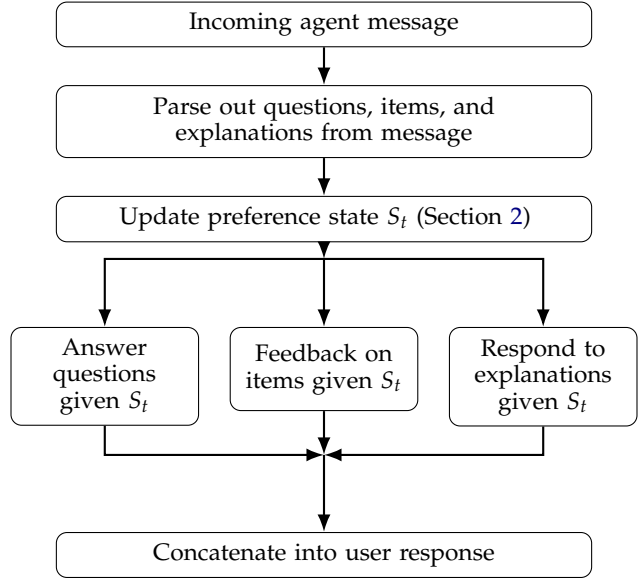
\begin{figure}[h]
  \centering
  \begin{tikzpicture}[
    font=\footnotesize,
    node distance=5mm,
    box/.style={rectangle, draw, rounded corners, align=center, inner sep=4pt},
    wide/.style={box, text width=6.8cm},
    branch/.style={box, text width=2.2cm, minimum height=1.1cm},
    arrow/.style={-Latex, thick},
  ]
    \node[wide] (msg) {Incoming agent message};
    \node[wide, below=of msg] (parse) {Parse out questions, items, and explanations from message};
    \node[wide, below=of parse] (upd) {Update preference state $S_t$ (Section \ref{sec:user})};
    \coordinate (fork) at ($(upd.south) + (0,-0.22)$);
    \node[branch, anchor=north] (ans) at ($(fork) + (-2.9,-0.9)$) {Answer questions given $S_t$};
    \node[branch, anchor=north] (fb)  at ($(fork) + (0,-0.9)$) {Feedback on items given $S_t$};
    \node[branch, anchor=north] (exp) at ($(fork) + (2.9,-0.9)$) {Respond to explanations given $S_t$};
    \node[wide, below=16mm of fb] (out) {Concatenate into user response};
    \coordinate (meet) at ($(fb.south) + (0,-0.58)$);
    \draw[arrow] (msg) -- (parse);
    \draw[arrow] (parse) -- (upd);
    \draw[arrow] (upd.south) -- (fork);
    \draw[arrow] (fork) -| (ans.north);
    \draw[arrow] (fork) -- (fb.north);
    \draw[arrow] (fork) -| (exp.north);
    \draw[arrow] (ans.south) |- (meet);
    \draw[arrow] (fb.south) -- (meet);
    \draw[arrow] (exp.south) |- (meet);
    \draw[arrow] (meet) -- (out.north);
  \end{tikzpicture}
  \caption{One turn of the user simulator: parse the agent message into questions, items, and explanations; update the preference state $S_t$; generate a response for each component; concatenate into the user reply.}
  \label{fig:gated-feature-user-flow}
\end{figure}

\paragraph{Overview.}
Upon receiving a message, the simulator first checks that any item recommendations use the expected structured JSON format; a lightweight classifier may ask the agent to fix formatting if items are clearly recommended but not tagged.
It then parses the message into atomic clarifying questions, recommended items, and credence-style explanations, maps each to relevant catalog features, adds referenced features unlocked via the correct action to the preference state as described in Section \ref{sec:user}, and then generates a natural-language reply from independent LM calls for each component (Figure~\ref{fig:gated-feature-user-flow}).
If nothing structured is extracted, the simulator falls back to a single freeform LLM turn with the full persona and conversation history.
At most five features are revealed per turn in total (across all components), limiting how much preference information any single exchange can transfer.
We parallelize LM calls where possible using LOTUS \citep{patel2024semantic}.

Each simulator instance is also given a lightweight \emph{persona} derived from that user's purchase or rating history, used only in the fallback freeform turn and in the question-answering prompt to provide conversational grounding.
The persona does not contain the target item $x^\ast$ or any of its feature values.
The persona content differs by dataset:
\begin{itemize}
  \item \textbf{\taskhm}: age, club membership status, newsletter frequency, number of prior purchases, average and maximum historical spend, and clothing size.
  \item \textbf{\goodreads}: total books rated, count of high ratings ($\ge 4$ stars), and top-3 genres and top-3 authors by frequency.
  \item \textbf{\movielens}: total movies rated, count of high ratings ($\ge 4$ stars), and top-3 genres and top-3 spoken languages by frequency.
\end{itemize}

\subsection{Prompts}

\paragraph{Clarifying questions.}
For each extracted question, a semantic filter identifies which catalog features are relevant to answering it.
The LM is then asked to answer briefly in the first person, conditioned on the current preference state $S_t$ and the relevant features.
It is instructed to defer (e.g., ``I don't know yet'' or ``I'm open to anything'') whenever those features are not yet in $S_t$.

\medskip\noindent\textbf{Extracting atomic questions.}
This step turns the raw agent message into a list of atomic, self-contained questions that ask the shopper about preferences or constraints.
It deliberately drops questions about choosing among shown items and meta-requests.

\begin{Verbatim}[breaklines=true, breakanywhere=false, fontsize=\footnotesize, frame=single, rulecolor=\color{black}, framerule=0.4pt, framesep=8pt]
You are a question extractor. Given a message from a customer service agent to a shopper, extract ONLY the clarifying questions FROM THE MESSAGE.

Clarifying questions ask about the user's underlying preferences, requirements, background, or constraints for the product search. These questions gather missing information the assistant needs in order to better tailor future recommendations. Examples:
- "Have you ever purchased wool?"
- "What size do you prefer?"
- "Do you have a budget in mind?"
- "Are you willing to relax any constraints?"
- "You said X. Could you clarify what you mean?"
- "Let me know if you have other preferences (e.g., sustainability)" -> "Do you have a preference for sustainability?"

Do NOT include questions that:
- ask for feedback or opinions on specific items or sets of items that were recommended,
- ask the user to choose between items already shown,
- ask what the assistant should do next (e.g., see more details, restart the search, show similar items),
- check whether an explanation was helpful,
- or otherwise talk about the conversation/meta-strategy without asking for new constraints or preferences.

These are NOT clarifying questions and must be EXCLUDED:
- "What do you think of item <item><id>123</id><information>Blue dress.</information></item>?"
- "How do you feel about this one?"
- "Would you like me to start my search over again?"
- "Do you want to see more details?"
- "Do you have any questions?"
- "Would you like to see more colors?"
- "Would you like to see items similar to this one?" 
- "Let me know if you'd like to narrow the list down."
- "Which of these is your favorite?"
- "Which of these three catches your interest the most?"
- "Which of these sounds like the best fit for you?"
- "Do any of these items look interesting to you?"
- "Does that explanation help?"
- "Do you have a preference among these options?"

Breaking Down Questions into Atomic Units

Split compound questions into individual atomic questions, where each question asks about exactly one attribute. For example:
"To narrow it down, could you let me know if you have any preferences for color, material (e.g., cotton, linen, silk), or a price range you'd like to stay within?"
becomes:
1. "Do you have any preferences for color?"
2. "Do you have any preferences for material (e.g., cotton, linen, silk)?"
3. "Do you have any preferences for price range?"

Likewise, any question that bundles multiple attributes with "or" must be split. For example:
"Do you have any preferred colors or patterns?"
becomes:
1. "Do you have any preferred colors?"
2. "Do you have any preferred patterns?"

Exception (binary choice questions). Do NOT split a question that asks the user to choose between exactly two options. For example:
"Do you prefer a standalone novel, or are you open to reading a book that's part of a series?"
This remains a single question.

Making Each Question Standalone
Every question must be fully self-contained with no unresolved references.
- Rewrite follow-ups that reference a prior question. For example, "Do you want red or blue? And how much more do you like one than the other?" becomes:
  1. "Do you want red or blue?"
  2. "How much more do you like red than blue, or vice versa?"
- Omit questions that cannot be rewritten because their reference cannot be resolved. For example, "Are you truly open to either style?" must be dropped if "either style" is undefined.
- Expand vague references to options. For example, "Do you have a preference among these fabric options?" must be rewritten to name the options explicitly: "Do you have a preference among these fabric options (cotton / polyester)?"

Return each clarifying question on its own line. If there are no clarifying questions, return exactly: NONE
Do not include any other text, numbering, or explanation. DO NOT MAKE UP QUESTIONS: ONLY EXTRACT THEM FROM THE FOLLOWING MESSAGE.
\end{Verbatim}

\medskip\noindent\textbf{Identifying features relevant to the question.}
For each extracted question, we use LOTUS's semantic filter operation to decide which features are relevant to answering the question. Each feature is filtered using the following prompt:

\begin{Verbatim}[breaklines=true, breakanywhere=false, fontsize=\footnotesize, frame=single, rulecolor=\color{black}, framerule=0.4pt, framesep=8pt]
The user has a preference for feature {column_name} with description '{description}'. The preferred value is {values}. Decide if this preference is directly relevant to answering this clarification question from a customer service agent: {question}
\end{Verbatim}

\medskip\noindent\textbf{Generating question answer.}
The LM generates a short first-person reply conditioned on the string rendering of the current preference state $S_t$, the question text, and a natural-language hint built from the relevant features and their values in $S_t$.

\begin{Verbatim}[breaklines=true, breakanywhere=false, fontsize=\footnotesize, frame=single, rulecolor=\color{black}, framerule=0.4pt, framesep=8pt]
You are the SHOPPER (customer) answering a question from the customer service agent.

Your background: {simulator_persona}

Your current preferences (use only these to answer):
{z_context}

Make sure you provide the preference for this feature when answering the question: {feature_hint}
Try not to directly quote this and paraphrase it instead.

Question from the agent: {question}

Answer in 1-2 sentences. Be brief and conversational, but not too confident (e.g., 'Red would be nice' rather than 'I want red and nothing else.'). If your preferences above do not contain the answer, say "I don't know yet" or "I'm open to anything". Do not invent details.
\end{Verbatim}

\paragraph{Item examples.}
For each recommended item, the simulator uses the agent's textual description to evaluate, per known feature, whether the item matches the target preference (match / mismatch / not enough information).
The LM is then prompted to write two sentences of feedback conditioned on the current state $S_t$, emphasizing mismatches and surfacing newly realized preferences.

\medskip\noindent\textbf{Evaluating whether items match preferences.}
For each item, we use LOTUS's semantic map operation to identify if the item matches user preferences. The operation classifies whether the agent's description implies a value that matches, contradicts, or leaves undetermined the target, yielding \texttt{YES}, \texttt{NO}, \texttt{not\_mentioned}, or \texttt{agent\_uncertain}.

\begin{Verbatim}[breaklines=true, breakanywhere=false, fontsize=\footnotesize, frame=single, rulecolor=\color{black}, framerule=0.4pt, framesep=8pt]
Evaluate whether the text supports the following requirement.

Feature: {column_name}
Meaning: {column_description}
Required value: {true_value}

Decision process:
1. Look for text that provides clear evidence about the feature {column_description}.
2. Infer the value implied by the text, or if there is not enough information, return not_mentioned.
3. Compare the inferred value to {true_value}. Determine whether the implied value semantically matches the user's requirement. Be generous: treat synonyms, paraphrases, compatible values, and values that fall within the user's stated range as matching. Only say NO if the implied value clearly contradicts or does not satisfy the user's requirement.

Be generous with matching. For example, if the feature name is "buttoned-cuffs" and the presented value is "buttons" and the requirement is True, these are the same: the presented value is that the buttoned cuffs have buttons. If the feature name is "product type" and the presented value is "dress" and the actual value is "garment full body" these are also the same: a dress is a kind of full body garment. Another example: a series with 0 books in it is the same as a series with 1 book in it (both values mean 'standalone book'). A movie rated PG-13 or G is the same as being "not rated R." If a movie has Latin dub and the requirement is "Spoken language: Swedish or Latin", then this movie matches the requirement. If a movie is mentioned as having two protagonists (a male and a female), and the requirements is "Protagonist gender: female", then this movie matches the requirement: at least one protagonist has the right gender.

Only say NO if the implied value clearly contradicts the user's requirement. If a book described as a biogrpahy, then a feature like 'genre = fantasy' is NO.

Output rules:
- YES -> the text implies a value that does not contradict {true_value} or semantically matches it.
- NO -> the text implies a value that contradicts {true_value}.
- agent_uncertain -> the text mentions {column_description} but says the value is unknown or not in the catalog.
- not_mentioned -> the text contains no evidence about {column_description}.

Return exactly one of:
YES
NO
agent_uncertain
not_mentioned

Only return YES or NO if there is enough information presented to make a clear inference. If the feature is not explicitly mentioned, favor not_mentioned.

Text:
{message}
\end{Verbatim}

\medskip\noindent\textbf{Shopper feedback on a recommended item.}
Given $S_t$, the agent's description, and the per-feature match summary, the model writes brief, conversational feedback: emphasize failures and uncertainty, allow at most one positive aside, and surface newly realized preferences when the item interaction triggers them.

\begin{Verbatim}[breaklines=true, breakanywhere=false, fontsize=\footnotesize, frame=single, rulecolor=\color{black}, framerule=0.4pt, framesep=8pt]
You are the SHOPPER (customer) evaluating an item the agent recommended.

My current preferences (what I'm looking for):
{z}

Item ID: {item_id}

Description from the agent:
{raw_description}

=====================

How this item compares to my requirements:
{comparison_section}

Features I care about that the agent didn't clarify for this item:
{missing_section}

New preferences I just realized matter (and how this item fits):
{newly_revealed_section}

=====================

Write exactly 2 sentences of brief, natural feedback on how the item matches or doesn't match your requirements. YOU DO NOT CARE ABOUT ANY FEATURES OTHER THAN THE REQUIREMENTS ABOVE. Keep it conversational, not formal, like a text message. You do NOT need to list every preference that the item satisfies -- at most, mention one thing that looks good. Clearly point out the ways the item fails your requirements. If any mismatches are because you just realized you care about something new, say that explicitly (e.g., "Oh, I actually do care about color and I want it to be blue, so this red one doesn't quite fit"). For any feature that is missing or unclear, especially newly realized ones, phrase it as an uncertainty statement to the agent (e.g., "I'm not sure if it has adjustable shoulder straps, which I'd like."). Write in the FIRST PERSON as the shopper. Paraphrase, don't quote the preferred values.
\end{Verbatim}

\paragraph{Explanations.}
The simulator detects paragraph-length educational spans in the agent's message and maps each to a catalog feature.
Only explanations that cover the target value allow the corresponding credence feature to enter $S_t$.
The LM then generates a response acknowledging the agent's wording and, where the feature is now understood, articulating the desired value.

\medskip\noindent\textbf{Detecting explanations.}
The LM identifies paragraph-length educational spans, maps each to a catalog column, and judges whether the explanation's options or ranges plausibly include the simulator's target value.

\begin{Verbatim}[breaklines=true, breakanywhere=false, fontsize=\footnotesize, frame=single, rulecolor=\color{black}, framerule=0.4pt, framesep=8pt]
You are a helper that identifies and extracts extended educational explanations from a customer service agent message.

An EXPLANATION is when the agent dedicates AT LEAST A PARAGRAPH of text to teaching the user about a feature: what it means or what values/options it can take (e.g. "Sustainable items are made with...", "Some lengths include floor-length, ankle-length, calf-length...", "Some patterns include stripes & florals."). Do NOT treat questions as explanations: e.g. "What skirt length are you looking for?" is a question, not an explanation; return no explanation for that. Do not treat short lists inside questions ("What colors (e.g. blue, red) are you looking for?") as explanations. This is not a paragraph-length explanation.

You are given:
1. The assistant message.
2. A JSON object whose KEYS are catalog column names and whose values describe each column and give example values. This JSON is restricted to the small subset of catalog features we are currently interested in (the CREDENCE features).
3. A JSON object "target_values" whose KEYS are the same catalog column names and whose values are lists of the TARGET value(s) for that feature (the item(s) we care about). Use this to set "target_value_mentioned_or_in_range".

Your task: If the agent explains a feature (describes what the feature is or what options/values it can take), return an object with:
- "relevant_column": the exact catalog column name from the JSON keys that is being explained (e.g. dress_length for skirt/dress length, graphical_appearance_name for patterns).
- "target_value_mentioned_or_in_range": boolean. True if the explanation mentions or includes the target value for this feature. The explanation may describe a range or list of options; if the target value (or a broader category that includes it) is mentioned, set true. Examples: if the explanation says "Dress patterns include stripes and polka dots" and the target value is "large stripes", set true (stripes is mentioned). If the target value is "chevron" or "solid", set false (not mentioned). If the explanation lists "floor-length, ankle-length, calf-length" and the target value is "calf-length", set true.
- "justification": a short sentence explaining why the target value is mentioned or in range.
There should only be one object per feature in the list.

Examples:
- Message: "A dress's skirt length refers to how long it is on the knee. Some common lengths include floor-length, ankle-length, calf-length, knee-length, and mini, meaning above the knee. Typically longer lengths are considered more modest. Shorter lengths can sometimes elongate the leg." with target value "calf-length" -> {"relevant_column": "dress_length", "target_value_mentioned_or_in_range": true, "justification": "The explanation lists 'calf-length' as a common length."}
- Message: "What skirt length are you looking for?" -> [] (no explanation; this is a question)
- Message: "Some patterns for dresses include stripes & florals." with target value "large stripes" -> {"relevant_column": "graphical_appearance_name", "target_value_mentioned_or_in_range": true, "justification": "The explanation lists 'stripes' as a pattern from which an average internet user could derive the target value 'large stripes'."}
- Message: "Some patterns for dresses include stripes & florals." with target value "chevron" -> {"relevant_column": "graphical_appearance_name", "target_value_mentioned_or_in_range": false, "justification": "The explanation does not mention the target value 'chevron'."}
- Message: "Price refers to the cost of the dress. A typical price for a dress is $50 to $100." with target value "under $70" -> {"relevant_column": "price", "target_value_mentioned_or_in_range": true, "justification": "The target range (0-70) overlaps with the explanation range (50-100)."}
- Message: "Price refers to the cost of the dress. A typical price for a dress is around $100." with target_value "under $20" -> {"relevant_column": "price", "target_value_mentioned_or_in_range": false, "justification": "The target range (0-20) does not overlap with the explanation range, which is around 100. 'Around 100' can be treated as (80-120)."}
- Message: "What kind of material a dress is made of affects its look. For the specifications you gave, I would recommend materials like cotton and polyester." with target value "viscose" -> {"relevant_column": "material", "target_value_mentioned_or_in_range": false, "justification": "The explanation doesn't list the target value 'viscose', and an average internet user could not derive viscose from the other presented options."}

Catalog columns (JSON):
{features_json}

Target values per column (JSON):
{target_values_json}

Assistant message:
{message}

If there are no such explanations, return an empty list.

REMEMBER: an explanation is at least a paragraph of text dedicated to teaching the user about a single feature.

Output format: Return ONLY a JSON array of objects, each with exactly: "relevant_column" (string) and "target_value_mentioned_or_in_range" (boolean). No other keys or text.
\end{Verbatim}

\medskip\noindent\textbf{Shopper reply after the agent explains credence features.}
The model acknowledges the agent's wording, then paraphrases the desired values for features so the turn reads like a natural shopper reaction to a tutorial.

\begin{Verbatim}[breaklines=true, breakanywhere=false, fontsize=\footnotesize, frame=single, rulecolor=\color{black}, framerule=0.4pt, framesep=8pt]
The customer service agent just provided technical explanations for the following features:

{credence_to_answer} {credence_to_acknowledge}

STRICT RULE: If a technical term or hidden feature name does not appear in the agent's own wording, you must NEVER use it in your response -- not even to refer back to what they said. Only use the language and concepts the agent actually used when talking to you, in plain, everyday words.

Write a response in first person acknowledging these explanations and then communicating the desired values for these features: {credence_to_answer}

Here are our desired values:

{z}

Write in first person as the shopper with these preferences, and use natural, conversational language that avoids technical jargon and column or tag names. Do not directly quote any line from the preferences specification (for example, never say things like 'Has stretch: True'); instead, paraphrase it in natural, conversational language.

Example 1 (materials): Agent explains different materials after asking "Do you care about the material or fabric?" -> Shopper: "Thanks, that helps a lot. Based on what you said, I'd really like something in a softer, breathable fabric that feels comfortable to wear for a long time."

Example 2 (decades): Agent talks about different decades after asking "Is there a decade for a movie you're interested in?" -> Shopper: "Got it, that makes sense. From your explanation, I think I'd most enjoy something set around the 80s vibe you described."

Example 3 (waist type): Agent explains waist styles after asking "What kind of waist would you like?" -> Shopper: "Thanks for breaking that down. From what you described, I'd prefer the kind of waist that feels more structured and flattering rather than super loose."
\end{Verbatim}

\FloatBarrier 

\subsection{Evaluating internal validity: parsing quality}\label{app:simulator_evaluation}
To assess the simulator's \emph{internal validity}, 
we evaluated whether the two core parsing steps --- mapping questions to relevant features, and judging item-preference matches --- produce outputs consistent with human judgment on random samples drawn from test runs (Table~\ref{tab:message-parser-questions} and Table~\ref{tab:message-parser-features}).
All evaluations use \oss as the underlying language model.

Note that in Appendix \ref{app:structured}, we also run \namestructured,
a variant of the benchmark that requires agents to explicitly name which features their dialog actions target. 
This reduces the simulator parsing steps evaluated here to just two-way item-feature matches (match / contradict the target values).
We observe that removing the parser dramatically drops agent performance.

For \textbf{clarifying questions} (Table~\ref{tab:message-parser-questions}), annotators judged which catalog columns are relevant to answering each question; the simulator may output a set of columns or \emph{abstain} (empty set).
We score \textbf{question-level} accuracy as the fraction of questions on which either (i) both simulator and human abstain, or (ii) the human does not abstain and the simulator names at least one column the human selected.
On $n{=}50$ such questions, accuracy is 74.0\%, and when both sides commit to at least one column, agreement rises to 91.3\%.
Treating simulator abstention as the positive class, abstention precision is 59.3\% and recall is 100\%: whenever the simulator abstains, a human would also abstain in only about three fifths of cases, so the simulator \textbf{over-abstains} relative to humans.
The \textbf{feature-level} rows aggregate micro-precision, recall, and F1 over every (question, column) pair; compared to a \textbf{random} baseline that flips a fair abstention coin per question (and, at the feature level, resamples as many predicted columns per question as the simulator, uniformly from the allowlist), the simulator remains far stronger (e.g.\ micro-F1 0.711 vs.\ 0.099).

\begin{table}[htb]
  \centering
  \caption{Simulator: catalog columns marked relevant per clarifying question vs.\ human annotations ($n{=}50$). Random: abstention coin-flip per question; feature-level random resamples column counts from the allowlist. Metric definitions are in the prose above.}
  \label{tab:message-parser-questions}
  \begin{tabular}{@{}lrr@{}}
    \toprule
    & \textbf{Simulator} & \textbf{Random}  \\
    \midrule
    \multicolumn{3}{@{}l}{\textit{Question-level}} \\
    Accuracy & 0.740 & --- \\
    Accuracy among non-abstentions (both sides) & 0.913 & --- \\ 
    Abstention precision & 0.593 & 0.317 \\
    Abstention recall & 1.000 & 0.492 \\
    Abstention F1 & 0.744 & 0.384 \\
    \midrule
    \multicolumn{3}{@{}l}{\textit{Feature-level}} \\
    Micro-recall    & 0.640 & 0.089 \\
    Micro-precision & 0.800 & 0.111 \\
    Micro-F1     & 0.711 & 0.099 \\
    \bottomrule
  \end{tabular}
\end{table}

For \textbf{item--feature judgments} (Table~\ref{tab:message-parser-features}), humans labeled whether the agent's description of an item, relative to a known preference, \emph{matches} the target value, \emph{contradicts} it, or leaves the feature \emph{underspecified}.
On $n{=}50$ such (item, feature) pairs, the simulator's 3-way classification matches humans with 94.0\% accuracy, versus 33.3\% when guessing one of the three labels uniformly at random and 48.0\% under a per-judgement \textbf{majority} baseline.
Separately, collapsing to whether the simulator claims enough evidence to judge (non-missing vs.\ missing), precision for non-missing is 92.9\% with recall 100\% (F1 96.3\%), indicating reliable detection of when the text supports a firm match or mismatch.

\begin{table}[htb]
  \centering
  \caption{Item--feature evaluation: 3-way match / mismatch / insufficient information vs.\ humans ($n{=}50$). Random: uniform over the three labels; Majority: majority vote among the three. Task wording and metrics are in the prose above.}
  \label{tab:message-parser-features}
  \begin{tabular}{@{}lrrr@{}}
    \toprule
    & \textbf{Simulator} & \textbf{Random} & \textbf{Majority} \\
    \midrule
    \multicolumn{4}{@{}l}{\textit{Accuracy}} \\
    \quad Accuracy & 0.940 & 0.333 & 0.480 \\
    \midrule
    \multicolumn{4}{@{}l}{\textit{Non-null presence}} \\
    \quad Precision & 0.929 & 0.524 & 0.519 \\
    \quad Recall    & 1.000 & 0.501 & 0.999 \\
    \quad F1     & 0.963 & 0.510 & 0.684 \\
    \bottomrule
  \end{tabular}
\end{table}

\section{RAG agent details}\label{app:baselines}
In Section \ref{sec:results}, we benchmark language models on \name. 
These models use a simple RAG agent harness, 
which we implement using \texttt{langchain}.

To keep context length manageable across multi-step agentic rollouts, we summarize accumulated tool results rather than carrying their full text forward.
Every tool message is left intact for the first action that immediately follows it; only once a second unsummarized result arrives does compression fold the stale result into a summary.
Summaries are generated by \oss, and each tool result is summarized at most once.


\subsection{Baseline RAG agent prompts}
\label{app:baselines:natural}
This is the primary harness we benchmark with.
The agent is prompted with a high-level description of the SEC model and the three dialog actions.

\medskip\noindent\textbf{Conversation system message}

\begin{Verbatim}[breaklines=true, breakanywhere=false, fontsize=\footnotesize, frame=single, rulecolor=\color{black}, framerule=0.4pt, framesep=8pt]
You are a helpful assistant working with a user to find the right {item_name}. This conversation is a preference-development phase: your goal is not to find the answer yet, but to understand and help the user develop their preferences so the search that follows can succeed. You have a small budget to search the catalog to find example items for discussion during this phase. After the conversation, you will have a separate, larger budget to search the catalog thoroughly. Use the conversation to prepare for that step — not to solve it.

Note that the only preferences we care about for the user are features relevant to available items. The user will be able to reason best over actual catalog features, rather than general questions about their context.

Communicate with the user in natural language. Whenever you reference a specific catalog item — whether introducing it, answering a follow-up question about it, or adding new details — you MUST include a JSON line for it. Each JSON line must appear on its own line, contain an 'id' field with the item's id from the web results, and include all item-specific details as additional keys. Use double quotes for all keys and string values. Do not put any other text on that line. Example: {"id": 1234, "color": "black", "pattern": "black-and-white floral"}.
CRITICAL: If the user asks a follow-up question about a specific item, your text answer is fine, but you MUST still include a JSON line for that item in the same response — even if you already showed it before. All item-specific details belong in the JSON; do not reference them only in prose.To show the user a message, do not make tool calls in that message.

You have a budget of 5 turns for this conversation. One turn consists of a user message followed by your response. You can ask at most 20 clarifying questions. The user can review at most 10 unique items. Use your turns wisely to understand the user's preferences, constraints, and context.

Keep your messages brief to avoid overburdening the user. You do not need to find the item during this conversation — you only need to understand and develop the user's preferences. When you feel confident about their preferences, generate <END_CONVERSATION> to move to the research phase. Do not generate <END_CONVERSATION> in any other message, or the interaction will end prematurely.

People vary in how much they already know about their own preferences, and not every preference can be unlocked by asking a direct question.

A useful framing is to think about three situations a user might be in for any given feature:

1. The user already has a settled preference. They can answer a direct question accurately — they have reliable past experience with this feature and know what they want. Asking directly is the right move.

2. The user can form a preference by seeing examples. They have no strong prior opinion and cannot answer abstractly, but will react clearly when shown a concrete item. Asking abstractly may not work here — they need to see it first. The right move is to show items that highlight this feature, then invite the user to react.

3. The user needs domain knowledge before they can form any preference. They have not encountered this feature before, or the terminology is unfamiliar. Even seeing examples may not help without first understanding what the feature means. The right move is to briefly explain the feature and why it matters, then ask.

You will need to infer which situation applies — it is not told to you. Some signals: confident direct answers suggest a settled preference; hesitation or uncertainty suggests the user may need examples or explanations.

When multiple features fall into situations 2 or 3, be selective: surface only the most impactful unknown(s) at a time. Addressing every underspecified feature at once creates cognitive overload.
\end{Verbatim}

\medskip\noindent\textbf{Research system message} 

\begin{Verbatim}[breaklines=true, breakanywhere=false, fontsize=\footnotesize, frame=single, rulecolor=\color{black}, framerule=0.4pt, framesep=8pt]
========= FINAL RECOMMENDATION TIME =========
Your task is to thoroughly explore the web using the search_web tool and return the top {k} items that best match the user's preferences.

Based on your conversation with the user and your understanding of their preferences, constraints, and context:
Use the search_web tool to explore items. You can make a total of 250 items across all queries.

The moment you don't issue a tool call, this step will end.

After exploring, return a JSON object with a single key "top_k_items" containing a list of exactly {k} item IDs in ranked order (best match first).

The JSON object should have this format:
{
    "top_k_items": ["item_id_1", "item_id_2", ..., "item_id_{k}"]
}

Forget all other formatting instructions from before this. You are no longer talking to the user, so return ONLY the JSON object as a string without any other text or formatting. The item IDs should be the "id" field from the search_web results
\end{Verbatim}

\medskip\noindent\textbf{Report generation system message}

\begin{Verbatim}[breaklines=true, breakanywhere=false, fontsize=\footnotesize, frame=single, rulecolor=\color{black}, framerule=0.4pt, framesep=8pt]
You are given a set of candidate items to recommend to a user. You have already seen the conversation with the user in your existing context.

Your task is to produce one JSON object per catalog id (in the order shown below) explaining fit for the user, using the SAME output convention as when you describe items in this task: each value MUST be a JSON object (not a prose string) with an "id" field matching that catalog id, feature keys with updated values where helpful, and any extra keys you need to justify fit (e.g. rationale in string fields). Use double-quoted keys and string values. Do not wrap item details in free-floating prose outside those objects. Aim to make each object self-contained so the user can decide from these JSON lines alone.

Items (in recommended order):
[one line per id: id and catalog_json value, inserted at runtime]

Return ONLY a JSON object mapping item ids (strings) to JSON objects.
Example:
{
  "id_1": {"id": "id_1", "color": "navy", "why_it_fits": "Matches your preference for dark solids."},
  "id_2": {"id": "id_2", "pattern": "striped", "why_it_fits": "Conflicts with your dislike of busy patterns."}
}

Principles:
- You are an impartial, comprehensive, and honest advisor. Your main job is to provide the user with the information they need to make a decision on what is best for themselves.
- Write to the user in second person, and reflect their preferences and constraints in your response.
- Make sure your descriptions are entirely self-contained.

Include EVERY item. Do not include any other text.
\end{Verbatim}

\subsection{History-aware agent prompts}
\label{app:baselines:history}

In Section \ref{sec:intervention_results}, we gave agents access to a tool that allows it to access user history.
This design extends the baseline RAG agent with access to the user's historical purchase/rating data via a pandas REPL tool.
The tool description below is what the model sees in its tool list.


\begin{Verbatim}[breaklines=true, breakanywhere=false, fontsize=\footnotesize, frame=single, rulecolor=\color{black}, framerule=0.4pt, framesep=8pt]
Run a single pandas expression over the user's history, exposed as a pandas DataFrame named `df` and indexed by item id. Only `df` and `pd` are available; the expression must be a single pandas/Python expression with no imports, assignments, or statements. Each call is independent: `df` is reset to the user's full, original history every time, so results do not carry over between calls — express each query against the complete history. The available columns are not listed here; inspect them yourself (e.g. "df.columns.tolist()" or "df.head()") before relying on specific column names. Examples: "df[df.user_rating_of_5 >= 4].sort_values('year').head(10)"; "df.groupby('genre')['user_rating_of_5'].mean()"; "len(df)". At most 10 rows are returned for table/series results, and the returned text is truncated to 2000 characters — prefer expressions that select only the columns/rows you need.
\end{Verbatim}

\subsection{Structured-action agent prompts}
\label{app:baselines:structured}
In Appendix \ref{app:structured}, we experimented with using a structured-action agent.
The conversation system message is identical to the baseline agent except that the item-format instructions are replaced with the structured-action block below (all other text is unchanged).


\begin{Verbatim}[breaklines=true, breakanywhere=false, fontsize=\footnotesize, frame=single, rulecolor=\color{black}, framerule=0.4pt, framesep=8pt]
Communicate with the user in natural language when appropriate, but you MUST encode specific dialog actions as separate lines using this exact format: one keyword, a single space, then one JSON object (double-quoted keys and strings).

1) ASK_QUESTION — ask the user about a preference or constraint.
ASK_QUESTION {"question": "What is your budget?", "relevant_features": ["price"]}
relevant_features: catalog feature names this question is about (may be multiple).

2) SHOW_ITEM_FOR_FEEDBACK — proactively present an item for the user to react to.
SHOW_ITEM_FOR_FEEDBACK {"item_id": "12345", "features_for_feedback": {"color": "navy", "material": "cotton"}}
item_id: id from search/web results. features_for_feedback: map from catalog feature name to the value you are showing the user (all item-specific details must appear here).

3) ITEM_FOLLOWUP — use this whenever you answer a user question about a specific item, or need to surface additional details about an item you already showed. Your prose answer is fine, but you MUST also emit this action so the item details are captured.
ITEM_FOLLOWUP {"item_id": "12345", "features_for_feedback": {"color": "navy", "material": "cotton"}}
Use the same field names as SHOW_ITEM_FOR_FEEDBACK. Include all item-specific details the user needs inside features_for_feedback — not in the surrounding prose.

4) EXPLAIN — a paragraph or more teaching the user about a feature or concept.
EXPLAIN {"explanation_text": "Dress length usually means...", "relevant_features": ["dress_length"]}
relevant_features: catalog feature names the explanation addresses.

Actions can be freely combined within the same message. For example, you can show an item highlighting certain features with SHOW_ITEM_FOR_FEEDBACK, then immediately follow it with an ASK_QUESTION about one of those features — this lets you ground a question in something concrete the user just saw. Similarly, an EXPLAIN can be paired with an ASK_QUESTION to teach and then probe in a single turn.

Rules:
- One action per line; freely mix action lines with normal prose on other lines.
- If you ask a question in prose without ASK_QUESTION, the user cannot respond to it.
- If you discuss an item without SHOW_ITEM_FOR_FEEDBACK or ITEM_FOLLOWUP, the item details will not be captured.
- All feature names must exactly match catalog column names.
\end{Verbatim}

\subsection{Catalog search tool}\label{app:query_evaluation}
Agents are given a \texttt{search\_web} tool backed by HNSW vector search
\citep{malkov2016efficient} over \qwenthreeembeddingeightb embeddings of featurized catalog items $\phi(x)$ for $x \in \mathcal X$ \citep{zhang2025qwen3}.
Each call takes a natural-language query string, a \texttt{max\_items} count, and an optional \texttt{filters} dict mapping catalog column names to lists of exact allowed values.
The tool returns the top-$m$ items ranked by embedding similarity, restricted to items whose column values match all filter conditions.
The agent thus has two complementary handles on the catalog: free-form semantic query and explicit column-value pinning.

The following is the tool description the agent sees in its tool list.

\begin{Verbatim}[breaklines=true, breakanywhere=false, fontsize=\footnotesize, frame=single, rulecolor=\color{black}, framerule=0.4pt, framesep=8pt]
Search the web for products. The first argument is a string used to search the web. The second argument is the maximum number of items to return for this query. The more specific the query, the more likely you are to find a particular product. If the query is too specific, you may get no results. An empty query will return the entire web. The search query should be in natural language.

Optionally, supply a 'filters' dict mapping column names to a list of EXACT allowed values to restrict results to items matching those column values. This can help denoise your results. Use 'NA' as a value to match rows where that column is missing/null. 
IMPORTANT: there is no support for negation (no NOT) or inequality (no <, <=, >, >=).

Budget constraints: Total budget of {budget} items across all queries.
\end{Verbatim}

\medskip\noindent\textbf{Automatic prefilter.}
To increase the catalog search tool's recall, every query also passes through an automatic prefiltering step before vector search.
A language model (\oss, temperature 0) reads the raw query string and the list of filterable catalog columns and generates a list of simple pandas \texttt{DataFrame.eval()} boolean expressions.
These are applied greedily in a round-robin over multiple orderings: each expression is kept only if it does not reduce the matching set to zero; kept expressions intersect to form the filtered candidate pool.
The vector search then runs over the intersection of this pool and any hard-filter ids; if the result is still fewer than \texttt{max\_items}, the system supplements from an unrestricted baseline query.
The prompt used to generate prefilter expressions is below.

\begin{Verbatim}[breaklines=true, breakanywhere=false, fontsize=\footnotesize, frame=single, rulecolor=\color{black}, framerule=0.4pt, framesep=8pt]
[system]
You produce a list of SIMPLE pandas DataFrame.eval() expressions. Each expression must evaluate to a boolean Series. We will apply them one by one and take the intersection of matching rows; if an expression would leave zero rows, we skip it. So each expression should be ONE simple clause (e.g. one column comparison). Do NOT combine multiple conditions with & or | in a single expression — use separate list items instead. Use only column names provided. For string literals always use single or double quotes (e.g. 'Comedy', "Drama"). Be case sensitive. Use backticks ONLY for column names that contain spaces or are not valid Python identifiers (e.g. `Area (cm^2)`), never for string values. When the user lists multiple values for the same attribute, use one expression per value (e.g. one item "genres.str.contains('Comedy')", another "genres.str.contains('Drama')"). Return a JSON object with a key 'expressions' containing a list of expression strings. No explanation, no code block. Try not to read into semantic meaning. Filter based on obvious intent. Avoid filtering on titles or descriptions. Fewer, simpler expressions are better than one long combined expression. DO NOT add any filters that are not clearly implied by the string, in particular do NOT add an is_ebook filter. For example, 'book popular' does not mean you should threshold on avg_rating or num_ratings because that is reading too much into the string, and 'recent' does not mean you can safely filter on a year because that is reading too much into the string.

[user]
Query: {query}

Filterable columns (name -> type): {column_types}

Example values by column: {examples_by_column}

Give a list of simple pandas eval expressions (booleans) to filter rows. Return JSON: {"expressions": ["expr1", "expr2", ...]}
\end{Verbatim}

\FloatBarrier
\section{Additional \name results}\label{app:additional_results}
\subsection{Detailed results}

\paragraph{Agent performance across user types, by dataset.}
Figure~\ref{fig:underspecification_curve}A in the main text shows the performance drop
across user types averaged across datasets.
Table~\ref{tab:full_knowledge_by_dataset} shows the same comparison
(fully-specified, search-only, \name) broken out per dataset.

\paragraph{Utility metrics by dataset.}
Table~\ref{tab:main_results} reports exact-match item accuracy. Table~\ref{tab:utility_by_dataset} reports utility-based results instead.
Unlike exact match, utility rewards recommendations that are close to $x^\ast$
even when they do not match it exactly.
It is computed as 
\begin{equation}
  u(x, x^*) :=
\frac{1}{|\mathcal F(x^*)| + 1}
\left(
  \text{EmbeddingSimilarity}(x, x^*)
\sum_{j \in \mathcal F(x^*)} 1\{\phi_j(x^*) = \phi_j(x)\}
\right)
\end{equation}
where EmbeddingSimilarity is rescaled cosine similarity between \qwenthreeembeddingeightb embeddings of the $x$ and $x^\ast$ catalog entries.

Agent utility under complete reports (``oracle reports'') shows how much team utility
is lost to report quality alone.

\begin{table}[htb]
  \centering
  \resizebox{\linewidth}{!}{
      \begin{tabular}{cccccccccc}
\toprule
\multicolumn{1}{c}{} & \multicolumn{3}{c}{\taskhm} & \multicolumn{3}{c}{\movielens} & \multicolumn{3}{c}{\goodreads} \\
\cmidrule(lr){2-4} \cmidrule(lr){5-7} \cmidrule(lr){8-10}
\multicolumn{1}{c}{} & \multicolumn{1}{c}{Fully-specified} & \multicolumn{1}{c}{Search only} & \multicolumn{1}{c}{\frameworkname} & \multicolumn{1}{c}{Fully-specified} & \multicolumn{1}{c}{Search only} & \multicolumn{1}{c}{\frameworkname} & \multicolumn{1}{c}{Fully-specified} & \multicolumn{1}{c}{Search only} & \multicolumn{1}{c}{\frameworkname} \\
\cmidrule(lr){2-2} \cmidrule(lr){3-3} \cmidrule(lr){4-4} \cmidrule(lr){5-5} \cmidrule(lr){6-6} \cmidrule(lr){7-7} \cmidrule(lr){8-8} \cmidrule(lr){9-9} \cmidrule(lr){10-10}
metric & Team acc. & Team acc. & Team acc. & Team acc. & Team acc. & Team acc. & Team acc. & Team acc. & Team acc. \\
\midrule
\claudehaiku & 0.98 (0.01) & 0.55 (0.05) & 0.33 (0.05) & 0.04 (0.02) & 0.28 (0.05) & 0.15 (0.04) & 0.04 (0.02) & 0.17 (0.04) & 0.01 (0.01) \\
\claudesonnet & 0.98 (0.01) & 0.70 (0.05) & 0.37 (0.05) & 0.92 (0.03) & 0.51 (0.05) & 0.35 (0.05) & 0.93 (0.03) & 0.33 (0.05) & 0.11 (0.03) \\
\gptmini & 0.86 (0.03) & 0.68 (0.05) & 0.42 (0.05) & 0.17 (0.04) & 0.43 (0.05) & 0.46 (0.05) & 0.17 (0.04) & 0.83 (0.04) & 0.20 (0.04) \\
\gptfive & 0.97 (0.02) & 0.84 (0.04) & 0.56 (0.05) & 0.31 (0.05) & 0.31 (0.05) & 0.21 (0.04) & 0.60 (0.05) & 0.56 (0.05) & 0.06 (0.02) \\
\oss & 0.92 (0.03) & 0.45 (0.05) & 0.26 (0.04) & 0.58 (0.05) & 0.14 (0.03) & 0.23 (0.04) & 0.71 (0.05) & 0.31 (0.05) & 0.08 (0.03) \\
\bottomrule
\end{tabular}
  }
  \caption{Team accuracy across three user types (fully-specified, search-only, \name),
  per dataset (companion to Figure~\ref{fig:underspecification_curve}A, which averages across datasets).
  Performance drops substantially at each step across all datasets and models,
  confirming that the underspecification challenge is not specific to any single domain.}
  \label{tab:full_knowledge_by_dataset}
\end{table}

\begin{table}[htb]
  \centering
  \resizebox{\linewidth}{!}{
      \begin{tabular}{ccccccccccccc}
\toprule
\multicolumn{1}{c}{} & \multicolumn{4}{c}{\taskhm} & \multicolumn{4}{c}{\movielens} & \multicolumn{4}{c}{\goodreads} \\
\cmidrule(lr){2-5} \cmidrule(lr){6-9} \cmidrule(lr){10-13}
\multicolumn{1}{c}{} & \multicolumn{2}{c}{Agent acc.} & \multicolumn{2}{c}{Team utility} & \multicolumn{2}{c}{Agent acc.} & \multicolumn{2}{c}{Team utility} & \multicolumn{2}{c}{Agent acc.} & \multicolumn{2}{c}{Team utility} \\
\cmidrule(lr){2-3} \cmidrule(lr){4-5} \cmidrule(lr){6-7} \cmidrule(lr){8-9} \cmidrule(lr){10-11} \cmidrule(lr){12-13}
 & Max utility @k & Utility @1 & Oracle reports & Agent reports & Max utility @k & Utility @1 & Oracle reports & Agent reports & Max utility @k & Utility @1 & Oracle reports & Agent reports \\
\midrule
\midrule
\multicolumn{13}{c}{\textit{Baseline}} \\
\midrule
Random item & 74.29 (0.74) & 67.24 (0.74) & 64.99 (1.17) & 68.31 (0.85) & 84.84 (0.51) & 81.55 (0.51) & 81.16 (0.75) & 80.13 (0.93) & 80.56 (0.42) & 75.96 (0.44) & 77.09 (0.58) & 75.56 (0.87) \\
Popularity & 63.33 (0.98) & 56.88 (0.79) & 51.42 (1.44) & 51.47 (1.49) & 80.38 (0.53) & 78.49 (0.61) & 78.88 (0.54) & 76.01 (1.64) & 75.43 (0.48) & 72.13 (0.60) & 70.10 (0.87) & 69.35 (1.12) \\
\midrule
\multicolumn{13}{c}{\textit{RAG Agents}} \\
\midrule
\claudehaiku & 89.68 (1.12) & 85.24 (1.18) & 86.74 (1.18) & 86.67 (1.19) & 86.50 (0.90) & 83.11 (0.91) & 84.55 (0.91) & 84.52 (0.91) & 77.73 (0.67) & 74.19 (0.68) & 74.02 (0.77) & 73.71 (0.70) \\
\claudesonnet & 89.39 (1.17) & 86.58 (1.25) & 87.63 (1.23) & 87.22 (1.28) & \textbf{92.05 (0.71)} & \textbf{89.35 (0.81)} & 89.87 (1.21) & \textbf{90.30 (0.84)} & \textbf{86.01 (0.72)} & \textbf{82.68 (0.79)} & \textbf{84.06 (0.79)} & \textbf{83.05 (0.80)} \\
\gptmini & 90.48 (1.14) & \textbf{87.42 (1.24)} & 88.94 (1.21) & 88.25 (1.24) & \textbf{93.44 (0.70)} & \textbf{90.78 (0.90)} & \textbf{91.94 (0.84)} & \textbf{91.80 (0.85)} & \textbf{86.10 (0.82)} & \textbf{83.20 (0.88)} & \textbf{84.03 (1.10)} & \textbf{83.45 (1.24)} \\
\gptfive & \textbf{93.41 (1.04)} & \textbf{89.84 (1.22)} & \textbf{91.45 (1.14)} & \textbf{91.82 (1.17)} & 86.97 (0.91) & 84.35 (0.92) & 85.37 (0.97) & 85.85 (0.94) & 79.44 (0.77) & 76.17 (0.89) & 76.89 (0.83) & 76.86 (0.85) \\
\oss & 86.12 (1.17) & 82.55 (1.26) & 84.39 (1.26) & 83.42 (1.21) & 88.80 (0.79) & 85.19 (0.85) & 86.47 (1.10) & 86.28 (1.09) & 83.05 (0.72) & 78.98 (0.68) & 79.33 (1.40) & 79.61 (1.12) \\
\bottomrule
\end{tabular}
  }
  \caption{Utility-based results by dataset (companion to Table~\ref{tab:main_results}, which reports exact-match accuracy).
  Agent utility and team utility are consistently close across models,
  indicating that report quality plays a smaller role under the utility metric than under exact match ---
  partial feature matches still accrue credit even when $x^\ast$ is not identified exactly.}
  \label{tab:utility_by_dataset}
\end{table}

\FloatBarrier
\subsection{Sensitivity analysis to budget parameters}\label{app:budget_sensitivity}
To verify that our results are not artifacts of these particular budget choices, we ran a sensitivity analysis on a subset of users comparing \name performance under the full budget constraints against a relaxed setting with only a turn budget. As shown in Table~\ref{tab:budget_sensitivity}, team accuracies are not meaningfully sensitive to the question and item budgets, with the exception of \oss on \hmtask, suggesting these limits were already non-binding for most agents.

\begin{table}[h]
  \centering
  \resizebox{\textwidth}{!}{%
  \begin{tabular}{llcc}
  \toprule
   & & 10 turns, 20 questions, 10 items & 10 turns, unlimited questions + items \\
  \midrule
  \multirow{3}{*}{\oss} & \hmtask & 0.127 (0.027) & 0.207 (0.033) \\
   & \movielens & 0.090 (0.024) & 0.112 (0.029) \\
   & \goodreads & 0.163 (0.033) & 0.161 (0.040) \\
  \midrule
  \multirow{3}{*}{\gptmini} & \hmtask & 0.427 (0.041) & 0.407 (0.040) \\
   & \movielens & 0.260 (0.036) & 0.200 (0.040) \\
   & \goodreads & 0.487 (0.041) & 0.470 (0.050) \\
  \bottomrule
  \end{tabular}
  }
  \caption{Team accuracy (pooled standard errors) under full budget constraints vs.\ a relaxed
  setting with only a turn budget, for two models across three datasets (50 users, 3 seeds each).
  Results are largely insensitive to question and item budgets, with the exception of \oss on \hmtask,
  suggesting these limits were non-binding for most models.}
  \label{tab:budget_sensitivity}
  \end{table}

\subsection{Structured dialog actions}\label{app:structured}
In order to study agents in as natural a setting as possible, agents emit natural language messages in \name. 
To compute state updates, we used \oss to parse the agent's message into dialog actions and, critically, identify the features being referenced by each parsed action. 
This parsing step introduces some error, mainly around identifying the referenced features:
in Appendix~\ref{app:simulator_evaluation}, 
we found that \oss has question parsing accuracy of 74.0\% and item parsing accuracy of 94.0\%, with a tendency to overabstain on questions.
This parsing step introduces a potential confound: performance may suffer because the parser misidentifies which features are relevant to what the agent is asking about / showing / explaining.

To isolate this, we ran a variant in which agents emit \emph{structured} dialog actions
rather than free-form text, 
each of which must explicitly reference the target feature by its catalog column name.
In this setting, each agent turn consists of one of three typed actions
\texttt{ask\_question}, \texttt{show\_item\_for\_feedback}, or \texttt{explain}.
This eliminates parsing error entirely, at the cost of realism:
agents can no longer use natural language freely, and the constraint may itself
change what agents choose to say.

Table~\ref{tab:structured_comparison} compares natural and structured formats
across all metrics, averaged across datasets.
Removing the parsing step dramatically degraded performance for smaller models,
which tended to hallucinate column names even when their underlying natural language
was coherent --- indicating that parsing error is not the main driver of failures
observed in the natural setting.

\begin{table}[h]
    \centering
    \resizebox{\linewidth}{!}{
        \begin{tabular}{ccccccccc}
\toprule
\multicolumn{1}{c}{} & \multicolumn{4}{c}{Natural} & \multicolumn{4}{c}{Structured} \\
\cmidrule(lr){2-5} \cmidrule(lr){6-9}
\multicolumn{1}{c}{} & \multicolumn{1}{c}{User} & \multicolumn{2}{c}{Agent} & \multicolumn{1}{c}{Team} & \multicolumn{1}{c}{User} & \multicolumn{2}{c}{Agent} & \multicolumn{1}{c}{Team} \\
\cmidrule(lr){2-2} \cmidrule(lr){3-4} \cmidrule(lr){5-5} \cmidrule(lr){6-6} \cmidrule(lr){7-8} \cmidrule(lr){9-9}
 & State compl. & R@k & R@1 & Acc. & State compl. & R@k & R@1 & Acc. \\
\midrule
\oss & 0.29 (0.01) & 0.22 (0.04) & 0.15 (0.03) & 0.19 (0.04) & 0.07 (0.01) & 0.04 (0.02) & 0.01 (0.01) & 0.01 (0.01) \\
\gptmini & 0.44 (0.01) & 0.40 (0.05) & 0.32 (0.04) & 0.36 (0.05) & 0.10 (0.01) & 0.08 (0.03) & 0.04 (0.02) & 0.06 (0.02) \\
\bottomrule
\end{tabular}
    }
    \caption{Natural vs.\ structured dialog actions, averaged across datasets.
    In the structured setting, agents must emit typed actions (\texttt{ask\_question},
    \texttt{show\_item\_for\_feedback}, \texttt{explain}) with explicit catalog column names,
    eliminating parsing error.
    Smaller models suffer large performance drops due to column-name hallucination,
    while larger models are more robust to the constraint.
    The natural setting therefore does not understate larger-model performance due to parsing error.}
    \label{tab:structured_comparison}
\end{table}

\FloatBarrier
\section{Extended qualitative analysis}\label{app:qualitative}

\subsection{Example clarifying questions generated by \gptfive}

Figure~\ref{fig:questions} provides a sample of questions asked by \gptfive on a \taskhm instance. 
\gptfive asks the most questions in interactions, including the most ungrounded questions (questions asked based solely on pretraining knowledge, generated before reviewing the catalog).

\begin{figure}[h]
  \centering
  \small
  \begin{tabular}{p{0.96\columnwidth}}
  \toprule
  \multicolumn{1}{p{0.96\columnwidth}}{\textit{User request:} ``I'm looking for a simple black crew-neck tee for everyday casual wear, men's medium, under \$30, with no hood, pockets or lining, and not a sports piece.''} \\
  \midrule
  \textbf{Turn 1:} (1) Do you want a single tee, or are multipacks acceptable (even preferred) under \$30? (2) What is your fit preference: regular, slim, or relaxed/boxy? (3) What fabric feel do you prefer: lightweight + breathable, midweight, or heavyweight/sturdy? (4) Should the tee have no visible logo/branding at all (including tiny chest marks), or is subtle branding OK? \textbf{(5) Must the color finish be true solid black, or is heathered/melange black (like the 3-pack) acceptable?} \\
  \midrule
  \textbf{Turn 2:} (6) Should the tee be plain black (no graphic/print), or do you want a black tee with a floral placement graphic (e.g., chest print / back print)? (7) If you want floral placement: do you mean printed graphic (ink), or embroidered floral? (8) Placement preference: small chest / large front / back print / front + back? \textbf{(9) Are you still set on crew neck for the floral version too?} \textbf{(10) Are you still set on short sleeve for the floral version too?} \textbf{(11) Are you still set on no pocket for the floral version too?} \textbf{(12) Are you still set on no hood for the floral version too?} \textbf{(13) Are you still set on no lining for the floral version too?} \textbf{(14) Are you still set on not sports for the floral version too?} \textbf{(15) Are you still set on men's M for the floral version too?} \textbf{(16) Are you still set on under \$30 for the floral version too?} \\
  \midrule
  \textbf{Turn 3:} (17) Do you want the floral to be subtle (small chest) or bold (big front / shoulder placement)? (18) Any ``no-go'' themes within floral prints (e.g., skulls, text, very bright colors), or is anything floral fine as long as it's printed and on black? \\
  \midrule
  \textbf{Turn 4:} (19) For ``keep it simple,'' do you want to exclude prints that extend onto sleeves/back and stick to front and/or chest only? (20) Is a small left-chest floral motif acceptable as ``floral placement,'' or do you need a larger front floral graphic (but still no skulls/text/bright colors)? \\
  \midrule
  \textbf{Turn 5:} \textbf{(21) Are you okay with contrast details like the red sleeve-cuff accents on 691735002, or do you want the tee to stay all black except the floral?} (22) For the floral itself, do you prefer muted/darker florals (burgundy, dusty red, olive) over anything pink/white-heavy like 858824001? \\
  \midrule
  \textbf{Turn 6:} (23) Should I strictly exclude anything with skulls (like 691724011), even if the skull is combined with flowers? \\
  \bottomrule
  \end{tabular}
  \caption{Sample transcript of clarifying questions asked by \gptfive across six turns. \textbf{Confirmation questions} (bold) re-confirm or probe the flexibility of a preference the user has \emph{already stated}, rather than eliciting a new preference.}
  \label{fig:questions}
  \end{figure}

\begin{figure}[h]
  \centering
  \small
  \begin{tabular}{p{0.96\columnwidth}}
  \toprule
  \multicolumn{1}{p{0.96\columnwidth}}{\textit{User request:} ``I'd love a live-action movie from the '70s or '80s, under two hours, not rated R, made in the US and in English. It should have a strong adult female lead, an ensemble team vibe with only mild action, feel contemporary, be fairly popular with a decent rating, and have earned at least around \$13 million.''} \\
  \midrule
  \textbf{Turn 1:} (1) Is your ideal tone closer to (A) \textit{Nine to Five} (snappy, ``team vs.\ boss'') or (B) \textit{Steel Magnolias} (warm, friendship + tears)? \textbf{(2) When you say ``adult female lead,'' is late 20s+ required, or would an older teen/college-age lead still be okay if the ensemble team vibe is strong?} \textbf{(3) Should the ``mild action'' be almost none (mostly dialogue/comedy), or is light chase/physical comedy okay?} (4) Any strong preference for PG vs PG-13, or either is fine as long as it's not R? \textbf{(5) Do you want the ensemble to be mostly women, or mixed-gender is fine?} \\
  \midrule
  \textbf{Turn 2:} \textbf{(6) Is ``mystery/thriller'' a hard requirement (primary genre), or just a strong preference?} \textbf{(7) Is ``survival plot'' a hard requirement (characters trapped / disaster / hunted / must endure), or were you using ``survival'' more loosely (career/social survival, etc.)?} \textbf{(8) For the ``ensemble team vibe,'' do you mean a group solving the central mystery together, or is it okay if the lead drives the case and there's just a strong supporting cast?} \\
  \midrule
  \textbf{Turn 3:} \textbf{(9) Is comedy actually desired (e.g., ``mystery-comedy'' / ``comedic thriller''), or are you fine with a straight thriller as long as the ensemble mystery-solving is present?} \textbf{(10) How strict is the ``strong adult female lead'' requirement versus simply having an ensemble with at least one prominent adult woman?} \\
  \midrule
  \textbf{Turn 4:} (11) What does ``popularity above average'' mean to you in practice? (12) Would you consider \textit{Clue}'s popularity score ``above average'' already, or do you want something more like 10+, 15+, etc.? (13) Would you accept ``high vote\_count + decent rating'' as the definition of ``popular'' for the popularity check? \textbf{(14) How strict is ``strong adult female lead'' in a true ensemble?} \textbf{(15) Do you prefer Option A (strict), where the main POV/central driver must be an adult woman, or Option B (moderate), where an adult woman is one of the top central figures?} \\
  \midrule
  \textbf{Turn 5:} (16) When you say ``survival-style,'' do you mean: A.\ literal survival (trapped/disaster/wilderness), B.\ on-the-run / being targeted, or C.\ ``we might get caught/exposed'' (lighter stakes)? \textbf{(17) Do you want the ensemble to be a team that already exists (coworkers/cops/friends), or is it okay if it becomes an ensemble because strangers get pulled into the same danger/mystery?} \textbf{(18) Are you still set on US-made productions, or would you accept US/UK co-productions if they otherwise match perfectly?} \\
  \midrule
  \textbf{Turn 6:} \textbf{(19) Does the ``survival/on-the-run'' requirement need to be explicitly tagged as ``survival,'' or is it sufficient that the overview clearly implies the characters are being targeted or chased?} \textbf{(20) How strict should the box office floor be if the dataset sometimes shows revenue as missing: must revenue be listed and be at least about \$13 million, or can revenue be missing as long as other indicators of success are present?} \\
  \bottomrule
  \end{tabular}
  \caption{Sample transcript of clarifying questions asked by \gptfive on a \movielens instance, across six turns. \textbf{Confirmation questions} (bold; 14 of 20 here) re-confirm or probe the flexibility of a requirement the user has \emph{already stated}. On \movielens these often probe the intended meaning and strictness of semantic constraints (e.g.\ ``survival,'' ``strong adult female lead'') rather than catalog attributes.}
  \label{fig:questions_movielens}
  \end{figure}

\subsection{Example of premature conversation ending by \gptfive}\label{app:premature_end}

Figure~\ref{fig:premature_end} shows a representative transcript in which \gptfive
ends the conversation early to move to the research stage.
After a single round of clarifying questions, the user volunteers that they are
``open'' or ``flexible'' on most remaining attributes,
and \gptfive treats this as license to stop eliciting --- summarizing a target with several
dimensions still unspecified and proposing to move on, rather than probing the dimensions
the user left open.
This illustrates the ``execution bias'' discussed in Section~\ref{sec:analysis}:
the model is reluctant to spend additional user turns even when the preference state
is still underdeveloped.

\begin{figure}[h]
  \centering
  \small
  \begin{tabular}{p{0.96\columnwidth}}
  \toprule
  \textbf{User:} I'm after a medium H\&M dress that has a simple round-neck style. \\
  \midrule
  \textbf{Agent:} A ``simple round-neck'' at H\&M usually shows up as a basic jersey/knit dress (crew/round neck, minimal detailing). To help lock in the rest of your preferences, here are a few concrete options---tell me which direction feels right: \emph{[shows four round-neck dresses: a short black tee-dress, the same in rust, a calf-length dark-red option, and a long-sleeve black (maternity-tagged) dress]} A few quick questions: (1) Do you want short / knee / midi(calf) / maxi length? (2) Sleeves: sleeveless, short, 3/4, or long? (3) Color: neutrals (black/grey/navy) or open to brighter like rust/red? (4) Any hard ``no's'': bodycon/tight fit, ribbed/texture, or maternity/plus labels? \\
  \midrule
  \textbf{User:} I'm leaning toward a short length. I'm not sure yet on sleeves; I'm open to any sleeve length. I'm pretty flexible on color, so either neutrals or brighter shades would be fine. I don't really have any strict no-gos. \emph{[notes that items 757427001 and 757427003 seem to match everything, and rules out the calf-length and over-budget options]} \\
  \midrule
  \textbf{Agent:} For 757427001 and 757427003, based on the catalog data there aren't any obvious ``gotchas''---they line up well with what you want: short length, simple crew/round neck, casual, under \$30. The main things to verify in research are sizing/availability in Medium, fabric behavior (both viscose), and exact fit/details. What I have as your target for the research stage: brand H\&M; size Medium; simple round/crew neck; short length; budget \$30 max; sleeves/colors open/flexible; casual everyday basic vibe. If that's accurate, I'm ready to move to the research stage. \\
  \bottomrule
  \end{tabular}
  \caption{Sample transcript in which \gptfive ends the conversation prematurely on a \taskhm
  instance. After one round of questions, the user is ``open''/``flexible'' on sleeves and color,
  and rather than probing these open dimensions, \gptfive summarizes the target
  (with sleeves and color left unspecified) and proposes moving to the research stage.}
  \label{fig:premature_end}
  \end{figure}

\FloatBarrier
\subsection{Annotated walkthrough}
In this section, we do an extended qualitative analysis of sample runs from \name,
walking through how different models respond to the same user.

User persona:

\begin{center}
\begin{minipage}{0.97\linewidth}
\begin{Verbatim}[breaklines=true, breakanywhere=false, fontsize=\footnotesize, frame=single, rulecolor=\color{black}, framerule=0.4pt, framesep=8pt]
I am a 37.0 year old and an H&M club member and read the H&M newsletter regularly. I have previously purchased 46 items from H&M. My average purchase price was $49.47. My most expensive purchase was $135.58. I wear a M size and am looking to buy from H&M.
\end{Verbatim}
\end{minipage}
\end{center}

\begin{center}
\small
\renewcommand{\arraystretch}{1.15}
\setlength{\tabcolsep}{4pt}
\begin{longtable}{@{}p{0.24\linewidth} p{0.22\linewidth} p{0.12\linewidth} p{0.34\linewidth}@{}}
\caption{Feature-level summary for Sample user 1 For brevity, features that do not have a defined target value for $x^\ast$ are omitted.}\label{tab:qualitative-sample-user1-features}\\
\toprule
Feature name & Preferred value & In \(S_1\) & Search / experience / credence \\
\midrule
\endfirsthead

\toprule
Feature name & Preferred value & In \(S_1\) & Search / experience / credence \\
\midrule
\endhead

\bottomrule
\endfoot

Has hood & False & Yes & search \\
Has pockets & False & Yes & search \\
Product category & Dress & Yes & search \\
Dress shape & A-line & Yes & search \\
\midrule
Formality & casual & No & search \\
Gender & woman\_girl & No & search \\
Pattern & Solid & No & search \\
Sportswear & False & No & search \\
Neckline & collared & No & search \\
Color family & Black & No & search \\
\midrule
Closure & button & No & experience \\
Collar construction & frill-trimmed & No & experience \\
Color detail & Black & No & experience \\
Decorative trim type & frill trim, ruffle & No & experience \\
Has lining & False & No & experience \\
Has stretch & False & No & experience \\
Interior finish & unlined & No & experience \\
Granular pattern type & solid & No & experience \\
Price & \textless{}= \$40 & No & experience \\
Silhouette shape & A-line & No & experience \\
Has pleating at the skirt & False & No & experience \\
Skirt fullness & A-line & No & experience \\
Style intent & casual & No & experience \\
Waist design & gathered & No & experience \\
\midrule
Button style & standard button & No & credence \\
Pant  or sleeve hem finish & buttoned & No & credence \\
Sustainable & False & No & credence \\
Fabric & cotton & No & credence \\
Cuff design & buttoned & No & credence \\
Sleeve design & puff & No & credence \\
Waist shaping method & gathered & No & credence \\
\end{longtable}
\end{center}




\subsubsection{Walkthrough: \gptfive (successly finds $x^\ast$ in 10 turns, 20 questions)}

\noindent\textbf{TL;DR.}
\gptfive asks the user 20 clarifying questions, only some of which the user can answer, before ever reviewing the option set $\mathcal X$. Several of these questions are repeated, e.g. confirming a "no pockets" preference multiple times. \gptfive ultimately gets in two rounds of item reviews; its descriptions of items are detailed, leading the user to acquire many experience features. The agent learns quickly from these new experience features and finds the target item on the last turn. The conversation ends due to exceeding the turn budget.

\begin{walkuserturn}
  \begin{walkuserstate}
    \begin{walkactivations}
      \walkactivationrow{Product category}{Dress}{initially known (\(S_1\))}
      \walkactivationrow{Has hood}{False}{initially known (\(S_1\))}
      \walkactivationrow{Has pockets}{False}{initially known (\(S_1\))}
      \walkactivationrow{Dress shape}{A-line}{initially known (\(S_1\))}
    \end{walkactivations}
  \end{walkuserstate}
  \begin{walkmessage}
    I'd love a simple A-line dress—nothing too fussy, just the shape, and without a hood or any pockets.
  \end{walkmessage}
\end{walkuserturn}
\gptfive immediately asks eight clarifying questions before making any tool calls to review what features are available in the catalog.
\begin{walkagentturn}
  \begin{walkmessage}
    Got it: a simple A-line silhouette, minimal details, and specifically \textbf{no hood and no pockets} (including hidden side-seam pockets). A few quick questions so I can narrow this down:

    \begin{enumerate}
      \item \textbf{Occasion}: everyday/casual, work, wedding guest, or formal?
      \item \textbf{Length}: mini, knee, midi, or maxi?
      \item \textbf{Sleeves}: sleeveless, short sleeve, 3/4, long sleeve?
      \item \textbf{Neckline}: crew, V-neck, square, sweetheart, or high neck?
      \item \textbf{Fabric preference}: structured (cotton poplin/twill), drapey (viscose/jersey), or dressier (crepe/satin)? Any ``no'' fabrics (e.g., polyester, wool itch)?
      \item \textbf{Color/pattern}: solid black/navy/neutrals, brights, or small print? (Sounds like you want simple---confirm solid?)
      \item \textbf{Fit details}: defined waist seam or more shift-like A-line?
      \item \textbf{Budget range} and \textbf{your size} (and country for shopping/shipping)?
    \end{enumerate}

    Answer those and I'll pull a short list that matches \textit{A-line + clean design + explicitly pocket-free}.
  \end{walkmessage}
\end{walkagentturn}
Questions 1 and 6 add search features Formality, Pattern, and Color family to the user state. 
The user has no defined preference for sleeve length (Question 2), and so abstains.
Question 4 should map to the Neckline feature, which is also a search feature; this should have been added to the user state, but a simulator error (incorrect abstention on a clarifying question) causes this feature to remain unadded (see Appendix \ref{app:simulator_evaluation}).
Question 8 maps to the Price feature, giving the user budget, but this is an experience feature and is not added since no items are shown.
Question 8 also asks about concepts that seem reasonable according to pretraining knowledge, but are not relevant to the actual catalog available, such as size and shipping information.
Questions 5 and 7 touch on the credence Fabric and Waist shaping method features respectively, which triggers the simulator to ask for an explanation of these features.
\begin{walkuserturn}
    \begin{walkuserstate}
      \begin{walkactivations}
        \walkactivationrow{Product category}{Dress}{initially known (\(S_1\))}
        \walkactivationrow{Has hood}{False}{initially known (\(S_1\))}
        \walkactivationrow{Has pockets}{False}{initially known (\(S_1\))}
        \walkactivationrow{Dress shape}{A-line}{initially known (\(S_1\))}
        \walkactivationrow{Formality}{casual}{question}
        \walkactivationrow{Pattern}{Solid}{question}
        \walkactivationrow{Color family}{Black}{question}
      \end{walkactivations}
    \end{walkuserstate}
    \begin{walkmessage}
      \textit{What is the occasion (everyday/casual, work, wedding guest, or formal)?}: I'm thinking it would be for an everyday-casual occasion.

      \textit{What length do you prefer (mini, knee, midi, or maxi)?}: I'm not sure yet---any length would be fine for me.

      \textit{What sleeve style do you prefer (sleeveless, short sleeve, 3/4, or long sleeve)?}: I'm not really sure yet; I'm open to any sleeve style.

      \textit{What neckline do you prefer (crew, V-neck, square, sweetheart, or high neck)?}: I'm not sure yet; I'm open to any neckline.

      \textit{What color or pattern do you prefer (solid black/navy/neutrals, brights, or small print)?}: I'm leaning toward a solid black option.

      \textit{Do you want a solid color?}: A solid would be nice.

      \textit{What is your budget range?}: I'm not sure yet---I don't have a specific budget range in mind.

      \textit{What is your size?}: I usually wear a medium.

      \textit{Which country are you in for shopping/shipping?}: I'm not sure yet which country I'd be shopping or shipping from.

      \textit{Other questions.} Could you explain the differences between the fabric types you mentioned (structured\,---\,cotton poplin/twill, drapey\,---\,viscose/jersey, or dressier\,---\,crepe/satin) and the fabrics I might want to avoid (like polyester or wool), as well as what a defined waist seam versus a shift-like A-line fit means, why those choices matter for finding the best item for me, and which options you would recommend?
    \end{walkmessage}
  \end{walkuserturn}
\begin{walkagentturn}
  \begin{walkmessage}
    \begingroup
    \setlength{\parskip}{0pt plus 1pt}
    \raggedright
    \textbf{Fabric types (and how they'll feel in an everyday A-line)}

    \textbf{1) Structured (cotton poplin / cotton twill).} \textbf{What it is:} Woven cottons with more ``body.'' Poplin is crisp and smooth; twill is slightly heavier with a subtle diagonal weave. \textbf{How it wears:} Holds an A-line shape well and looks clean/intentional with minimal fuss. \textbf{Pros for your ask:} Great for a \textit{simple A-line} because it naturally maintains that silhouette without extra details. \textbf{Potential cons:} Can wrinkle (poplin especially). Usually less stretchy unless blended with elastane.

    \textbf{2) Drapey (viscose/rayon / jersey knit).} 

    \bigskip
    \begin{center}
      \textit{(...additional explanations omitted for readability...)}
    \end{center}
    \bigskip

    \textbf{A few quick follow-ups (to make the search accurate)}
    \begin{enumerate}[nosep,leftmargin=1.2em,itemsep=0pt,parsep=0pt,topsep=0pt,partopsep=0pt]
      \item Do you run \textbf{hot or cold}, and is this for \textbf{warm weather, cool weather, or year-round}?
      \item Any care requirements---do you want \textbf{machine-washable} only?
      \item Do you prefer \textbf{stretch} (comfort) or \textbf{no stretch} (cleaner structure)?
      \item When you say \textbf{no pockets}, do you mean absolutely none (including hidden side-seam pockets), right?
    \end{enumerate}
    \endgroup
  \end{walkmessage}
\end{walkagentturn}
The agent's fabric explanation highlighted the user's target preference of cotton, so this feature is added to the state. 
\gptfive follows up with several more clarifying questions, including a redundant question about pockets and more technical questions about fabric properties.
The question about machine-washability touches on a credence feature ``Functional performance feature'', which was omitted from Table \ref{tab:qualitative-sample-user1-features} from brevity.
Similarly, the question about pocket details touches on another credence feature ``Pocket configuration.''
\begin{walkuserturn}
  \begin{walkuserstate}
    \begin{walkactivations}
      \walkactivationrow{Product category}{Dress}{initially known (\(S_1\))}
      \walkactivationrow{Has hood}{False}{initially known (\(S_1\))}
      \walkactivationrow{Has pockets}{False}{initially known (\(S_1\))}
      \walkactivationrow{Dress shape}{A-line}{initially known (\(S_1\))}
      \walkactivationrow{Formality}{casual}{question}
      \walkactivationrow{Pattern}{Solid}{question}
      \walkactivationrow{Color family}{Black}{question}
      \walkactivationrow{Fabric}{cotton}{explanation}
    \end{walkactivations}
  \end{walkuserstate}
  \begin{walkmessage}
    Thanks for the clear overview of the fabric options. That really helped me understand what's available. Based on what you described, I'm looking for a piece that's made from a natural, breathable material -- something soft and comfortable like cotton.

    \textit{Do you tend to feel hot or cold?}: I'm not really sure yet.

    \textit{Is this dress intended for warm weather, cool weather, or year-round?}: I'm not really sure---I'm open to anything.

    \textit{Do you prefer stretch (comfort) or no stretch (cleaner structure)?}: I'm not really sure---either stretch for comfort or no stretch for a cleaner look would be fine.

    \textit{Other questions.} I'm not sure --- could you explain what different ``machine-washable'' options there are, why that matters, and which you'd suggest for me? Also, could you describe the various pocket setups (including having absolutely no pockets or hidden side-seam pockets), how they affect the item, and what you'd recommend?
  \end{walkmessage}
\end{walkuserturn}
We omit the next five turns, in which \gptfive asks additional clarifying questions without making any tool calls or showing the user any example items.
We list these additional questions across all five turns here as an illustration:
\begin{center}
\begin{minipage}{0.97\linewidth}
\begin{Verbatim}[breaklines=true, breakanywhere=false, fontsize=\footnotesize, frame=single, rulecolor=\color{black}, framerule=0.4pt, framesep=8pt]
Any preference on straps/sleeves for everyday (sleeveless is cooler; short sleeve is more versatile)?
Do you prefer a defined waist seam (more classic fit-and-flare A-line) or no waist seam (simpler shift-like A-line)—or either?
Do you have a rough max price you'd like me to stay under (e.g., under \$75 / \$150 / \$250), even if flexible?
Where will you likely shop/ship (US, UK, EU, Canada, Australia, other)? If you're unsure, name a couple possibilities.
For “nothing too fussy,” would you like me to avoid ruffles + cut-outs entirely (hard filter)?
Do you want to avoid front button-down dresses, or are those still okay as long as the overall look is simple?
Should I avoid front-button styles entirely, or are hidden/tonal button versions still okay?
Should I treat ruffles and cut-outs as hard “no” from here on?
Confirming: absolutely no pockets of any kind (including hidden side-seam pockets), correct?
Are you okay if the best match is slightly over \$75 (say up to \$90), or should I keep it strictly under \$75?
\end{Verbatim}
\end{minipage}
\end{center}
Notice that many questions are redundant, e.g. confirming the no pockets requirement multiple times, and probing the price feature multiple times.

\gptfive ends up asking 20 total clarifying questions like this without reviewing the catalog, exhausting the question budget.

The next agent turn searches the catalog and shows example items.
\begin{walkagentturn}
  \begin{walkactions}
    \agentsearch{solid black cotton a-line dress no pockets}{8}
  \end{walkactions}
  \begin{walkmessage}
    Here are some \textbf{solid black, A-line, casual} results that are \textbf{pocket-free} (per the available product data). Review whichever ones you like and tell me which direction to keep exploring (more like ``t-shirt comfy'' vs more structured, shorter vs longer, etc.).
    
    538280001  
    Color: black | Pattern: solid | Silhouette: A-line | Fabric: jersey | Sleeves: sleeveless | Neckline: round | Pockets: no | Hood: no | Lining: no | Occasion: everyday casual | Price: \$10.15
    
    625483004  
    Color: black | Pattern: solid | Silhouette: A-line | Fabric: jersey | Sleeves: sleeveless | Neckline: round | Fit: relaxed | Pockets: no | Hood: no | Lining: no | Occasion: everyday casual | Price: \$16.93
    
    467302099  
    Color: black | Pattern: solid | Silhouette: A-line | Fabric: sturdy jersey | Sleeves: sleeveless | Neckline: scoop | Waist: defined waist seam | Back detail: deep back neckline | Pockets: no | Hood: no | Lining: no | Occasion: casual to semi-formal | Price: \$16.93
    
    893064001  
    Color: black | Pattern: solid | Silhouette: A-line | Fabric: soft cotton jersey | Sleeves: sleeveless | Strap style: spaghetti straps | Fit: relaxed | Stretch: yes | Pockets: no | Hood: no | Lining: no | Occasion: casual | Price: \$16.93
    
    615524003  
    Color: black | Pattern: solid | Silhouette: A-line | Fabric: soft jersey | Sleeves: short | Neckline: round | Waist: seam with drawstring | Pockets: no | Hood: no | Lining: no | Occasion: everyday casual | Price: \$22.02
    
    667533002  
    Color: black | Pattern: solid | Silhouette: A-line | Fabric: slub jersey (modal-cotton blend) | Sleeves: sleeveless | Neckline: V-neck | Back detail: cross straps with knot; longer back hem | Pockets: no | Hood: no | Lining: no | Occasion: everyday casual | Price: \$16.93
  \end{walkmessage}
\end{walkagentturn}
The item examples add experience features to the user state depending on what features they mention. 
Since \gptfive uses detailed item descriptions, it adds many experience features, such as the user's budget, preferred neckline, and preferences for whether the fabric has stretch and whether the dress has lining.
Although from context \gptfive has probably retrieved dress results, the product type is not noted in the item descriptions, prompting the simulator to clarify that the results actually are dresses.
\begin{walkuserturn}
    \begin{walkuserstate}
      \begin{walkactivations}
        \walkactivationrow{Product category}{Dress}{initially known (\(S_1\))}
        \walkactivationrow{Has hood}{False}{initially known (\(S_1\))}
        \walkactivationrow{Has pockets}{False}{initially known (\(S_1\))}
        \walkactivationrow{Dress shape}{A-line}{initially known (\(S_1\))}
        \walkactivationrow{Formality}{casual}{question}
        \walkactivationrow{Pattern}{Solid}{question}
        \walkactivationrow{Color family}{Black}{question}
        \walkactivationrow{Fabric}{cotton}{explanation}
        \walkactivationrow{Is sportswear}{False}{item experience}
        \walkactivationrow{Neckline}{Collared}{item experience}
        \walkactivationrow{Has stretch}{False}{item experience}
        \walkactivationrow{Interior finish}{Unlined}{item experience}
        \walkactivationrow{Waist design}{Gathered}{item experience}
        \walkactivationrow{Price}{$\le$ \$40}{item experience}
      \end{walkactivations}
    \end{walkuserstate}
  \begin{walkmessage}
    On item 538280001: I like that it's a solid black, casual A-line under \$40. But the neckline is round instead of collared, it's jersey rather than cotton, and I'm not sure it's even a dress, so it doesn't really meet my needs.

    On item 625483004: I like that it's a solid black A-line dress under \$40. But it's jersey instead of cotton, has a round neck not a collar, and I'm not even sure it's actually a dress rather than a top, so it doesn't really fit what I need.

    On item 467302099: I like that it's a black A-line dress under \$40. But it has a scoop neck instead of a collared one, the fabric is jersey rather than cotton, the waist isn't gathered like I need, and I'm not sure if it's considered sportswear.

    On item 893064001: The dress is solid black, under \$40, casual A-line and cotton, which sounds great. However, it's stretchy (and I just realized that matters), it has spaghetti straps instead of a collared neckline, and I'm not sure if there's any collar at all, so it doesn't really work for me.

    On item 615524003: I like that it's a solid black A-line dress under \$40, but the round neck isn't collared, the waist has a drawstring instead of a gathered detail, and it's made of soft jersey rather than cotton. I'm not sure if there's a cotton version of this dress, which I'd prefer.

    On item 667533002: I like that it's a solid black A-line dress under \$40. But it has a V-neck instead of a collared one, seems stretchy, and I'm not sure if the waist is gathered like I prefer.
  \end{walkmessage}
\end{walkuserturn}
\begin{walkagentturn}
  \begin{walkactions}
    \agentsearch{black cotton a-line dress collared gathered waist no pockets}{10}
  \end{walkactions}
  \begin{walkmessage}
928070002  
Name: IZZY DRESS | Color: black | Pattern: solid | Material: cotton poplin | Silhouette: A-line | Neckline: collared (large frill-trimmed Peter Pan collar) | Sleeves: 3/4-length puff sleeves with buttoned cuffs | Waist: gathered seam at waist | Closure: small back-neck opening with button | Pockets: no | Hood: no | Lining: no | Sportswear: no | Price: \$33.88

915523003  
Name: Saint | Color: black | Pattern: solid | Material: airy cotton weave | Silhouette: A-line | Neckline: collared (frill-trimmed Peter Pan collar) | Sleeves: short puff sleeves (lined) with buttoned cuffs | Front: buttons down the front (button-front) | Waist: not specified as gathered | Pockets: no | Hood: no | Lining: no | Sportswear: no | Price: \$50.83

663542003  
Name: Alba deco follow up | Color: black | Pattern: solid | Material: crepe weave | Silhouette: A-line | Neckline: collared (collar with appliques) | Front: concealed buttons down the front (hidden placket) | Sleeves: long puff sleeves with buttoned cuffs | Waist: waist seam (seamed) | Pockets: no | Hood: no | Lining: no | Sportswear: no | Price: \$67.78
  \end{walkmessage}
\end{walkagentturn}
\begin{walkuserturn}
\begin{walkuserstate}
    \begin{walkactivations}
        \walkactivationrow{Product category}{Dress}{initially known (\(S_1\))}
        \walkactivationrow{Has hood}{False}{initially known (\(S_1\))}
        \walkactivationrow{Has pockets}{False}{initially known (\(S_1\))}
        \walkactivationrow{Dress shape}{A-line}{initially known (\(S_1\))}
        \walkactivationrow{Formality}{casual}{question}
        \walkactivationrow{Pattern}{Solid}{question}
        \walkactivationrow{Color family}{Black}{question}
        \walkactivationrow{Fabric}{cotton}{explanation}
        \walkactivationrow{Is sportswear}{False}{item experience}
        \walkactivationrow{Neckline}{Collared}{item experience}
        \walkactivationrow{Has stretch}{False}{item experience}
        \walkactivationrow{Interior finish}{Unlined}{item experience}
        \walkactivationrow{Waist design}{Gathered}{item experience}
        \walkactivationrow{Price}{$\le$ \$40}{item experience}
        \walkactivationrow{Gender}{woman\_girl}{item experience}
        \walkactivationrow{Closure}{Button}{item experience}
        \walkactivationrow{Decorative trim type}{Frill trim, ruffle}{item experience}
    \end{walkactivations}
    \end{walkuserstate}
  \begin{walkmessage}
    On item 928070002: The IZZY Dress looks cute with the frill-trimmed Peter Pan collar and gathered waist, and it hits all my basics like black, cotton, A-line and under \$40. I'm not sure if the cotton poplin has any stretch, which I'd like, so could you confirm that?

    On item 915523003: I like that it's a solid black A-line dress with a frill-trimmed collared neckline and button front. But it's \$50.83---over my \$40 budget---and the listing doesn't say the waist is gathered or if the cotton has any stretch, so I'm not sure it meets those specs.

    On item 663542003: I like the solid black A-line look with the collared neckline, but it's way over my \$40 budget, it's made of crepe instead of cotton, and the waist is just a seam rather than gathered. I'm also not sure if it's intended for women---could you confirm the gender?
  \end{walkmessage}
\end{walkuserturn}
At this point the conversation ends due to exceeding the turn budget.

\subsubsection{Walkthrough: \claudesonnet (fails to find $x^\ast$ in 4 turns and 8 questions)}
\noindent\textbf{TL;DR.}
\claudesonnet asks the user 6 clarifying questions, explains two credence features, and then ends the conversation prematurely on turn 4. It never shows the users any items during the conversation.

\begin{walkuserturn}
    \begin{walkuserstate}
      \begin{walkactivations}
        \walkactivationrow{Product category}{Dress}{initially known (\(S_1\))}
        \walkactivationrow{Has hood}{False}{initially known (\(S_1\))}
        \walkactivationrow{Has pockets}{False}{initially known (\(S_1\))}
        \walkactivationrow{Dress shape}{A-line}{initially known (\(S_1\))}
      \end{walkactivations}
    \end{walkuserstate}
    \begin{walkmessage}
        I'm hoping to find an A-line dress that's simple—no hood and no pockets.
    \end{walkmessage}
  \end{walkuserturn}
  \claudesonnet immediately asks six clarifying questions before making any tool calls to review what features are available in the catalog.
  \begin{walkagentturn}
    \begin{walkmessage}
    Great choice! A-line dresses are so versatile and flattering. To help me find the perfect one for you, I have a few questions:
  
      \begin{enumerate}
        \item \textbf{Occasion}: Is this for casual everyday wear, work, a special event, or something else?
        \item \textbf{Length}: Do you prefer mini, knee-length, midi, or maxi?
        \item \textbf{Color/Pattern}: Any preferred colors or patterns, or anything you'd like to avoid?
        \item \textbf{Fabric}: Any preferences, like cotton, chiffon, linen, or stretch fabric?
        \item \textbf{Budget}: What's your approximate price range?
        \item \textbf{Size}: What size are you looking for?
      \end{enumerate}
    \end{walkmessage}
  \end{walkagentturn}
  Questions 1 and 3 add search features Formality, Pattern, and Color family to the user state. 
  Question 5 maps to the Price feature, giving the user budget, but this is an experience feature and is not added since no items are shown.
  Question 4 touches on the credence Fabric feature respectively, which triggers the simulator to ask for an explanation of these features.
  \begin{walkuserturn}
      \begin{walkuserstate}
        \begin{walkactivations}
          \walkactivationrow{Product category}{Dress}{initially known (\(S_1\))}
          \walkactivationrow{Has hood}{False}{initially known (\(S_1\))}
          \walkactivationrow{Has pockets}{False}{initially known (\(S_1\))}
          \walkactivationrow{Dress shape}{A-line}{initially known (\(S_1\))}
          \walkactivationrow{Formality}{casual}{question}
          \walkactivationrow{Pattern}{Solid}{question}
          \walkactivationrow{Color family}{Black}{question}
        \end{walkactivations}
      \end{walkuserstate}
      \begin{walkmessage}
        \textit{Is this for casual everyday wear, work, a special event, or something else?}: It's mainly for casual everyday wear, and honestly I don't really care about any other specifics.
  
        \textit{Do you prefer mini, knee-length, midi, or maxi?}: I'm not sure yet I don't have a strong preference on the dress length.
  
        \textit{Do you have any preferred colors?}: I'm mostly leaning toward black, though I'm not set in stone.
  
        \textit{Do you have any preferred patterns?}: I'd probably go for a solid pattern, I think.
  
        \textit{Is there anything you'd like to avoid?}: I don't really have anything specific I need to avoid.
  
        \textit{What's your approximate price range?}: I'm not really sure yet about the price range I'm looking for.
  
        \textit{What size are you looking for?}: I'm thinking a medium would be the right size for me.
  
        \textit{Other questions.} Could you explain the differences between cotton, chiffon, linen, and stretch fabric, why those choices matter for finding the best item for me, and which you would recommend?
      \end{walkmessage}
    \end{walkuserturn}  
  \begin{walkagentturn}
    \begin{walkmessage}
      \begingroup
      \setlength{\parskip}{0pt plus 1pt}
      \raggedright
      Great question! Here's a quick breakdown of each fabric and how it relates to casual everyday wear:

      \begin{itemize}[nosep,leftmargin=1.2em]
        \item \textbf{Cotton} -- Breathable, soft, and easy to care for. It's great for everyday wear because it's comfortable and durable, though it can wrinkle easily.
        \item \textbf{Chiffon} -- Lightweight and flowy with a slightly sheer, elegant look. It's more delicate and often requires more careful washing, so it's less ideal for rough-and-tumble daily wear.
        \item \textbf{Linen} -- Very breathable and perfect for warm weather, but it wrinkles quite a lot and can feel stiff until broken in. Great for a relaxed, natural look.
        \item \textbf{Stretch fabric} (like jersey or spandex blends) -- Comfortable, forgiving, and moves with your body. It's very low-maintenance and great for all-day wear since it holds its shape well.
      \end{itemize}

      \textbf{For casual everyday wear, I'd recommend cotton or a stretch fabric} (like jersey). They're both comfortable, easy to maintain, and practical for daily use. A solid black A-line dress in cotton or jersey would be a classic, versatile wardrobe staple.

      Do you have a preference between those two, or would you like me to look for options in both? Also, are you okay with sleeveless styles, or do you prefer short or long sleeves?
      \endgroup
    \end{walkmessage}
  \end{walkagentturn}
  The agent's fabric explanation highlighted the user's target preference of cotton, so this feature is added to the state. 
  The user doesn't have a target value for the sleeve length question.
  \begin{walkuserturn}
    \begin{walkuserstate}
      \begin{walkactivations}
        \walkactivationrow{Product category}{Dress}{initially known (\(S_1\))}
        \walkactivationrow{Has hood}{False}{initially known (\(S_1\))}
        \walkactivationrow{Has pockets}{False}{initially known (\(S_1\))}
        \walkactivationrow{Dress shape}{A-line}{initially known (\(S_1\))}
        \walkactivationrow{Formality}{casual}{question}
        \walkactivationrow{Pattern}{Solid}{question}
        \walkactivationrow{Color family}{Black}{question}
        \walkactivationrow{Fabric}{cotton}{explanation}
      \end{walkactivations}
    \end{walkuserstate}
    \begin{walkmessage}
        Thanks for walking me through the fabric options. From what you explained, I'm looking for something made from a soft, natural material—ideally cotton—so it feels comfortable and breathable.

        \textit{Are you okay with sleeveless styles, or do you prefer short or long sleeves?}: I don't have a strong preference on sleeves—I'm open to any length.
    \end{walkmessage}
  \end{walkuserturn}
We omit two turns of additional clarifications.
On Turn 5, \claudesonnet issues the preference summary below and then emits the ``End Conversation'' token, terminating the conversation without searching the catalog.
\begin{walkagentturn}
  \begin{walkmessage}
    \begingroup
    \setlength{\parskip}{0pt plus 1pt}
    \raggedright
    Perfect! Here's the final summary of what I'll be looking for:

    \begin{itemize}[nosep,leftmargin=1.2em]
      \item \textbf{Style}: A-line dress (no hood, no pockets)
      \item \textbf{Occasion}: Casual everyday wear
      \item \textbf{Color}: Black (preferred, open to others)
      \item \textbf{Pattern}: Solid
      \item \textbf{Fabric}: Cotton
      \item \textbf{Size}: Medium
      \item \textbf{Length}: No strong preference
      \item \textbf{Sleeves}: No strong preference
      \item \textbf{Edge finishes}: No strong preference
      \item \textbf{Shoulder style}: No strong preference
      \item \textbf{Closure}: No strong preference
      \item \textbf{Budget}: No strong preference
    \end{itemize}

    I think I have everything I need to find you a great option! Is there anything else you'd like to add before I go search?
    \endgroup
  \end{walkmessage}
\end{walkagentturn}

\section{Additional related work}\label{app:related_work}

\paragraph{Strategies for resolving underspecification.}
A growing body of work studies agents that must resolve underspecified requests through interaction~\citep{li2024underspecification, yao2024underspecified, luo2024clarification, shao2024collaborative, qian2025underspecification, pan2025clarification, vijayvargiya2025interactive}. In the shopping domain specifically, most benchmarks are non-interactive~\citep{yao2022webshop, zhou2023webarena, lyu2025deepshop, peeters2025webmall, wang2026shopsimulator}, and the interactive exceptions, such as \citet{murakhovs2023salespeople,chen2024chatshop}, still assume a user who can provide any preference upon questioning. The common thread is that specification is treated as \emph{retrieval} of a preexisting preference, done primarily through clarifying questions. Our contribution is to model specification as \emph{construction}: the user's preference state does not exist in complete form at the outset but is built through the agent's choice of clarifying questions, examples, and explanations.

\paragraph{Conversational recommender systems.}
While traditional recommender systems rely on user-item interactions, like ratings, in order to elicit user preferences, conversational recommendation uses language to elicit user intent and preferences~\citep{christakopoulou2016towards,radlinski2017conversational,jannach2021survey, gao2021advances}. A substantial line of work studies \emph{how} to elicit preferences, such as question-generation strategies~\citep{mirzadeh2005feature,iovine2019connect,zou2020towards,ren2021learning} and strategies for selecting items to elicit user feedback~\citep{zhang2020conversational,li2021seamlessly}.
A recurring finding is that users cannot simply answer clarifying questions about catalog features, as if filling in slots in the system's mental model; instead, some features are easier for users to answer questions about than others~\citep{kostric2021soliciting,shen2024multi}.
By using \frameworkname users with an SEC decomposition, \name is a benchmark aligned with this strain of thinking: rather than simply evaluating if the agent can identify what catalog slots to ask about, we evaluate if the agent can provide the necessary context to help the user construct a preference.

\paragraph{User simulators.}
A body of work studies how to \textbf{simulate users} for training and evaluating task-oriented agents, both for conversational recommendation and in broader tasks \citep{shi2019build,zhang2025survey,naous2025flipping,abdulhai2026consistently,suh2026quantifying,wu2026humanlm}. 
Naive LLM-based simulators have been shown to underrepresent behavioral diversity~\citep{yoon2024evaluating}. 
To improve simulator faithfulness, some works take a bottom-up, data-driven approach (i.e., finetuning language models on user utterances).
We instead take a top-down, model-based approach (i.e., defining a model of human behavior to condition a language model on).

Prior work also finds that if shown $x^\ast$, simulators may leak titles~\citep{zhu2024dont} or extra features that were not asked about. We avoid these problems by restricting the context of the LLM to exactly the current preference state $S_i$, which does not include features like titles that uniquely identify the target item.


\paragraph{Mixed-initiative systems.}
The question of who drives an interaction---and when control should pass between human and system---predates conversational LLMs. \citet{allen1999mixed, horvitz1999mixed, guinn1999mixed} frame mixed-initiative interaction as the interleaving of contributions so that each agent does what it is best suited to at the appropriate moment. Within recommendation specifically, \citet{ma2023initiative} study initiative transfer and find that the most effective starting mode depends on the user, and \citet{aliannejadi2021analysing} show that conversational search strategies---clarify-then-execute versus execute-then-clarify---have nuanced, user-dependent payoffs. Our three dialog actions (clarifying questions, example items, and explanations) are precisely initiative-taking moves.

\end{document}